\RequirePackage{fix-cm}
\documentclass[smallextended]{svjour3}       % onecolumn (second format)
\smartqed  % flush right qed marks, e.g. at end of proof
\usepackage{graphicx}
\usepackage{amssymb}
\usepackage{mathdots}
\usepackage[classicReIm]{kpfonts}
\usepackage{array}
\usepackage[square,numbers]{natbib}
\usepackage{subfigure}
\usepackage{wrapfig}
\usepackage{wasysym}
\usepackage{enumitem}
\usepackage{adjustbox}
\usepackage{ragged2e}
\usepackage[svgnames,table]{xcolor}
\usepackage{tikz}
\usepackage{longtable}
\usepackage{changepage}
\usepackage{setspace}
\usepackage{hhline}
\usepackage{multicol}
\usepackage{tabto}
\usepackage{float}
\usepackage{multirow}
\usepackage{makecell}
\usepackage{fancyhdr}
\usepackage[toc,page]{appendix}
\usepackage[hidelinks]{hyperref}
\usetikzlibrary{shapes.symbols,shapes.geometric,shadows,arrows.meta}
\tikzset{>={Latex[width=1.5mm,length=2mm]}}
\usepackage[utf8]{inputenc}
\usepackage[T1]{fontenc}
\TabPositions{0.14in,0.28in,0.42in,0.56in,0.7in,0.84in,0.98in,1.12in,1.26in,1.4in,1.54in,1.68in,1.82in,1.96in,2.1in,2.24in,2.38in,2.52in,2.66in,2.8in,2.94in,3.08in,3.22in,3.36in,3.5in,3.64in,3.78in,3.92in,4.06in,4.2in,4.34in,4.48in,4.62in,4.76in,4.9in,5.04in,5.18in,5.32in,5.46in,5.6in,5.74in,5.88in,6.02in,6.16in,6.3in,6.44in,6.58in,6.72in,6.86in,7.0in,7.14in,}
\begin{document}
{\noindent\onecolumn{
\noindent\textcopyright 2020 Springer. Personal use of this material is permitted. Permission from Springer must be obtained for all other uses, in any current or future media, including reprinting/republishing this material for advertising or promotional purposes, creating new collective works, for resale or redistribution to servers or lists, or reuse of any copyrighted component of this work in other works.}}

\newpage
\title{Effective Fusion of Deep Multitasking Representations for Robust Visual Tracking}
%%%%%%%%%%%%%%%%%%%%%%%%%%%%%%%%%%%%%%%%%%%%%%%%%%%%%%%%%%%%%%%%%%%%%%%%%%%%%%%%%%%%%%%%%%%%%%%%%%%%%%%%%%
\author{Seyed~Mojtaba~Marvasti-Zadeh    \and
        Hossein~Ghanei-Yakhdan    \and
        Shohreh~Kasaei    \and
        Kamal~Nasrollahi    \and
        Thomas~B.~Moeslund
}
\institute{  \textbullet\ S. M. Marvasti-Zadeh \at
              Digital Image and Video Processing Lab (DIVPL), Department of Electrical Engineering, Yazd University, Yazd, Iran. He is also a member of Image Processing Lab (IPL), Sharif University of Technology, Tehran, Iran, and Vision and Learning Lab, University of Alberta, Edmonton, Canada (\email{mojtaba.marvasti@ualberta.ca}). \\  
             \textbullet\ H. Ghanei-Yakhdan (corresponding author) \at
              Digital Image and Video Processing Lab (DIVPL), Department of Electrical Engineering, Yazd University, Yazd, Iran (\email{hghaneiy@yazd.ac.ir}). \\
            \textbullet\ S. Kasaei \at
              Image Processing Lab (IPL), Department of Computer Engineering, Sharif University of Technology, Tehran, Iran (\email{kasaei@sharif.edu}).\\
            \textbullet\ K. Nasrollahi \at
              Visual Analysis of People Lab (VAP), Department of Architecture, Design, and Media Technology, Aalborg University, Aalborg, Denmark (\email{kn@create.aau.dk}). \\ 
            \textbullet\ T. B. Moeslund \at
              Visual Analysis of People Lab (VAP), Department of Architecture, Design, and Media Technology, Aalborg University, Aalborg, Denmark (\email{tbm@create.aau.dk}). 
}
%%%%%%%%%%%%%%%%%%%%%%%%%%%%%%%%%%%%%%%%%%%%%%%%%%%%%%%%%%%%%%%%%%%%%%%%%%%%%%%%%%%%%%%%%%%%%%%%%%%%%%%%%%
\date{Received: date / Accepted: date}
\maketitle
%%%%%%%%%%%%%%%%%%%%%%%%%%%%%%%%%%%%%%%%%%%%%%%%%%%%%%%%%%%%%%%%%%%%%%%%%%%%%%%%%%%%%%%%%%%%%%%%%%%%%%%%%%
\begin{abstract}
Visual object tracking remains an active research field in computer vision due to persisting challenges with various problem-specific factors in real-world scenes. Many existing tracking methods based on discriminative correlation filters (DCFs) employ feature extraction networks (FENs) to model the target appearance during the learning process. However, using deep feature maps extracted from FENs based on different residual neural networks (ResNets) has not previously been investigated. This paper aims to evaluate the performance of twelve state-of-the-art ResNet-based FENs in a DCF-based framework to determine the best for visual tracking purposes. First, it ranks their best feature maps and explores the generalized adoption of the best ResNet-based FEN into another DCF-based method. Then, the proposed method extracts deep semantic information from a fully convolutional FEN and fuses it with the best ResNet-based feature maps to strengthen the target representation in the learning process of continuous convolution filters. Finally, it introduces a new and efficient semantic weighting method (using semantic segmentation feature maps on each video frame) to reduce the drift problem. Extensive experimental results on the well-known OTB-2013, OTB-2015, TC-128, and VOT-2018 visual tracking datasets demonstrate that the proposed method effectively outperforms state-of-the-art methods in terms of precision and robustness of visual tracking. 
%%%%%%%%%%%%%%%%%%%%%%%%%%%%%%%%%%%%%%%%%%%%%%%%%%%%%%%%%%%%%%%%%%%%%%%%%%%%%%%%%%%%%%%%%%%%%%%%%%%%%%%%%%
\keywords{Appearance modeling \and discriminative correlation filters \and deep convolutional neural networks \and robust visual tracking}
\end{abstract}
%%%%%%%%%%%%%%%%%%%%%%%%%%%%%%%%%%%%%%%%%%%%%%%%%%%%%%%%%%%%%%%%%%%%%%%%%%%%%%%%%%%%%%%%%%%%%%%%%%%%%%%%%%
\vspace{-4mm}
\section{Introduction}
\label{intro}
\vspace{-2mm}
Generic visual tracking aims to estimate the trajectory of an arbitrary visual target (given the initial state in the first frame) over time despite many challenging factors, including fast motion, background clutter, deformation, and occlusion. This constitutes a fundamental problem in many computer vision applications (e.g., self-driving cars, autonomous robots, human-computer interactions). Extracting a robust target representation is the critical component of state-of-the-art visual tracking methods to overcome these challenges. Hence, to robustly model target appearance, these methods utilize a wide range of handcrafted features (e.g., \cite{HC-HCOF,HC-SA,HC-HCF,HC-MS,HC-FM,APIN1} which exploit histogram of oriented gradients (HOG) \cite{HOG}, histogram of local intensities (HOI), and Color Names (CN) \cite{CN}), deep features from deep neural networks (e.g., \cite{HCFT,HCFTs,DCFNet,SiamFC,APIN2,APIN3}, or both (e.g., \cite{DeepSRDCF,ECO,LCTdeep}). 

Deep features are generally extracted from either FENs (i.e., pre-trained deep convolutional neural networks (CNNs)) or end-to-end networks (EENs), which directly evaluate target candidates \cite{SurveyDeepTracking}. However, most EEN-based visual trackers train or fine-tune FENs on visual tracking datasets. Numerous recent visual tracking methods exploit powerful generic target representations from FENs trained on large-scale object recognition datasets, such as the ImageNet \cite{ImageNet}. By combining deep discriminative features from FENs with efficient online learning formulations, visual trackers based on discriminative correlation filters (DCF) have achieved significant performance in terms of accuracy, robustness, and speed. Despite the existence of modern CNN models, which are mainly based on the ResNet architectures \cite{ResNet}, most DCF-based visual trackers still use VGG-M \cite{VGGM}, VGG-16 \cite{VGGNet}, and VGG-19 \cite{VGGNet} models, which have a simple stacked multi-layer topology with moderate representation capabilities. These models provide a limited power of representation compared to the state-of-the-art ResNet-based models such as ResNet \cite{ResNet}, ResNeXt \cite{ResNeXt}, squeeze-and-excitation networks (SENets) \cite{SE-Res-CVPR,SE-Res-PAMI}, and DenseNet \cite{DenseNet} models. Besides, most of the visual tracking methods exploit well-known CNN models which are trained for the visual recognition task. However, the transferability of deep representations from other related tasks (e.g., semantic segmentation or object detection) may help the DCF-based visual trackers to improve their robustness. Motivated by these two goals, the proposed paper surveys the performance of state-of-the-art ResNet-based FENs and exploits deep multitasking representations for visual tracking purposes.

The main contributions of the paper are as follows. \\
1) The effectiveness of twelve state-of-the-art ResNet-based FENs is evaluated: ResNet-50 \cite{ResNet}, ResNet-101 \cite{ResNet}, ResNet-152 \cite{ResNet}, ResNeXt-50 \cite{ResNeXt}, ResNeXt-101 \cite{ResNeXt}, SE-ResNet-50 \cite{SE-Res-CVPR,SE-Res-PAMI}, SE-ResNet-101 \cite{SE-Res-CVPR,SE-Res-PAMI}, SE-ResNeXt-50 \cite{SE-Res-CVPR,SE-Res-PAMI}, SE-ResNeXt-101 \cite{SE-Res-CVPR,SE-Res-PAMI}, DenseNet-121 \cite{DenseNet}, DenseNet-169 \cite{DenseNet}, and DenseNet-201 \cite{DenseNet}. These have been trained on the large-scale ImageNet dataset. To the best of our knowledge, this is the first paper to comprehensively survey the performance of ResNet-based FENs for visual tracking purposes. \\
2) The generalization of the best ResNet-based FEN is investigated by a different DCF-based visual tracking method. \\
3) A visual tracking method is proposed that fuses deep representations; these are extracted from the best trained ResNet-based network and the FCN-8s network \cite{FCN-8s}, which have been trained on the PASCAL VOC \cite{PascalVOC} and MSCOCO \cite{MSCOCO} datasets for semantic image segmentation tasks. \\
4) The proposed method uses the extracted deep features from the FCN-8s network to semantically weight the target representation in each video frame. \\
5) The performance of the proposed method is extensively compared with state-of-the-art visual tracking methods on four well-known visual tracking datasets.

The rest of the paper is organized as follows. In Section \ref{sec:2}, an overview of related work, including DCF-based visual trackers exploiting deep features from the FENs, is outlined. In Section \ref{sec:3} and Section \ref{sec:4}, the survey of ResNet-based FENs and the proposed visual tracking method are presented, respectively. In Section \ref{sec:5}, extensive experimental results on the large visual tracking datasets are given. Finally, the conclusion is summarized in Section \ref{sec:6}.
%%%%%%%%%%%%%%%%%%%%%%%%%%%%%%%%%%%%%%%%%%%%%%%%%%%%%%%%%%%%%%%%%%%%%%%%%%%%%%%%%%%%%%%%%%%%%%%%%%%%%%%%%%
\vspace{-4mm}
\section{Related Work}
\label{sec:2}
\vspace{-2mm}
In this section, the exploitation of different FENs by state-of-the-art visual trackers is categorized and described. Although some visual tracking methods use recent FENs (e.g., R-CNN \cite{RCNN} in \cite{CNN-SVM}), popular FENs in the most visual trackers are VGG-M, VGG-16, and VGG-19 models (see Table 1). These networks provide moderate accuracy and complexity for visual target modeling.\\
%%%%%%%%%%%%%%%%%%%%%%%%%%%%%%%%%%%%%%%%%%%%%%%%%%%%%%%%%%%%%%%%%%%%%%%%%%%%%%%%%%%%%%%%%%%%%%%%%%%%%%%%
\begin{table*}\label{tab1}
\caption{Exploited FENS in some visual tracking methods.} % title of Table
\vspace{-2mm}
\centering
\resizebox{\textwidth}{!}{
\begin{tabular}{c c c c}
\hline\hline
Visual Tracking Method & Model & Pre-training Dataset & Name of Exploited Layer(s) \\
\hline
DeepSRDCF \cite{DeepSRDCF} & VGG-M & ImageNet & Conv1 \\
C-COT \cite{CCOT} & VGG-M & ImageNet & Conv1, Conv5 \\
ECO \cite{ECO} & VGG-M & ImageNet & Conv1, Conv5 \\
DeepSTRCF \cite{STRCF} & VGG-M & ImageNet & Conv3 \\
WAEF \cite{WAEF} & VGG-M &	ImageNet &	Conv1, Conv5 \\
WECO \cite{WECO} & VGG-M &	ImageNet &	Conv1, Conv5 \\
VDSR-SRT \cite{VDSR-SRT} & VGG-M &	ImageNet &	Conv1, Conv5 \\
RPCF \cite{RPCF} & VGG-M &	ImageNet &	Conv1, Conv5 \\
DeepTACF \cite{DeepTACF} & VGG-M &	ImageNet &	Conv1 \\
ASRCF \cite{ASRCF} & VGG-M, VGG-16 & ImageNet &	Norm1, Conv4-3 \\
DRT \cite{DRT} & VGG-M, VGG-16 & ImageNet &	Conv1, Conv4-3 \\
ETDL \cite{ETDL} & VGG-16 & ImageNet & Conv1-2 \\
FCNT \cite{FCNT} & VGG-16 & ImageNet & Conv4-3, Conv5-3 \\
Tracker \cite{DeepMotionFeatures} & VGG-16 & ImageNet & Conv2-2, Conv5-3 \\
DNT \cite{DNT} & VGG-16 & ImageNet & Conv4-3, Conv5-3 \\
CREST \cite{CREST} & VGG-16 & ImageNet &	Conv4-3 \\
CPT \cite{CPT} & VGG-16 & ImageNet & Conv5-1, Conv5-3 \\
DeepFWDCF \cite{DeepFWDCF} & VGG-16 & ImageNet & Conv4-3 \\
DTO \cite{DTO} & VGG-16, SSD &	ImageNet &	Conv3-3, Conv4-3, Conv5-3 \\
DeepHPFT \cite{DeepHPFT} & VGG-16, VGG-19, and GoogLeNet &	ImageNet &	Conv5-3, Conv5-4, and icp6-out \\
HCFT \cite{HCFT} & VGG-19 & ImageNet & Conv3-4, Conv4-4, Conv5-4 \\
HCFTs \cite{HCFTs} & VGG-19 & ImageNet & Conv3-4, Conv4-4, Conv5-4 \\
LCTdeep \cite{LCTdeep} & VGG-19 & ImageNet & Conv5-4 \\
Tracker \cite{CF-CNN} & VGG-19 & ImageNet & Conv3-4, Conv4-4, Conv5-4 \\
HDT \cite{HDT} & VGG-19 & ImageNet & Conv4-2, Conv4-3, Conv4-4, Conv5-2, Conv5-3, Conv5-4 \\
IBCCF \cite{IBCCF} & VGG-19 & ImageNet &	Conv3-4, Conv4-4, Conv5-4 \\
DCPF \cite{DCPF} &	VGG-19 & ImageNet &	Conv3-4, Conv4-4, Conv5-4 \\
MCPF \cite{MCPF} &	VGG-19 & ImageNet &	Conv3-4, Conv4-4, Conv5-4 \\
DeepLMCF \cite{DeepLMCF} &	VGG-19 & ImageNet &	Conv3-4, Conv4-4, Conv5-4 \\
STSGS \cite{STSGS} & VGG-19 & ImageNet &	Conv3-4, Conv4-4, Conv5-4 \\
MCCT \cite{MCCT} &	VGG-19 & ImageNet &	Conv4-4, Conv5-4 \\
DCPF2 \cite{DCPF2} & VGG-19 & ImageNet & Conv3-4, Conv4-4, Conv5-4 \\
ORHF \cite{ORHF} &	VGG-19 & ImageNet &	Conv3-4, Conv4-4, Conv5-4 \\
IMM-DFT \cite{IMM-DFT} & VGG-19 & ImageNet & Conv3-4, Conv4-4, Conv5-4 \\
MMLT \cite{MMLT} & VGGNet, Fully-convolutional Siamese network & ImageNet, ILSVRC-VID  & Conv5 \\	
CNN-SVM \cite{CNN-SVM} & R-CNN & ImageNet &	First fully-connected layer \\
TADT \cite{TADT} &	Siamese matching network &	ImageNet &	Conv4-1, Conv4-3 \\
\hline
\end{tabular}
}
\vspace{-4mm}
\end{table*}
%%%%%%%%%%%%%%%%%%%%%%%%%%%%%%%%%%%%%%%%%%%%%%%%%%%%%%%%%%%%%%%%%%%%%%%%%%%%%%%%%%%%%%%%%%%%%%%%%%%%%%%%
\indent\textbf{VGG-M Model:} With the aid of deep features and spatial regularization weights, the spatially regularized discriminative correlation filters (DeepSRDCF) \cite{DeepSRDCF} learn more discriminative target models on larger image regions. The formulation of DeepSRDCF provides a larger set of negative training samples by penalizing the unwanted boundary effects resulting from the periodic assumption of standard DCF-based methods. Moreover, the method based on deep spatial-temporal regularized correlation filters (DeepSTRCF) \cite{STRCF} incorporates both spatial and temporal regularization parameters to provide a more robust appearance model. Then, it iteratively optimizes three closed-form solutions via the alternating direction method of multipliers (AD-MM) algorithm \cite{ADMM}. To explore efficiency of Tikhonov regularization in temporal domain, the weighted aggregation with enhancement filter (WAEF) \cite{WAEF} organizes deep features according to the average information content and suppresses uncorrelated frames for visual tracking. The continuous convolution operator tracker (C-COT) \cite{CCOT} fuses multi-resolution deep feature maps to learn discriminative continuous-domain convolution operators. It also exploits an implicit interpolation model to enable accurate sub-pixel localization of the visual target. To reduce the computational complexity and the number of training samples and to improve the update strategy of the C-COT, the ECO tracker \cite{ECO} uses a factorized convolution operator, a compact generative model of training sample distribution, and a conservative model update strategy. The weighted ECO \cite{WECO} utilizes weighted sum operation and normalization of deep features to take advantages of multi-resolution deep features. By employing the ECO framework, the VDSR-SRT method \cite{VDSR-SRT} uses a super-resolution reconstruction algorithm to robustly track targets in low-resolution images. To compress model size and improve the robustness against deformations, the region of interest (ROI) pooled correlation filters (RPCF) \cite{RPCF} accurately localizes a target while it uses deep feature maps with smaller sizes. The target-aware correlation filters (TACF) method \cite{DeepTACF} guides the learned filters for visual tracking by focusing on a target and preventing from background information.\\
\indent\textbf{VGG-16 Model:} By employing the DeepSRDCF tracker, Gladh et al. \cite{DeepMotionFeatures} have investigated the fusion of handcrafted appearance features (e.g., HOG and CN) with deep RGB and motion features in the DCF-based visual tracking framework. This method demonstrates the effectiveness of deep feature maps for visual tracking purposes. Besides, the CREST method \cite{CREST} reformulates correlation filters to extract more beneficial deep features by integrating the feature extraction and learning process of DCFs. The dual network-based tracker (DNT) \cite{DNT} designs a dual structure for embedding the boundary and shape information into the feature maps to better utilize deep hierarchical features. The deep fully convolutional networks tracker (FCNT) \cite{FCNT} uses two complementary feature maps and a feature-map selection method to design an effective FEN-based visual tracker. It exploits two different convolutional layers for category detection and distraction separation. Moreover, the feature-map selection method helps the FCNT to reduce computation redundancy and discard irrelevant feature maps. The method based on enhanced tracking and detection learning (ETDL) \cite{ETDL} employs different color bases for each color frame, adaptive multi-scale DCF with deep features, and a detection module to re-detect the target in failure cases. To prevent unexpected background information and distractors, the adaptive feature weighted DCF (FWDCF) \cite{DeepFWDCF} calculates target likelihood to provide spatial-temporal weights for deep features. The channel pruning method (CPT) \cite{CPT} utilizes average feature energy ratio method to exploit low-dimensional deep features, adaptively. \\
\indent\textbf{VGG-19 Model:} Similar to the FCNT, the hierarchical correlation feature-based tracker (HCFT) \cite{HCFT} exploits multiple levels of deep feature maps to handle considerable appearance variation and simultaneously to provide precise localization. Furthermore, the modified HCFT (namely HCFTs or HCFT*) \cite{HCFTs} not only learns linear correlation filters on multi-level deep feature maps; it also employs two types of region proposals and a discriminative classifier to provide a long-term memory of target appearance. Furthermore, the IMM-DFT method \cite{IMM-DFT} exploits adaptive hierarchical features to interactively model a target due to the insufficiency of linear combination of deep features. To incorporate motion information with spatial deep features, the STSGS method \cite{STSGS} solves a compositional energy optimization to effectively localize an interested target. Also, the MCCT method \cite{MCCT} learns different target models by multiple DCFs that adopt different features, and then it makes the decision based on the reliable result. The hedged deep tracker (HDT) \cite{HDT} ensembles weak CNN-based visual trackers via an online decision-theoretical Hedge algorithm, aiming to exploit the full advantage of hierarchical deep feature maps. The ORHF method \cite{ORHF} selects useful deep features according to the estimated confidence scores to reduce the computational complexity problem. To enhance the dicrimination power of target among its background, the DCPF \cite{DCPF} and DeepLMCF \cite{DeepLMCF} exploit deep features into the particle filter and structured SVM based tracking, respectively. To handle significant appearance change and scale variation, the deep long-term correlation tracking (LCTdeep) \cite{LCTdeep} uses multiple adaptive DCF, deep feature pyramid reconfiguration, aggressive and conservative learning rates as short- and long-term memories combined with an incrementally learned detector to recover tracking failures.
Furthermore, Ma et al. \cite{CF-CNN} exploited different deep feature maps and a conservative update scheme to learn the mapping as a spatial correlation and maintain the long-memory of target appearance. At last, the DCPF2 \cite{DCPF2} and IBCCF \cite{IBCCF} methods aims to handle the aspect ratio of a target for deep DCF-based trackers. \\
\indent\textbf{Custom FENs:} In addition to the mentioned popular FENs, some methods exploit either a combination of FENs (e.g., ASRCF \cite{ASRCF}, DRT \cite{DRT}, DTO \cite{DTO},  MMLT \cite{MMLT}, and DeepHPFT \cite{DeepHPFT}) or other types of FENs (e.g., CNN-SVM \cite{CNN-SVM}, and TADT \cite{TADT}) for visual tracking. The ASRCF \cite{ASRCF} and DRT \cite{DRT} methods modify the DCF formulation to achieve more reliable results. To alleviate inaccurate estimations and scale drift, the DeepHPFT \cite{DeepHPFT} ensembles various handcrafted and deep features into the particle filter framework. Although the DTO method \cite{DTO} employs a trained object detection model (i.e., single shot detector (SSD) \cite{SSD}) to evaluate the impact of object category estimation for visual tracking, its ability is limited to some particular categories. The MMLT method \cite{MMLT} employs the Siamese and VGG networks to not only robustly track a target but also provide a long-term memory of appearance variation. On the other hand, the CNN-SVM \cite{CNN-SVM} and TADT \cite{TADT} methods use different models to improve the effectiveness of FENs for visual tracking.\\
\indent According to the Table 1, the recent visual tracking methods do not only use the state-of-the-art FENs as their feature extractor; they also exploit different feature maps in the DCF-based framework. In the next section, the popular ResNet-based FENs will be surveyed. Then, the proposed visual tracking method, with the aid of fused deep representation, will be described.
%%%%%%%%%%%%%%%%%%%%%%%%%%%%%%%%%%%%%%%%%%%%%%%%%%%%%%%%%%%%%%%%%%%%%%%%%%%%%%%%%%%%%%%%%%%%%%%%%%%%%%%%%%
\vspace{-4mm}
\section{Survey of ResNet-based FENs}
\label{sec:3}
\vspace{-2mm}
This section has two main aims. First, twelve state-of-the-art ResNet-based FENs in the DCF-based framework are comprehensively evaluated for visual tracking purposes, and the best ResNet-based FEN and its feature maps are selected to be exploited in the DCF-based visual tracking framework. Second, the generalization of the best FEN is investigated on another DCF-based visual tracking method.
\vspace{-4mm}
\subsection{Performance Evaluation}
\label{sec:3.1}
\vspace{-2mm}
The ECO framework \cite{ECO} is employed as the baseline visual tracker. Aiming for accurate and fair evaluations, this tracker was modified to extract deep features from FENs with the directed acyclic graph topology and exploit all deep feature maps without any down sampling or dimension reduction. Table 2 shows the exploited ResNet-based FENs and their feature maps, which are used in the proposed method. All these FENs are evaluated by well-known precision and success metrics \cite{OTB2013,OTB2015} on the OTB-2013 dataset \cite{OTB2013}, which includes more than 29,000 video frames. The characteristics of visual tracking datasets which are used in this work are shown in Table 3. The OTB-2013, OTB-2015, and TC-128 visual tracking datasets \cite{OTB2013,OTB2015,TC128} have common challenging attributes, including illumination variation (IV), out-of-plane rotation (OPR), scale variation (SV), occlusion (OCC), deformation (DEF), motion blur (MB), fast motion (FM), in-plane rotation (IPR), out-of-view (OV), background clutter (BC), and low resolution (LR), which may occur for different classes of visual targets in real-world scenarios. Also, the VOT-2018 includes six main challenging attributes of camera motion, illumination change, motion change, occlusion, and size change which are annotated per each video frame. To measure visual tracking performance, the OTB-2013, OTB-2015, and TC-128 toolkits use precision and success plots to rank the methods according to the area under curve (AUC) metric while the VOT-2018 utilizes the accuracy-robustness (AR) plot to rank the visual trackers based on the TraX protocol \cite{TraX}. The precision metric is defined as the percentage of frames where the average Euclidean distance between the estimated and ground-truth locations is smaller than a given threshold (20 pixels in this work). Moreover, the overlap success metric is the percentage of frames where the average overlap score between the estimated and the ground-truth bounding boxes is more than a particular threshold (50\% overlap in this work). For the VOT-2018, the accuracy measure the overlap of the estimated bounding boxes with the ground-truth ones while the number of failures is considered as the robustness metric. Although the OTB-2013, OTB-2015, and TC-128 toolkits assess the visual trackers according to one-pass evaluation, the VOT-2014 detects the tracking failures of each method and restart the evaluations for each method after five frames of failure on every video sequence based on the TraX protocol.
%%%%%%%%%%%%%%%%%%%%%%%%%%%%%%%%%%%%%%%%%%%%%%%%%%%%%%%%%%%%%%%%%%%%%%%%%%%%%%%%%%%%%%%%%%%%%%%%%%%%%%%%
\begin{table*}\label{tab2}
\caption{Exploited state-of-the-art ResNet-based FENs and their feature maps [Level of feature map resolution is denoted as L].} % title of Table
\vspace{-2mm}
\centering
\resizebox{\textwidth}{!}{
\begin{tabular}{c c c c c} % centered columns (4 columns)
\hline \hline 
\multirow{2}{*}{Name of FENs} & \multicolumn{4}{c}{Name of Output Layers (Number of Feature Maps)} \\\cline{2-5}
 & L1 & L2 & L3 & L4 \\ \hline
ResNet-50 &	Conv1-relu (64) &	Res2c-relu (256) &	Res3d-relu (512) &	Res4f-relu (1024) \\  
ResNet-101 &	Conv1-relu (64) &	Res2c-relu (256) &	Res3b3-relu (512) &	Res4b22-relu (1024) \\  
ResNet-152 &	Conv1-relu (64) &	Res2c-relu (256) &	Res3b7-relu (512) &	Res4b35-relu (1024) \\  
ResNeXt-50 &	Features-2 (64) &	Features-4-2-id-relu (256) &	Features-5-3-id-relu (512) &	Features-6-5-id-relu (1024) \\  
ResNeXt-101 &	Features-2 (64)	 & Features-4-2-id-relu (256) &	Features-5-3-id-relu (512) &	Features-6-22-id-relu (1024) \\  
SE-ResNet-50 &	Conv1-relu-7x7-s2 (64) &	Conv2-3-relu (256) &	Conv3-4-relu (512) &	Conv4-6-relu (1024) \\  
SE-ResNet-101 &	Conv1-relu-7x7-s2 (64) &	Conv2-3-relu (256) &	Conv3-4-relu (512) &	Conv4-23-relu (1024) \\  
SE-ResNeXt-50 &	Conv1-relu-7x7-s2 (64) &	Conv2-3-relu (256) &	Conv3-4-relu (512) &	Conv4-6-relu (1024) \\  
SE-ResNeXt-101 &	Conv1-relu-7x7-s2 (64) &	Conv2-3-relu (256) &	Conv3-4-relu (512) &	Conv4-23-relu (1024) \\  
DenseNet-121 &	Features-0-relu0 (64) &	Features-0-transition1-relu (256) &	Features-0-transition2-relu (512) &	Features-0-transition3-relu (1024) \\  
DenseNet-169 &	Features-0-relu0 (64) &	Features-0-transition1-relu (256) &	Features-0-transition2-relu (512) &	Features-0-transition3-relu (1024) \\ 
DenseNet-201 &	Features-0-relu0 (64) &	Features-0-transition1-relu (256) &	Features-0-transition2-relu (512) &	Features-0-transition3-relu (1792) \\
\hline
\end{tabular}
}
\vspace{-2mm}
\end{table*}
%%%%%%%%%%%%%%%%%%%%%%%%%%%%%%%%%%%%%%%%%%%%%%%%%%%%%%%%%%%%%%%%%%%%%%%%%%%%%%%%%%%%%%%%%%%%%%%%%%%%%%%%
\begin{table*}\label{tab3}
\caption{Exploited visual tracking datasets in this work.} % title of Table
\vspace{-2mm}
\centering
\resizebox{\textwidth}{!}{
\begin{tabular}{c c c c c c c c c c c c c c} % centered columns (4 columns)
\hline \hline 
\multirow{2}{*}{Visual Tracking Dataset} & \multirow{2}{*}{Number of Videos} & \multirow{2}{*}{Number of Frames} &\multicolumn{11}{c}{Number of Videos Per Attribute} \\\cline{4-14}
 &  &  & IV  &	OPR  &	SV  &	OCC  &	DEF  &	MB  &	FM  &	IPR  &	OV  &	BC  &	LR\\ \hline
OTB-2013 \cite{OTB2013} &	51 &	29491 &	25 &	39 &	28 &	29 &	19 &	12 &	17 &	31 &	6 &	21 &	4  \\
OTB-2015 \cite{OTB2015} &	100 &	59040 &	38 &	63 &	65 &	49 &	44 &	31 &	40 &	53 &	14 &	31 &	10 \\
TC-128 \cite{TC128} &	129 &	55346 &	37 &	73 &	66 &	64 &	38 &	35 &	53 &	59 &	16 &	46 &	21 \\
VOT-2018 \cite{VOT-2018} &	70   &	25504   & \multicolumn{11}{c}{Frame-based attributes} \\
\hline
\end{tabular}
}
\vspace{-2mm}
\end{table*}
%%%%%%%%%%%%%%%%%%%%%%%%%%%%%%%%%%%%%%%%%%%%%%%%%%%%%%%%%%%%%%%%%%%%%%%%%%%%%%%%%%%%%%%%%%%%%%%%%%%%%%%%

To survey the performance of twelve ResNet-based FENs for visual tracking, the results of the comprehensive precision and success evaluations on the OTB-2013 dataset are shown in Fig. 1. On this basis, the L3 feature maps of these FENs have provided the best representation of targets, which is favorable for visual tracking. The deep features in the third level of ResNet-based FENs provide an appropriate balance between semantic information and spatial resolution and yet are invariant to significant appearance variations. The performance comparison of different ResNet-based FENs in terms of precision and success metrics is shown in Table 4. According to the results, the DenseNet-201 model and the feature maps extracted from the L3 layers have achieved the best performance for visual tracking purposes. 
%%%%%%%%%%%%%%%%%%%%%%%%%%%%%%%%%%%%%%%%%%%%%%%%%%%%%%%%%%%%%%%%%%%%%%%%%%%%%%%%%%%%%%%%%%%%%%%%%%%%%%%%
\begin{figure*}
\justify
\subfigure{\includegraphics[width=0.24\textwidth]{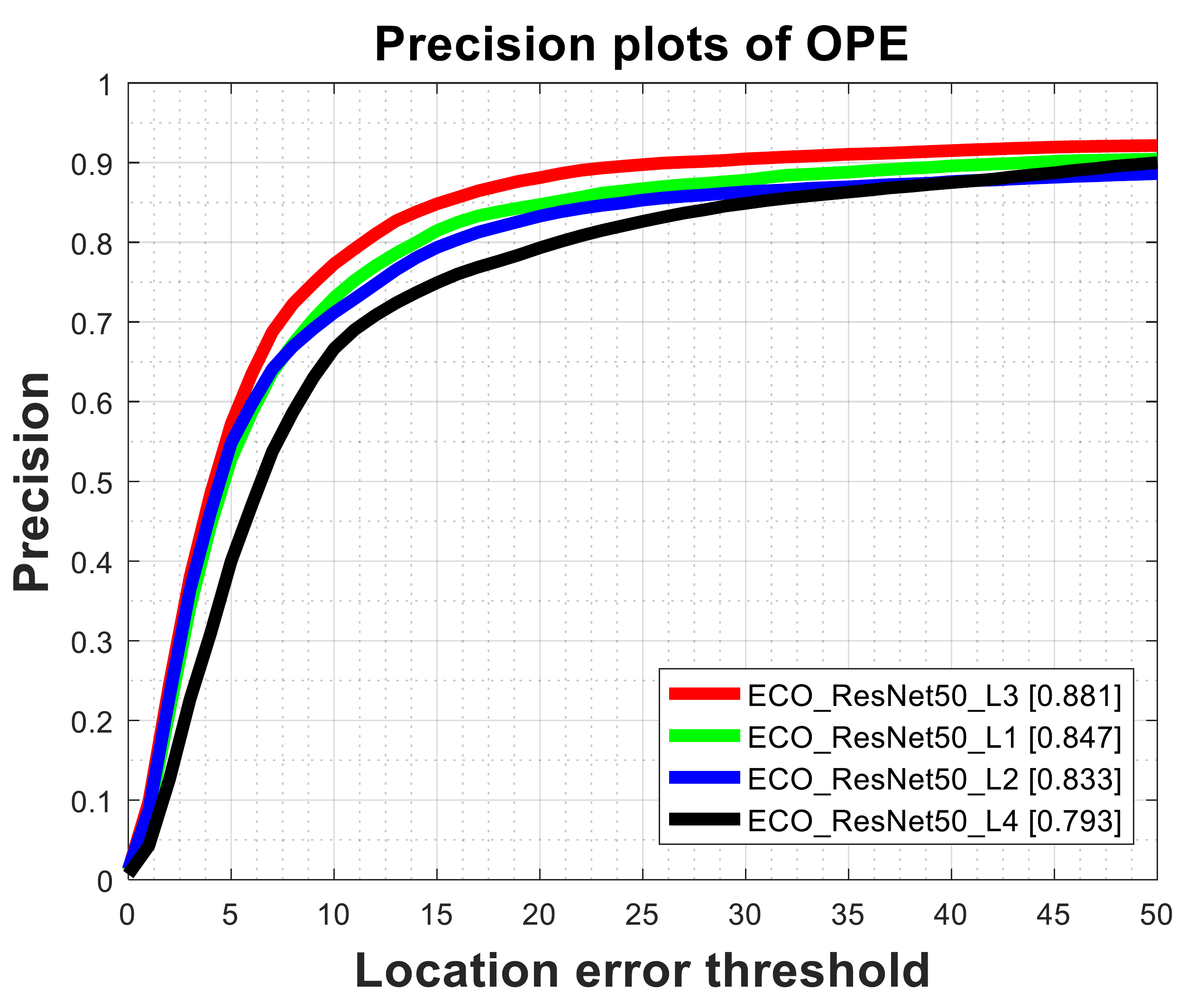}} 
\subfigure{\includegraphics[width=0.24\textwidth]{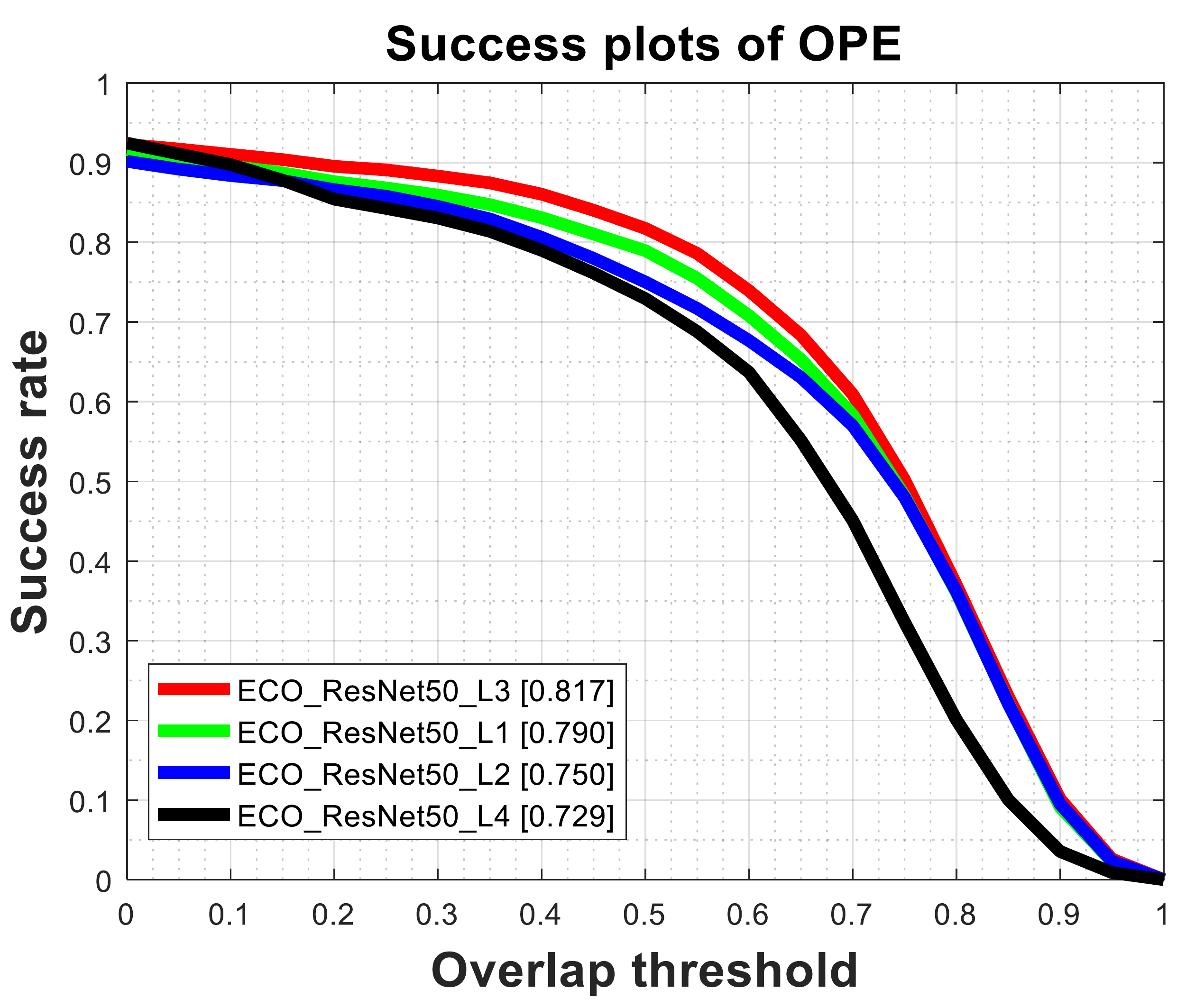}}
\subfigure{\includegraphics[width=0.24\textwidth]{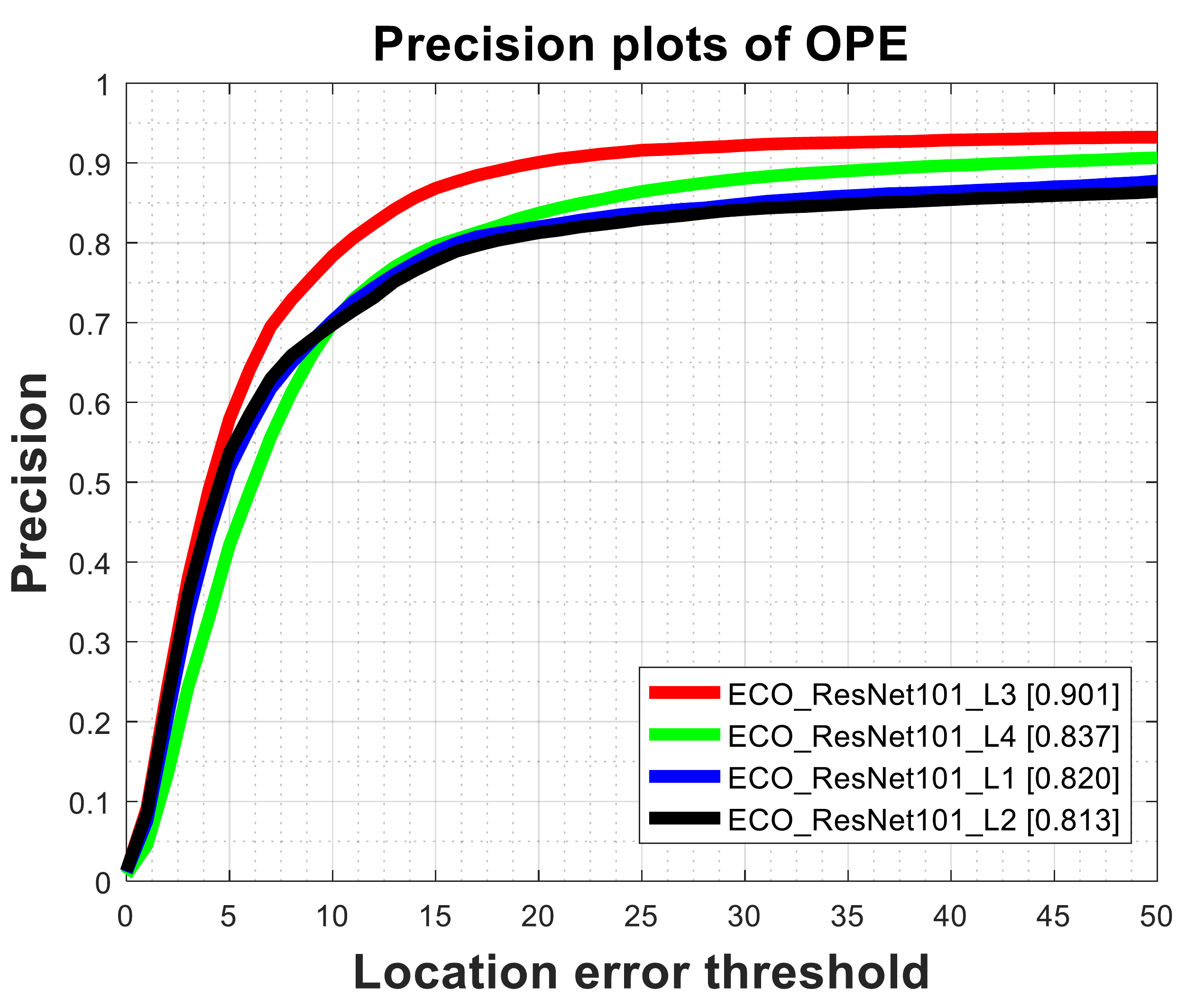}}
\subfigure{\includegraphics[width=0.24\textwidth]{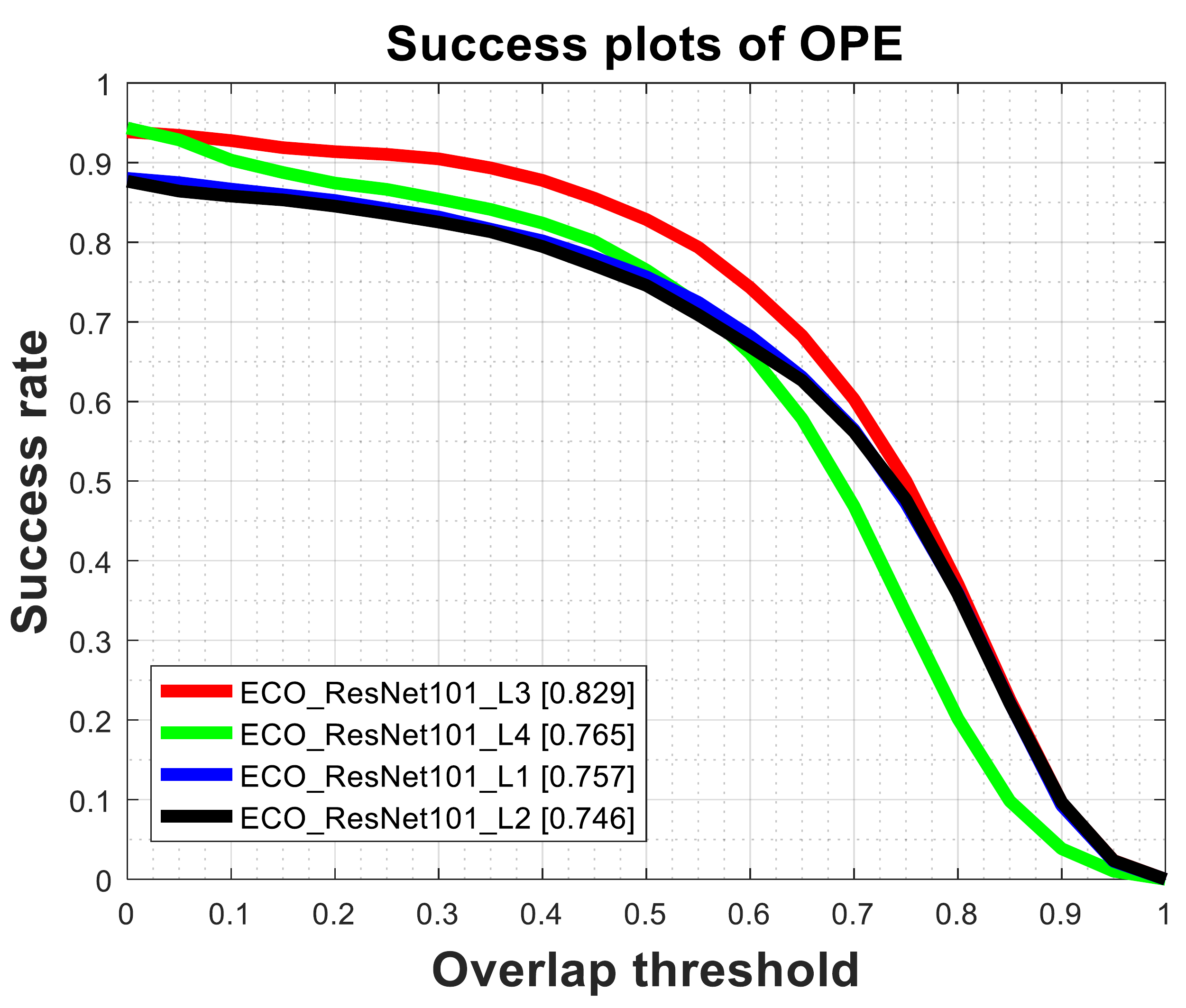}}
\vspace{-6mm}
\justify
\subfigure{\includegraphics[width=0.24\textwidth]{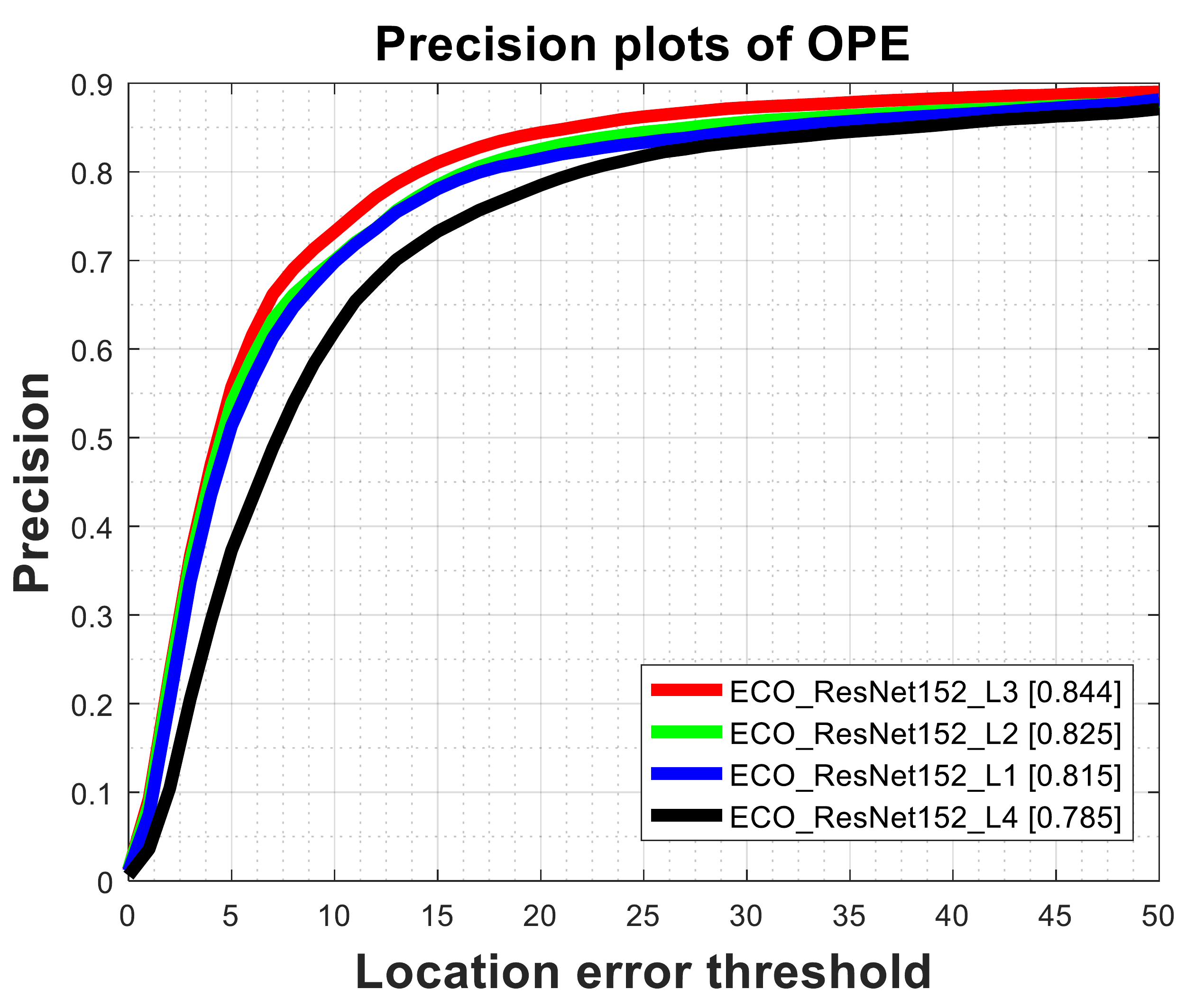}} 
\subfigure{\includegraphics[width=0.24\textwidth]{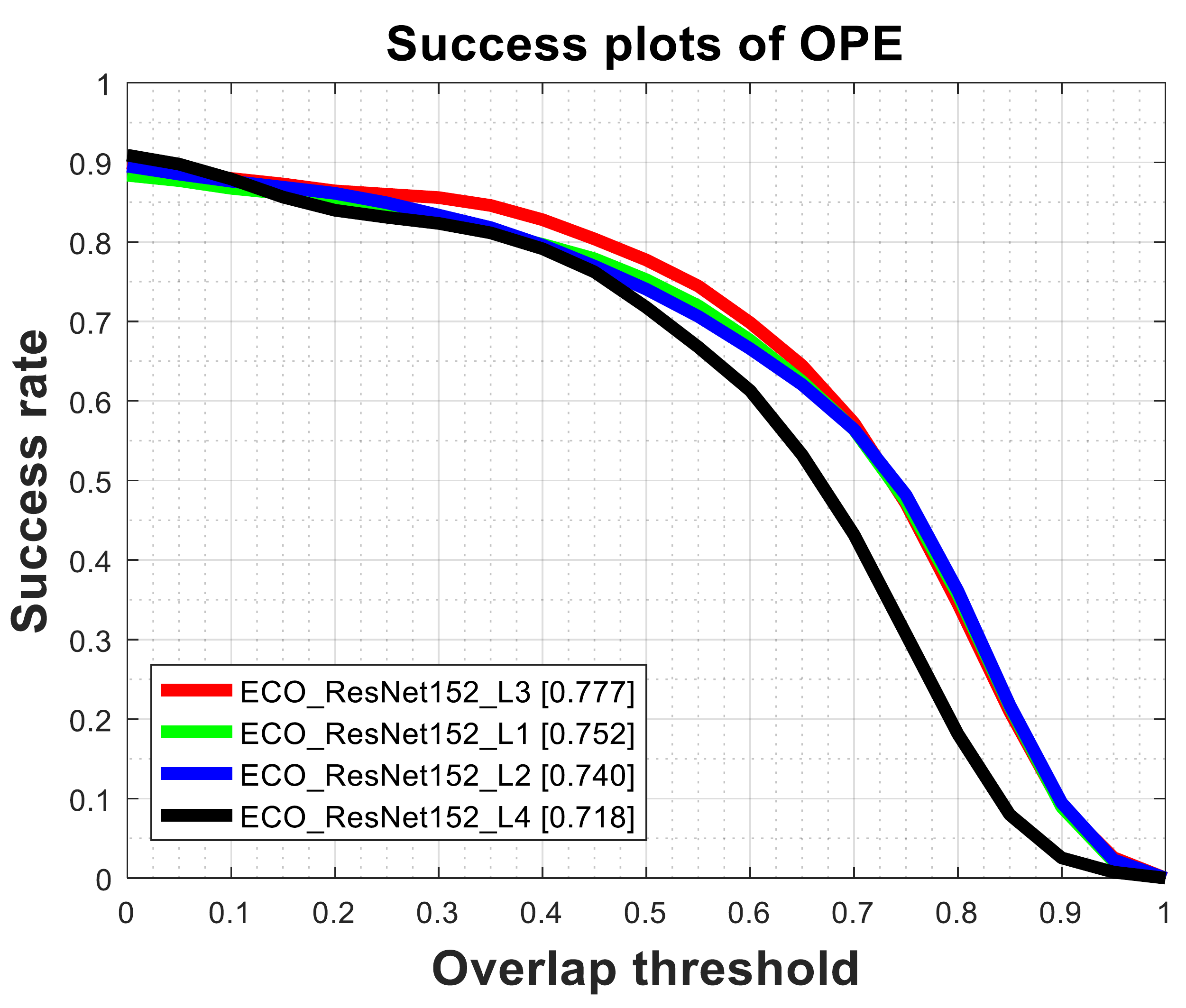}}
\subfigure{\includegraphics[width=0.24\textwidth]{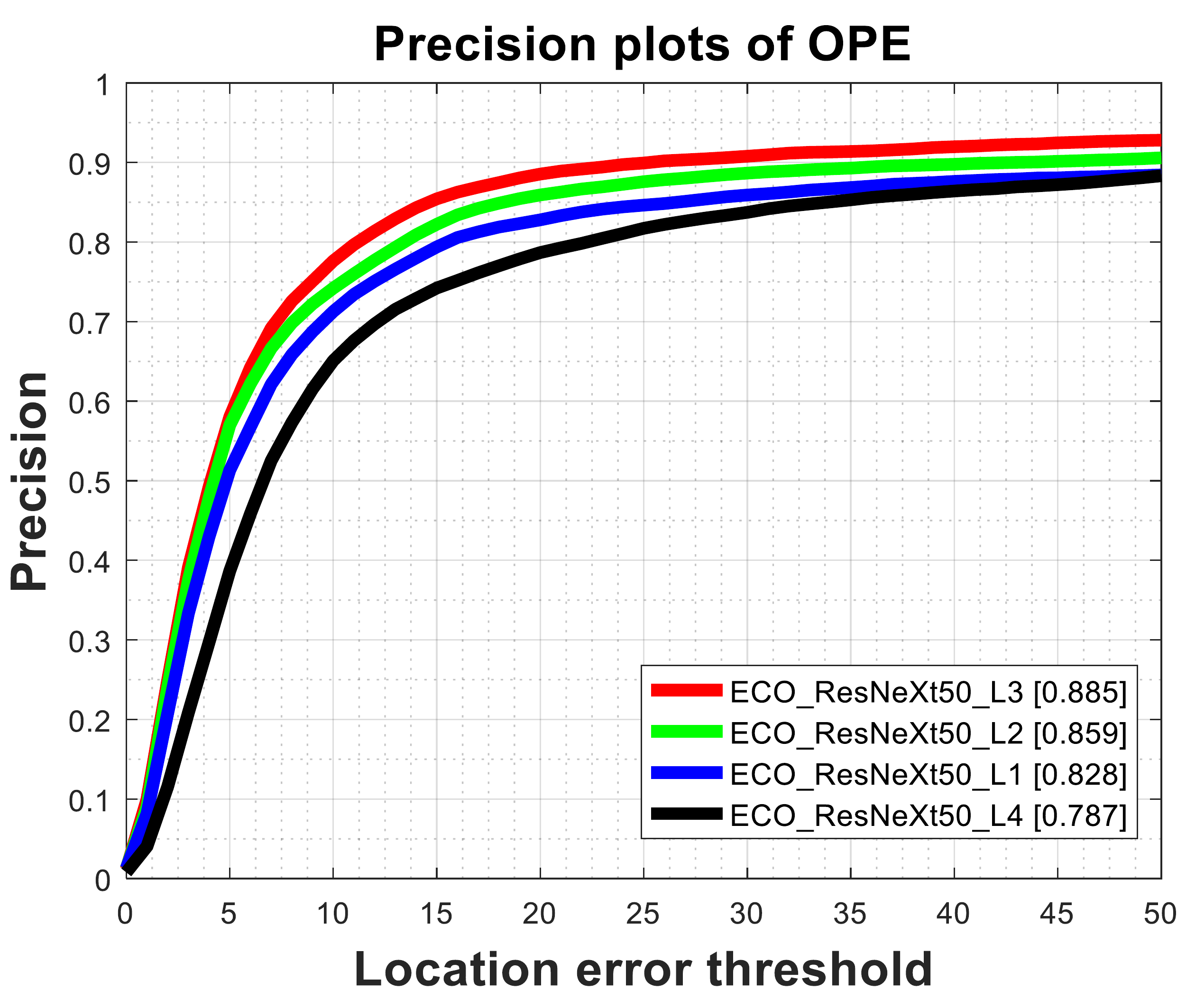}}
\subfigure{\includegraphics[width=0.24\textwidth]{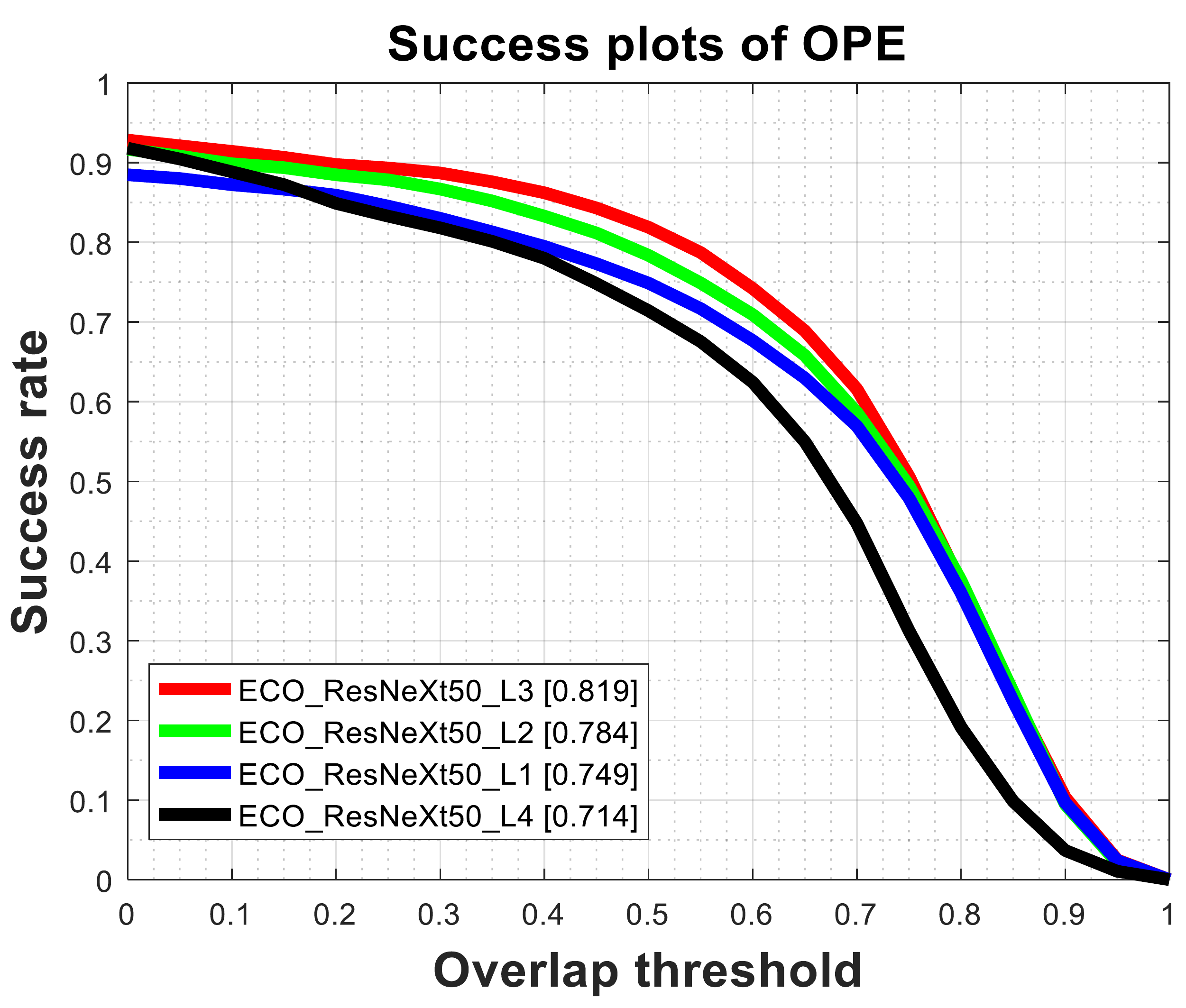}}
\vspace{-4mm}
\justify
\subfigure{\includegraphics[width=0.24\textwidth]{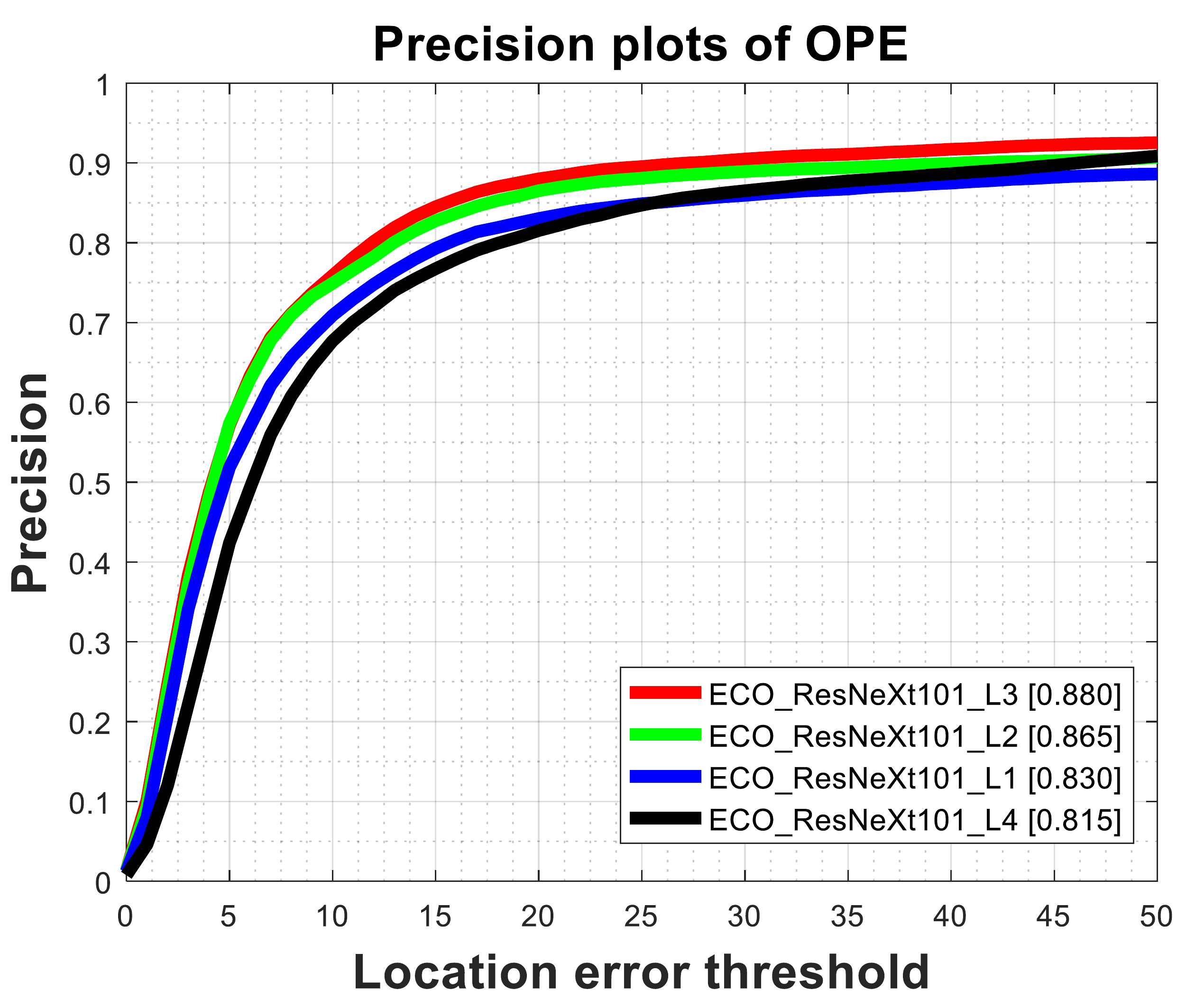}} 
\subfigure{\includegraphics[width=0.24\textwidth]{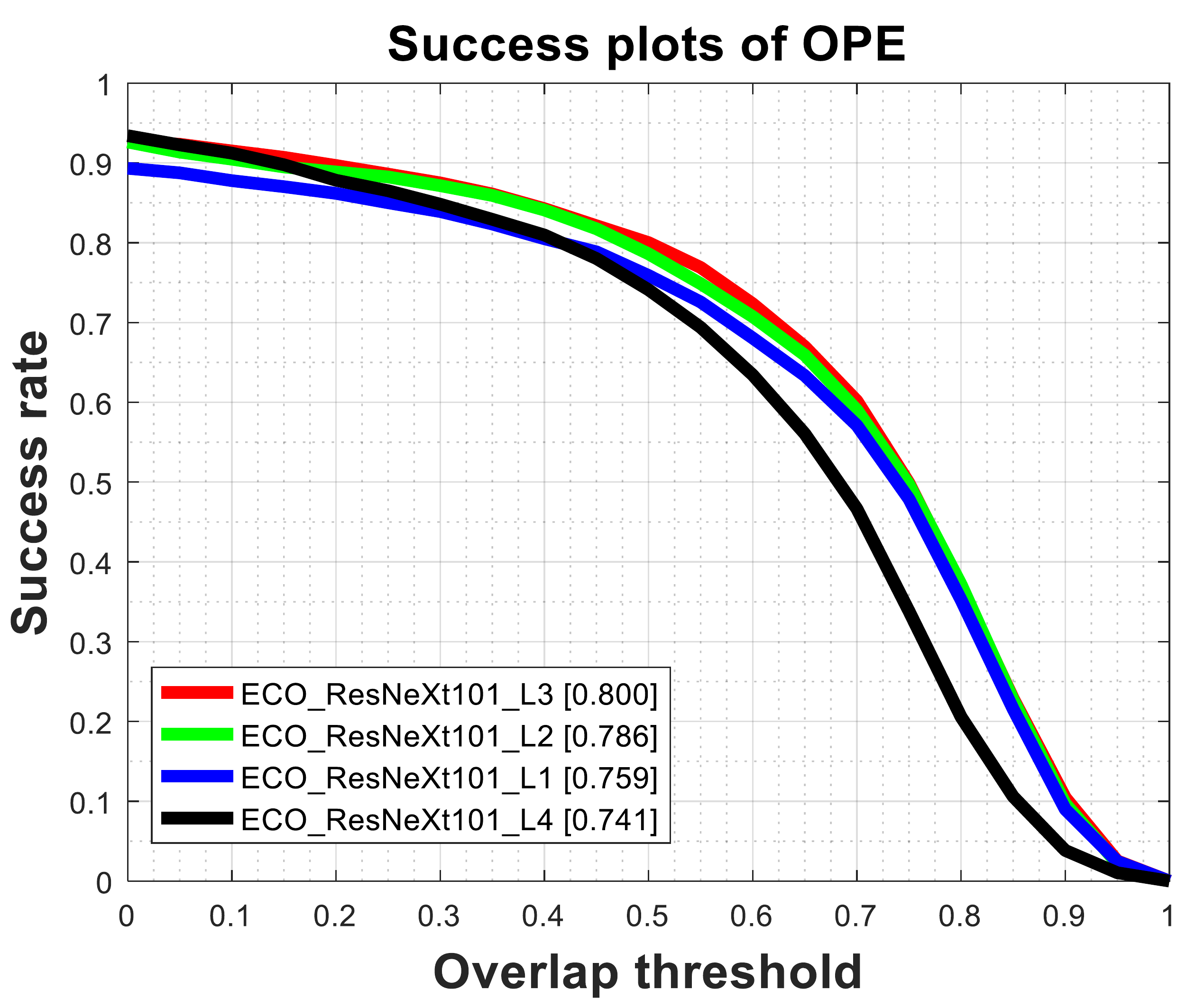}}
\subfigure{\includegraphics[width=0.24\textwidth]{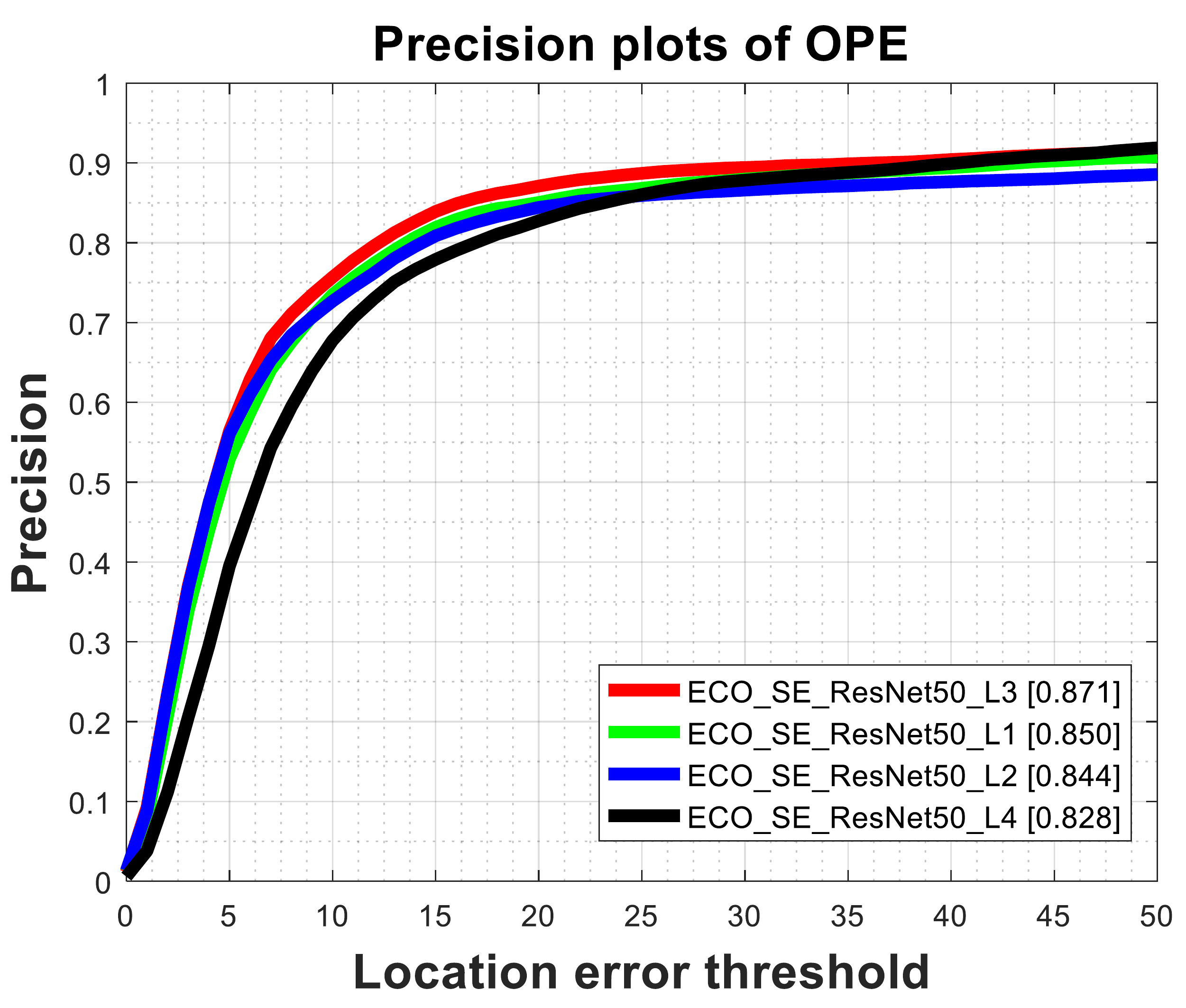}}
\subfigure{\includegraphics[width=0.24\textwidth]{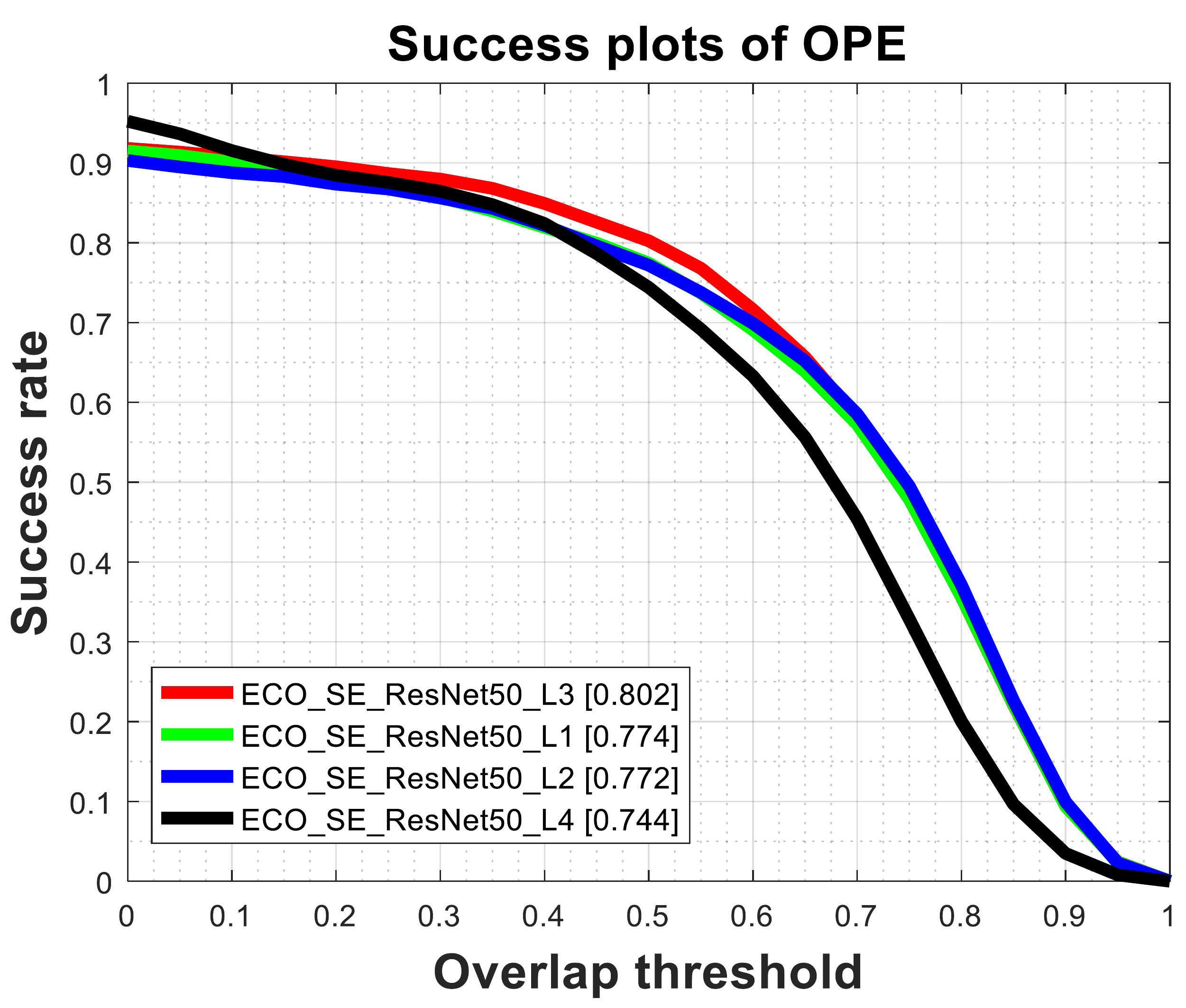}}
\vspace{-6mm}
\justify
\subfigure{\includegraphics[width=0.24\textwidth]{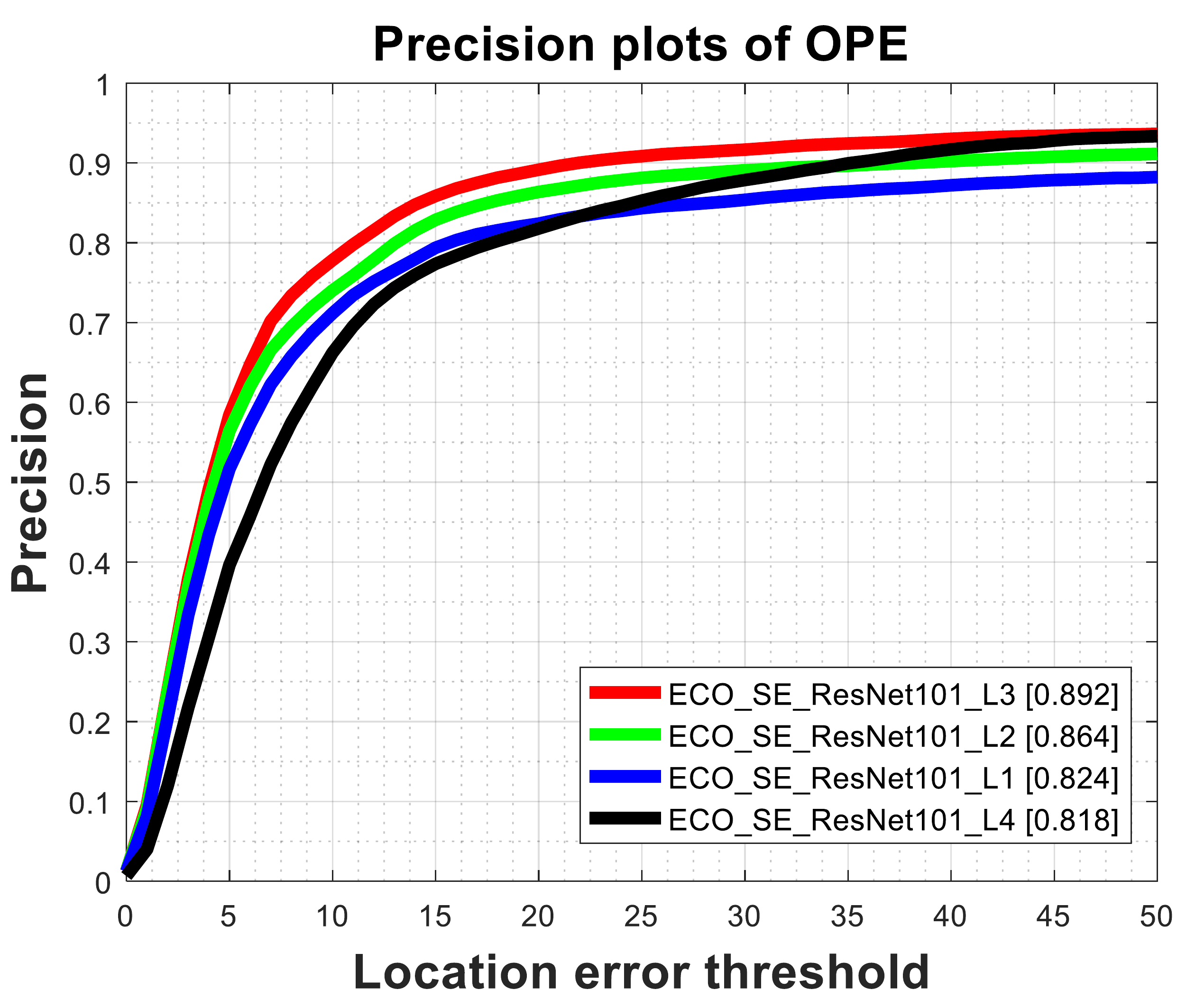}} 
\subfigure{\includegraphics[width=0.24\textwidth]{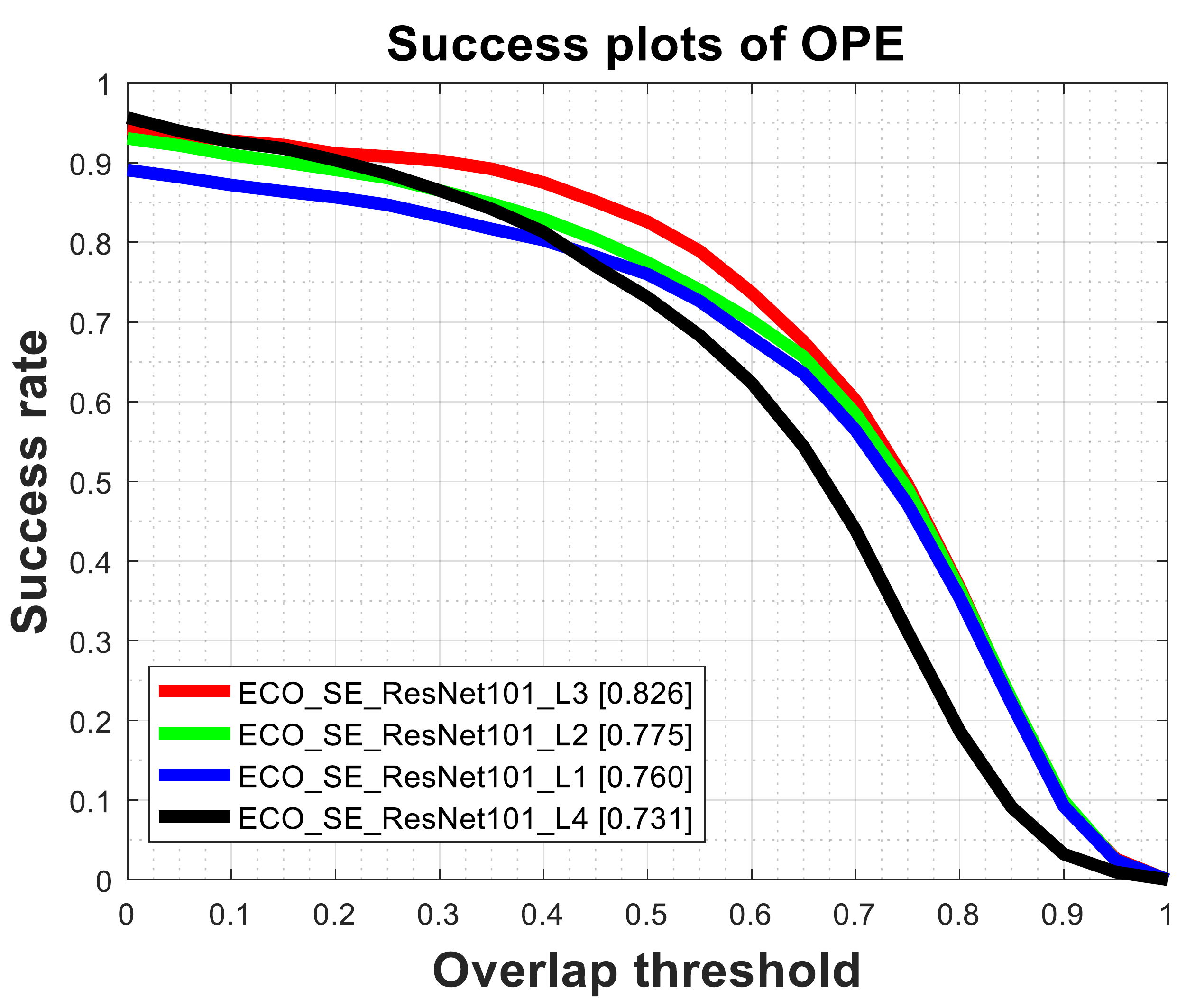}}
\subfigure{\includegraphics[width=0.24\textwidth]{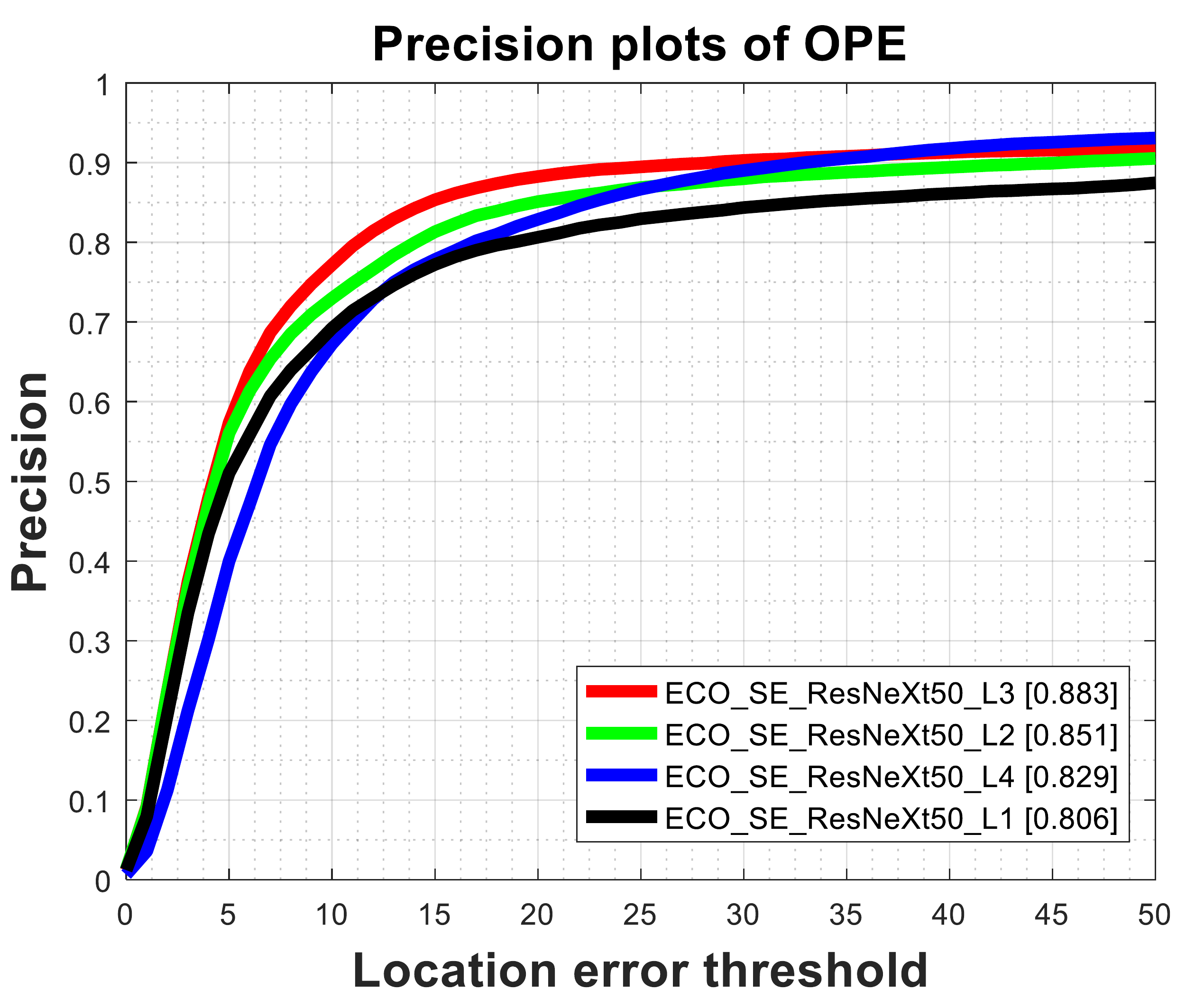}}
\subfigure{\includegraphics[width=0.24\textwidth]{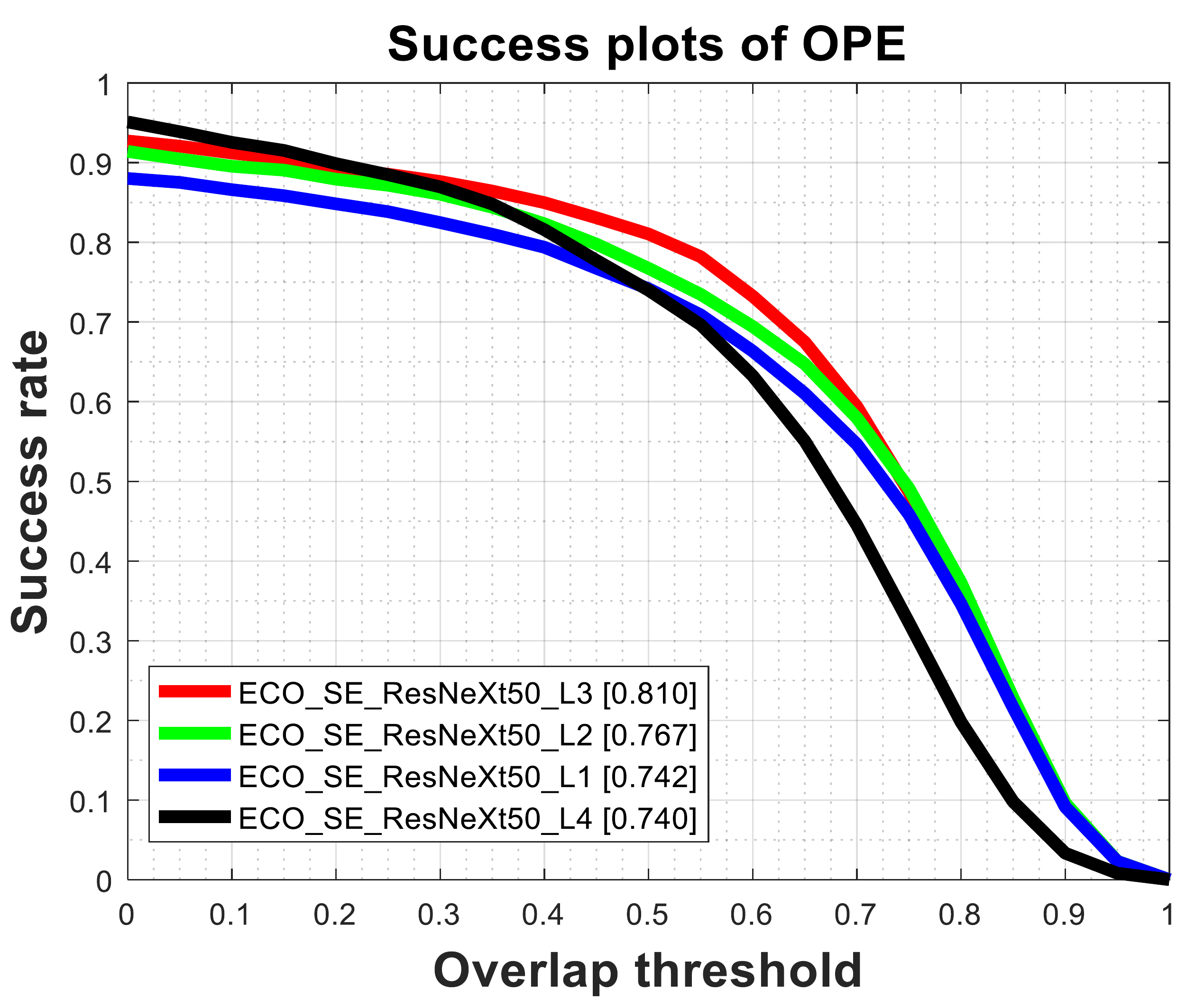}}
\vspace{-6mm}
\justify
\subfigure{\includegraphics[width=0.24\textwidth]{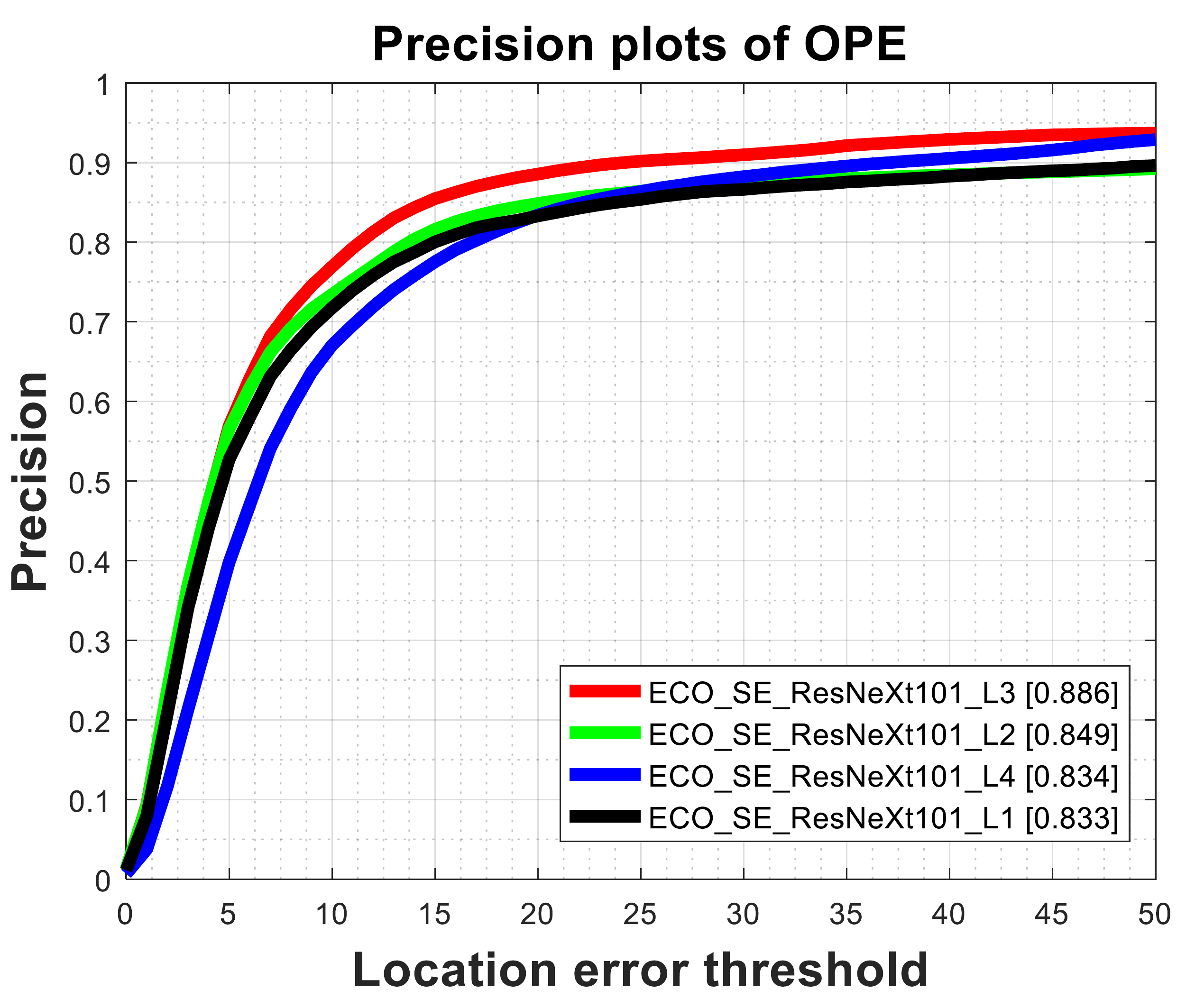}} 
\subfigure{\includegraphics[width=0.24\textwidth]{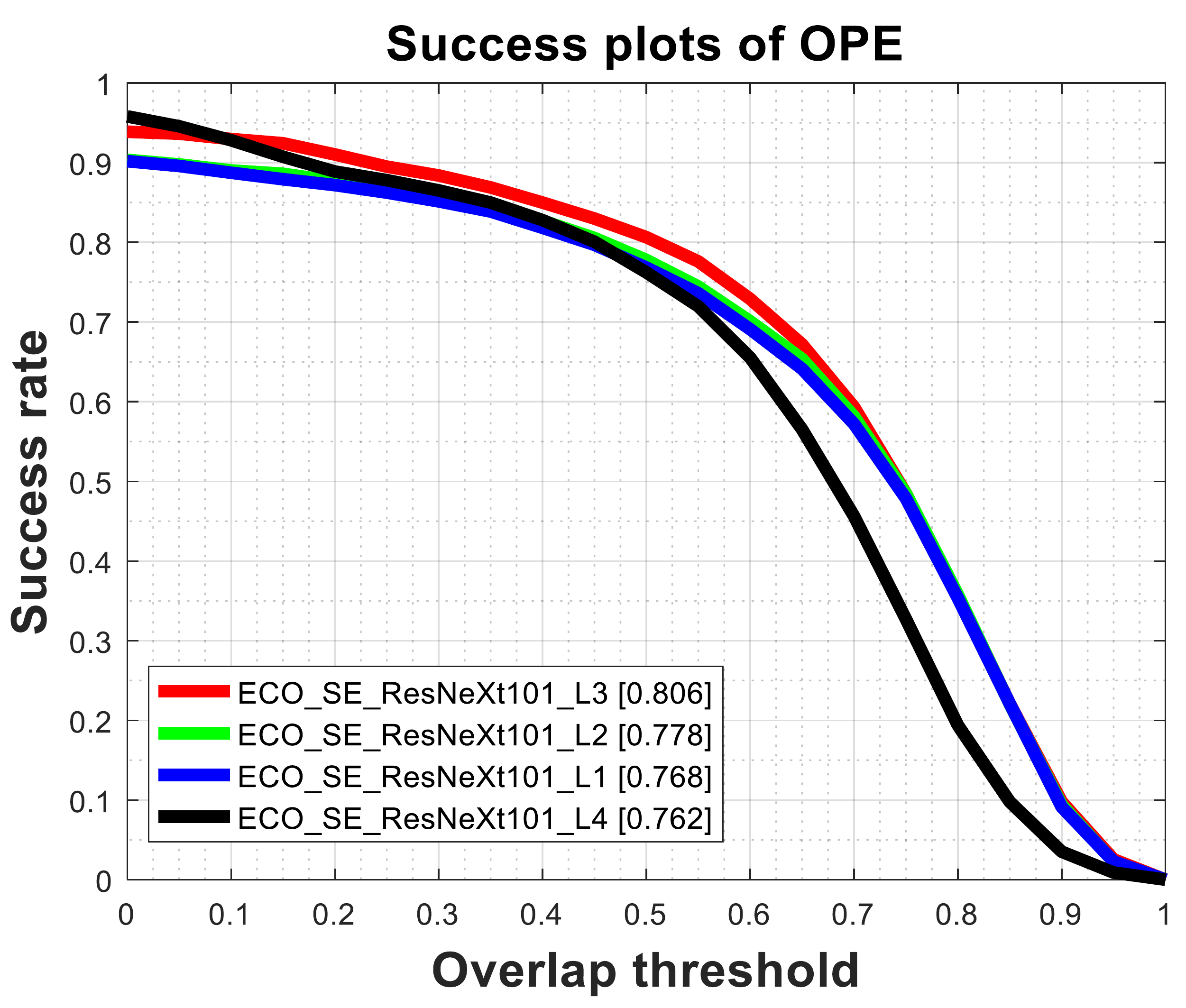}}
\subfigure{\includegraphics[width=0.24\textwidth]{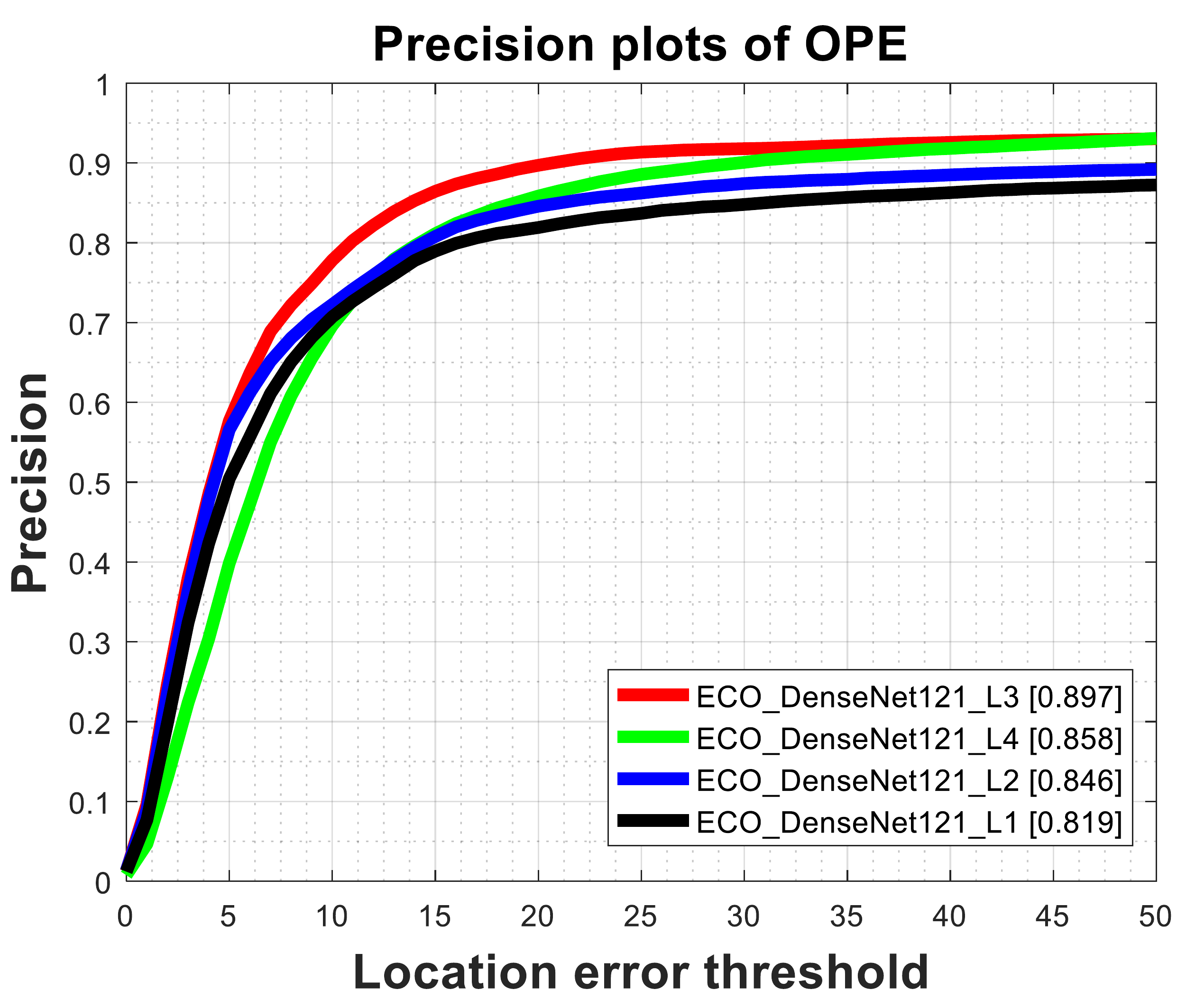}}
\subfigure{\includegraphics[width=0.24\textwidth]{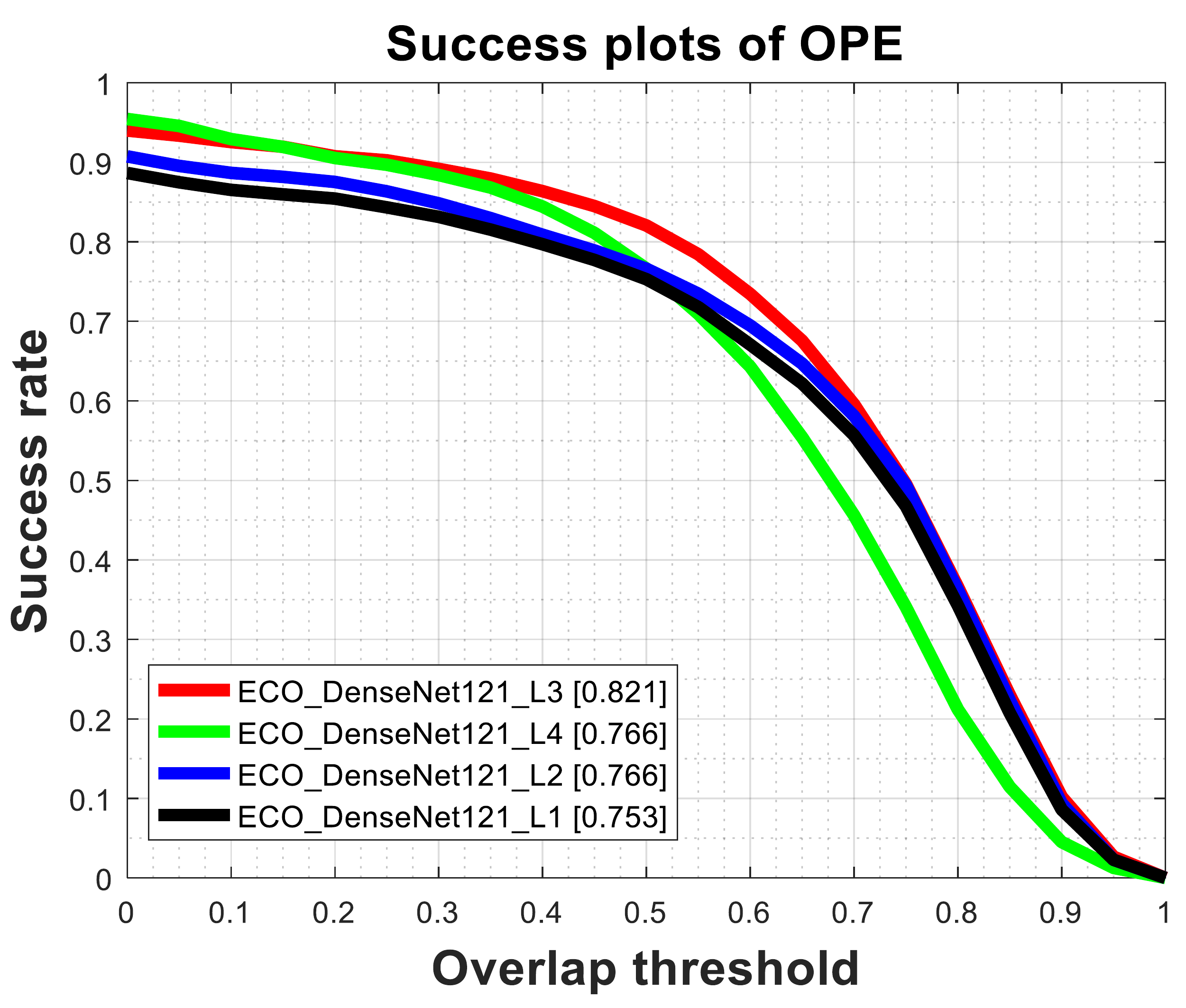}}
\vspace{-6mm}
\justify
\subfigure{\includegraphics[width=0.24\textwidth]{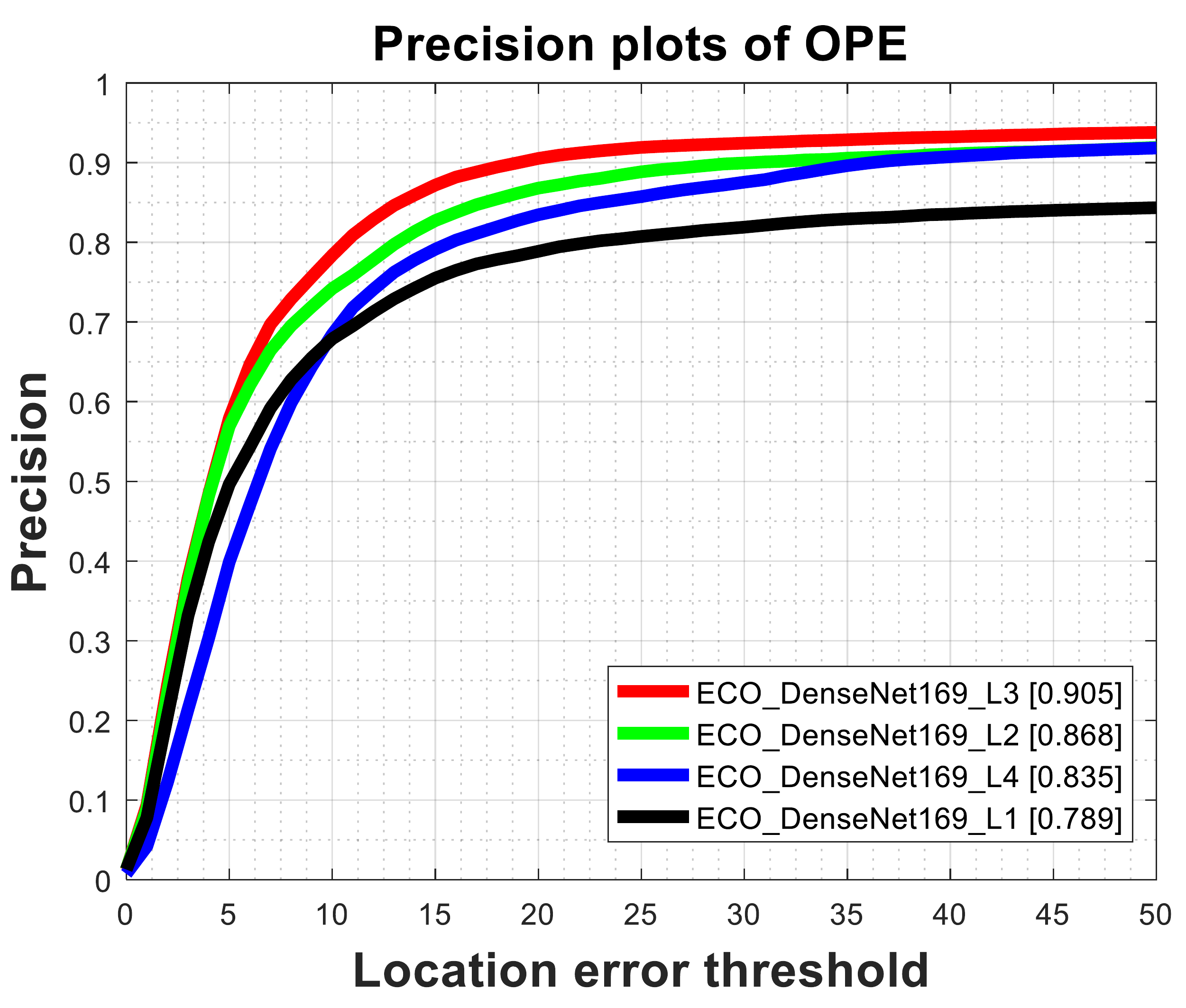}} 
\subfigure{\includegraphics[width=0.24\textwidth]{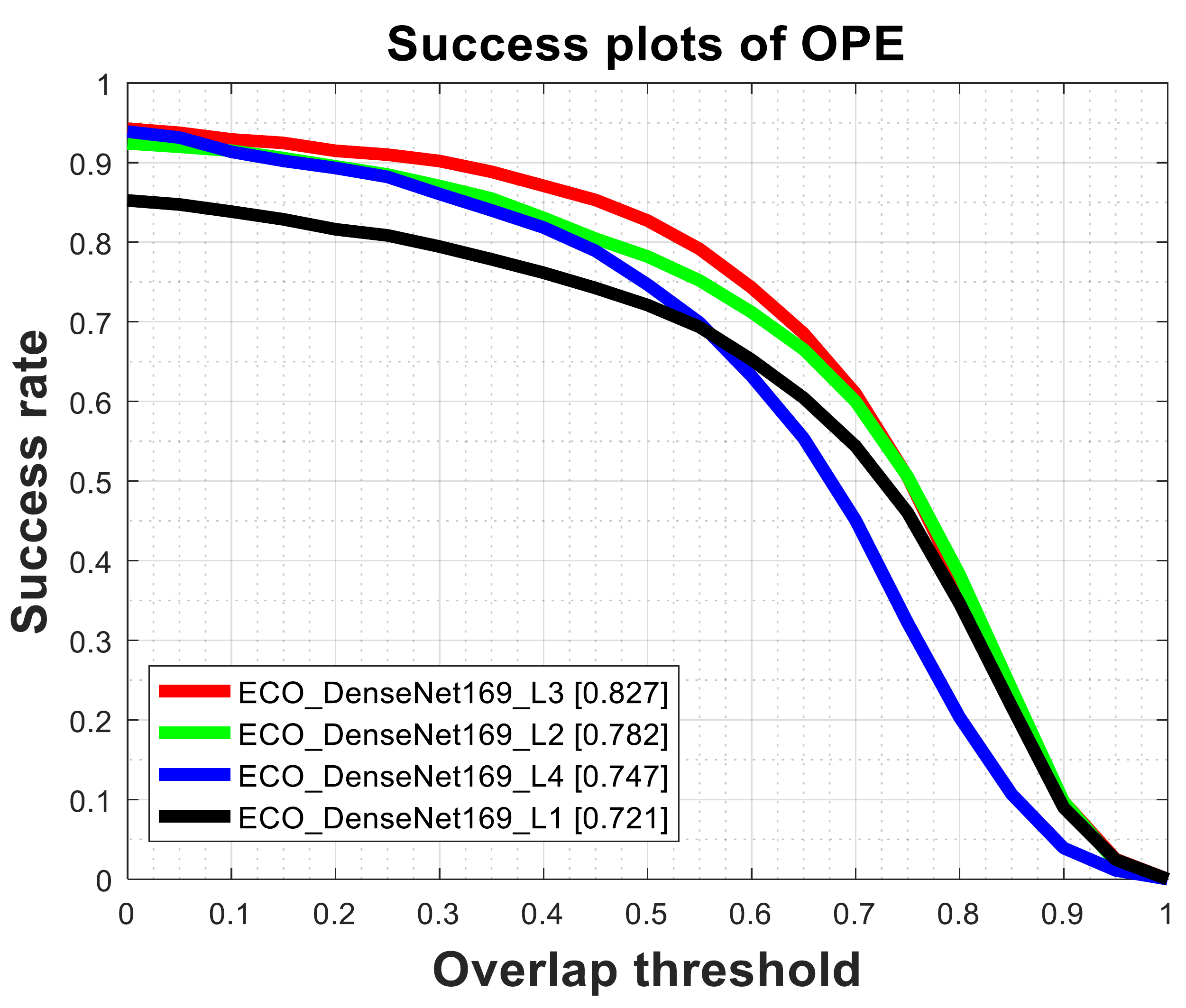}}
\subfigure{\includegraphics[width=0.24\textwidth]{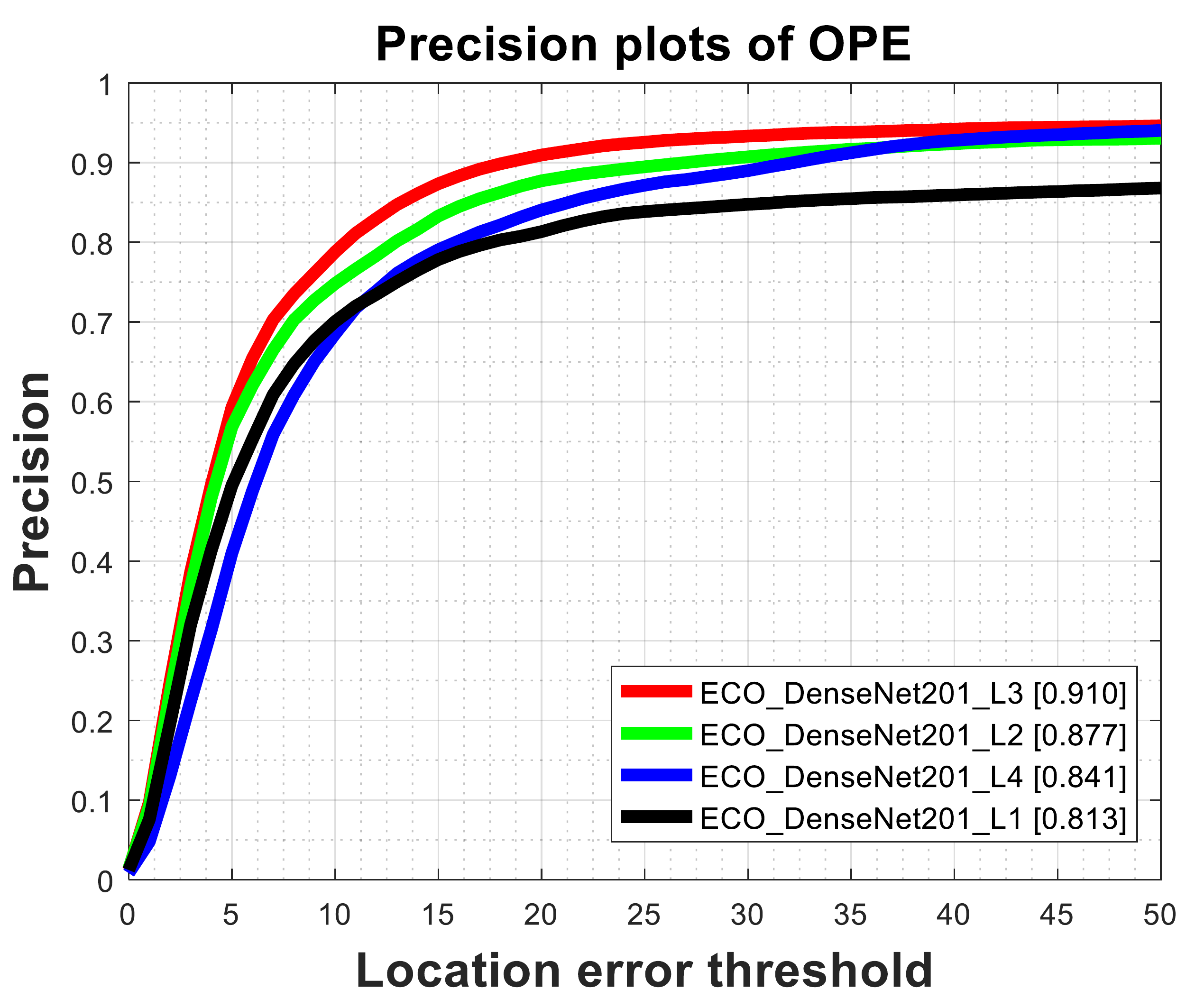}}
\subfigure{\includegraphics[width=0.24\textwidth]{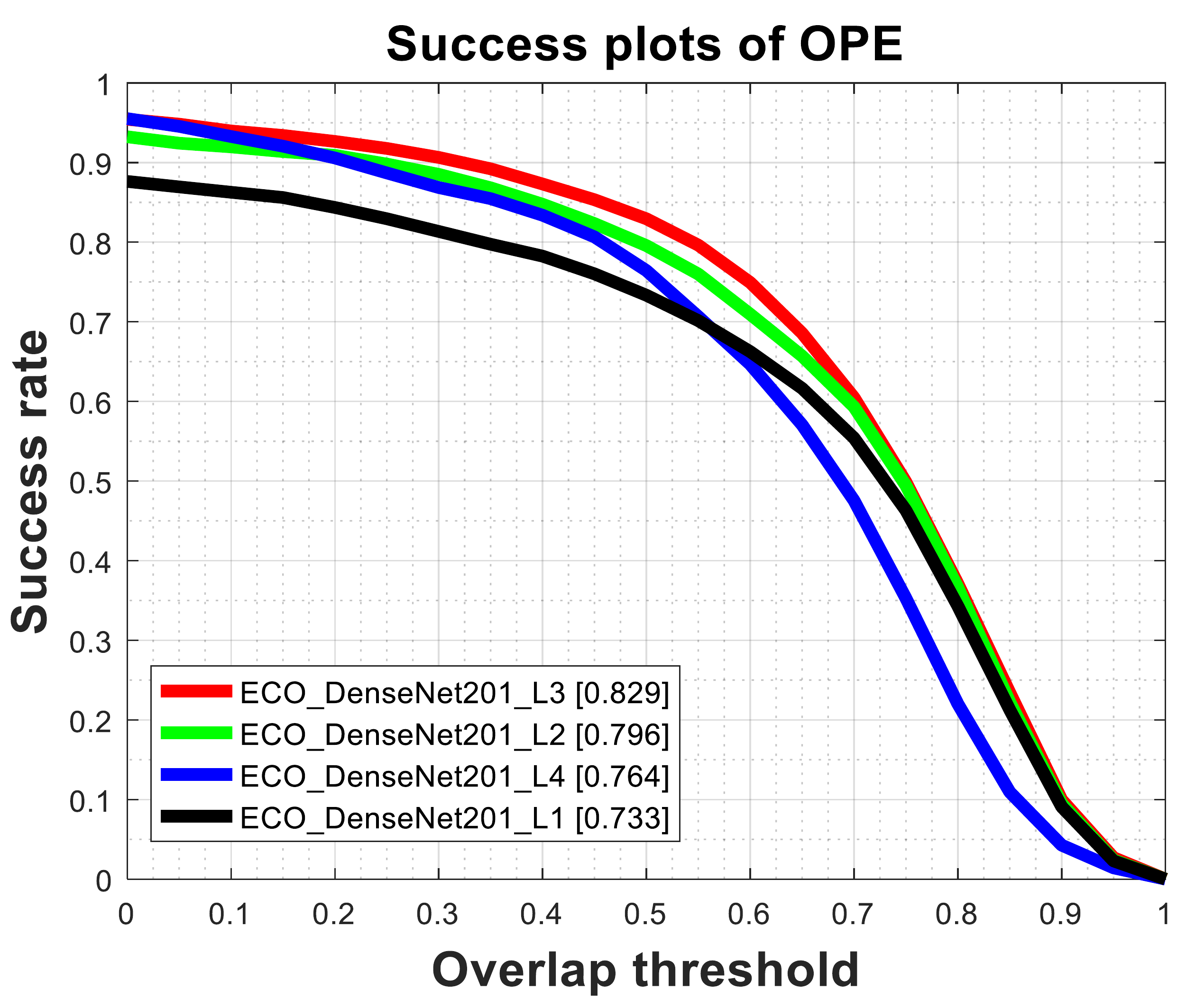}}
\vspace{-2mm}
\caption{Precision and success evaluation results of state-of-the-art ResNet-based FENs on the OTB-2013 dataset.}
\end{figure*}
%%%%%%%%%%%%%%%%%%%%%%%%%%%%%%%%%%%%%%%%%%%%%%%%%%%%%%%%%%%%%%%%%%%%%%%%%%%%%%%%%%%%%%%%%%%%%%%%%%%%%%%%
%%%%%%%%%%%%%%%%%%%%%%%%%%%%%%%%%%%%%%%%%%%%%%%%%%%%%%%%%%%%%%%%%%%%%%%%%%%%%%%%%%%%%%%%%%%%%%%%%%%%%%%
\begin{table*}\label{tab4}
\caption{Performance comparison of the L3 representations of ResNet-based models [\textcolor{green}{first}, \textcolor{blue}{second}, and \textcolor{red}{third} FENs are shown in color].} % title of Table
\vspace{-2mm}
\centering
\tiny
%\resizebox{\textwidth}{!}{
\begin{tabular*}{\textwidth}{c @{\extracolsep{\fill}} ccc}
\hline\hline
Name of FENs &	Precision &	Success &	Average \\
\hline
ResNet-50      &	0.881 &	0.817 &	0.849 \\
ResNet-101     &	\textcolor{red}{0.901} &	\textcolor{green}{0.829} &	\textcolor{red}{0.865} \\
ResNet-152     &	0.844 &	0.777 &	0.810 \\
ResNeXt-50     &	0.885 &	0.819 &	0.852 \\
ResNeXt-101    &	0.880 &	0.800 &	0.840 \\
SE-ResNet-50   &	0.871 &	0.802 &	0.836 \\
SE-ResNet-101  &	0.892 &	\textcolor{red}{0.826} &	0.859 \\
SE-ResNeXt-50  &	0.883 &	0.810 &	0.846 \\
SE-ResNeXt-101 &	0.886 &	0.806 &	0.846 \\
DenseNet-121   &	0.897 &	0.821 &	0.859 \\
DenseNet-169   &	\textcolor{blue}{0.905} &	\textcolor{blue}{0.827} &	\textcolor{blue}{0.866} \\
DenseNet-201   &	\textcolor{green}{0.910} &	\textcolor{green}{0.829} &	\textcolor{green}{0.869} \\
\hline
\end{tabular*}
%}
\vspace{-4mm}
\end{table*}
%%%%%%%%%%%%%%%%%%%%%%%%%%%%%%%%%%%%%%%%%%%%%%%%%%%%%%%%%%%%%%%%%%%%%%%%%%%%%%%%%%%%%%%%%%%%%%%%%%%%%%%%
Furthermore, the comprehensive results show that the DenseNet-201 network provides superior deep representations of target appearance in challenging scenarios compared to the other ResNet-based FENs. This network features reuse property, which concatenates the feature maps learned by different layers to strengthen feature propagation and improve the efficiency. While this network is selected as the FEN in the proposed method (see Sec.~\ref{sec:4}), the exploitation of its best feature maps will be investigated in the next subsection.
\vspace{-8mm}
\subsection{Selection and Generalization Evaluation of Best Feature Maps}
\label{sec:3.2}
\vspace{-2mm}
Two aims are investigated in the following section; exploiting the best feature maps of the DenseNet-201 model and performing generalization evaluation of this network in other DCF-based visual trackers.

In the first step, all single and fused deep feature maps extracted from the DenseNet-201 model were extensively evaluated on the OTB-2013 dataset (see Fig. 2(a)). Based on these results, the feature maps extracted from the L3 layer appeared to have achieved the best visual tracking performance in terms of average precision and success metrics. Although the original ECO tracker exploits fused deep features extracted from the first and third convolutional layers of the VGG-M network, the fusion of different levels of feature maps from the DenseNet network did not lead to better visual tracking performance. These results may help the visual trackers to prevent redundant feature maps and considerably reduce the computational complexity. 

In the second step, the generalization of exploiting the third convolutional layer of the DenseNet-201 model was evaluated on the DeepSTRCF tracker \cite{STRCF}. To ensure a fair comparison, the visual tracking performance of both the original and the proposed versions of DeepSTRCF were evaluated with the aid of only deep features (without fusing with handcrafted features) extracted from VGG-M and DenseNet201 models on the OTB-2013 dataset. As shown in Fig. 2(b), the exploitation of the DenseNet-201 network does not only significantly improve the accuracy and the robustness of visual tracking in challenging scenarios; it can also generalize well into other DCF-based visual trackers.
%%%%%%%%%%%%%%%%%%%%%%%%%%%%%%%%%%%%%%%%%%%%%%%%%%%%%%%%%%%%%%%%%%%%%%%%%%%%%%%%%%%%%%%%%%%%%%%%%%%%%%%
\begin{figure*}
\justify
\subfigure{\includegraphics[width=11.5cm, height=4.2cm]{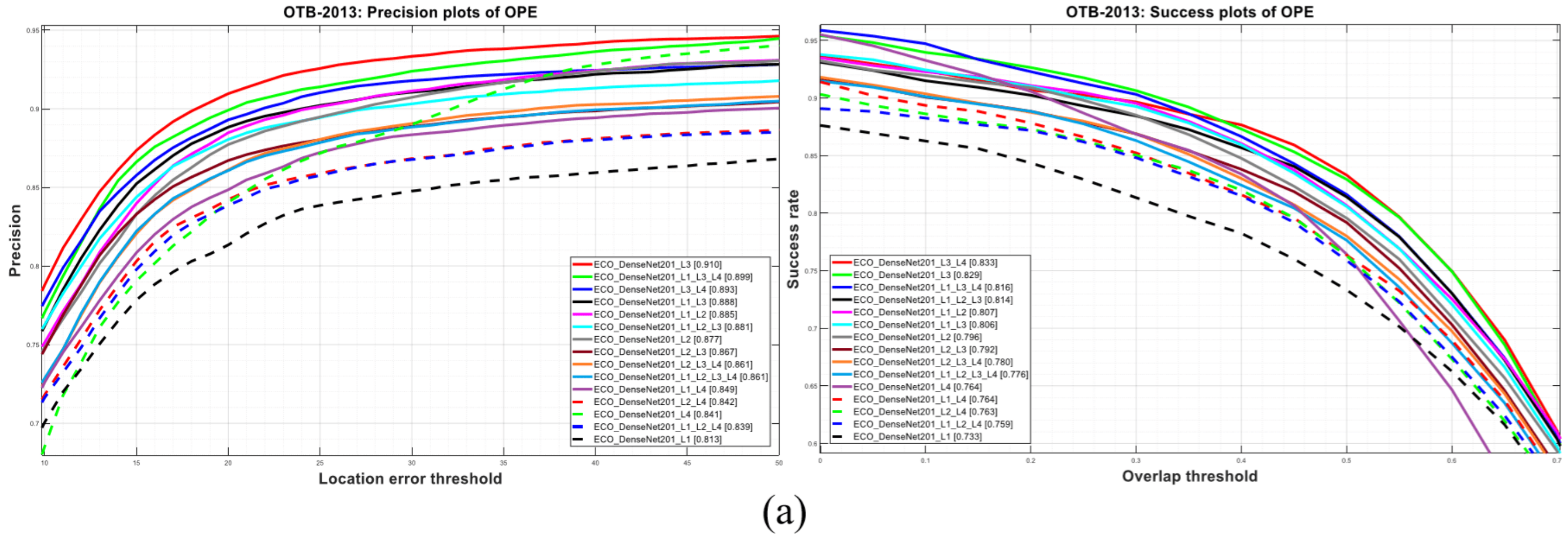}} 
\vspace{-6mm}
\centering
\subfigure{\includegraphics[width=11cm, height=5cm]{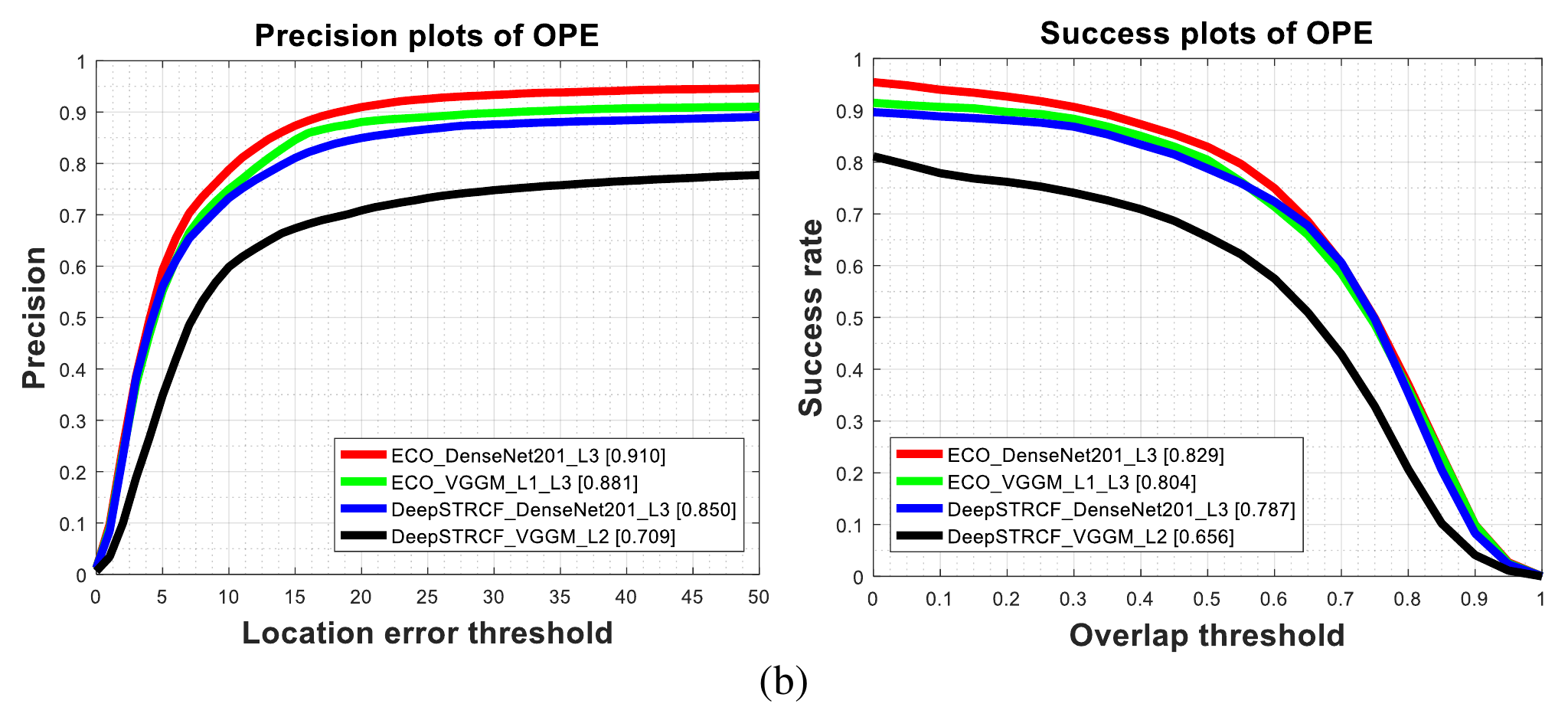}} 
\vspace{2mm}
\caption{Evaluation of exploiting fused feature maps of the pre-trained DenseNet-201 network and generalization of the best of them into the DeepSTRCF tracker.}
\end{figure*}
%%%%%%%%%%%%%%%%%%%%%%%%%%%%%%%%%%%%%%%%%%%%%%%%%%%%%%%%%%%%%%%%%%%%%%%%%%%%%%%%%%%%%%%%%%%%%%%%%%%%%%%
%%%%%%%%%%%%%%%%%%%%%%%%%%%%%%%%%%%%%%%%%%%%%%%%%%%%%%%%%%%%%%%%%%%%%%%%%%%%%%%%%%%%%%%%%%%%%%%%%%%%%%%%%%
\vspace{-6mm}
\section{Proposed Visual Tracking Method}
\label{sec:4}
\vspace{-2mm}
The proposed tracking method exploits fused deep features of multi-task FENs and introduces semantic weighting to provide a more robust target representation for the learning process. To effectively track generic objects in challenging scenarios, the proposed method aims to exploit fused deep representation (extracted from FENs) and semantic weighting of target regions. The algorithm 1 is shown a brief outline of the proposed method.
\vspace{-2mm}
\begin{table}[H]
\centering
\begin{tabular}{p{3.45in}}
\hline
%row no:1
\multicolumn{1}{|p{4.5in}|}{{Algorithm\ 1.  Brief Outline of Proposed Method}} \\
\hhline{-}
%row no:2
\multicolumn{1}{|p{4.5in}|}{\begin{itemize}[leftmargin=*] \vspace{-2mm}
	\item {\textbf{Pre-requisitions:}} \par 
	\begin{itemize} 
	\item {Survey of twelve ResNet-based models which have been pre-trained on ImageNet dataset} \par 
	\item {Selection of the best feature maps of the DenseNet-201 model and their generalization} \par 
	\item {Selection of the DenseNet-201 model as the best FEN} \par 
	\item {Selection of the FCN-8s model which has been pre-trained on PASCAL VOC and MSCOCO datasets} \par 	\item {Selection of the pixel-wise prediction maps as the semantic features}
\end{itemize} \par 	
\item {\textbf{Evaluation of proposed visual tracking method on the OTB-2013, OTB-2015, and TC-128 datasets (without any pre-training of translation and scale filters):}}
\end{itemize} \par {Initialization of the proposed method with a given bounding box in the first frame:} \par \begin{enumerate}
	\hangindent=2em 
	\item {Extract deep features from DenseNet-201 and FCN-8s models} \par 	
	\item {Fuse deep multi-tasking features with semantic windowing using Eq. (2) to Eq. (7)} \par 	
	\item {Learn the continuous translation filters using Eq. (9) to robustly model the target } \par 
	\item {Extract HOG features} \par 
	\item {Learn continuous scale filters using a multi-scale search strategy} \par 
	\item {Update the translation and scale filters using Eq. (11) (for frame t > 1)} \par 
	\item {Go to the next frame} \par 
	\item {Select the search region according to the previous target location} \par 
	\item {Extract deep features from DenseNet-201 and FCN-8s models} \par 
	\item {Fuse deep multi-tasking features with semantic windowing using Eq. (2) to Eq. (7)} \par 
	\item {Calculate the confidence map Eq. (9)} \par 
	\item {Estimate the target location by finding the maximum of Eq. (10)} \par 
	\item {Extract multi-scale HOG features centered at the estimated target location} \par 
	\item {Estimate the target scale using the multi-scale search strategy} \par 
	\item {Select the target region conforming with the estimated location and scale} \par 
	\item {Return to Step 1}
	\vspace{-2mm}
\end{enumerate}} \\
\hhline{-}
\end{tabular}
 \end{table}
 \vspace{-4mm}
\vspace{-2mm}
Using the ECO framework, the proposed method minimizes the following objective function in the Fourier domain \cite{ECO}
\begin{equation}\tag{1}
min_{H}~  \sum _{i=1}^{m} \alpha _{i} \Vert  \sum _{j=1}^{n}H^{j} F_{i}^{j} K_{j}-Y_{i} \Vert _{l^{2}}^{2}+ \sum _{j=1}^{n} \Vert P\ast H^{j} \Vert _{l^{2}}^{2}
\end{equation}
in which  \( H \),  \( F \),  \( Y \), and  \( K \) denote the multi-channel convolution filters, weighted fused deep feature maps from FENs, desirable response map, and interpolation function (to pose the learning problem in the continuous domain), respectively. The penalty function (to learn filters on target region), number of training samples, number of convolution filters, and weights of training samples are indicated by the \( P \),  \( m \),  \( n \), and  \(  \alpha _{i} \) variables, respectively. Also, the \( \ast \), and  \( l^{2} \) are referred to the circular convolution operation and L2 norm, respectively. Capital letters represent the discrete Fourier transform (DFT) of variables.

In addition to utilizing the feature maps of the DenseNet-201 network, the proposed method extracts deep feature maps from the FCN-8s network (i.e., 21 feature maps from the \textit{``score\_final''} layer) to learn a more discriminative target representation by fusing the feature maps as
\begin{equation}\tag{2}
\overline{X}= \left[ X_{DenseNet}^{1} , \ldots ,X_{DenseNet}^{S1} , X_{FCN}^{1} , \ldots ,X_{FCN}^{S2} \right] 
\end{equation}
where  \( \overline{X} \),  \( X \),  \( S1 \), and  \( S2 \)  are the fused deep representation, deep feature maps, number of feature maps extracted from the DenseNet-201 network, and number of feature maps extracted from the FCN-8s network, respectively. 

Although the DCF-based visual trackers use a cosine window to weight the target region and its surroundings, this weighting scheme may contaminate the target model and lead to drift problems. Some recent visual trackers, e.g., \cite{WindowedCF}, utilize different windowing methods, such as computation of target likelihood maps from raw pixels. However, the present work is the first method to use deep feature maps to semantically weight target appearances. The proposed method exploits the feature maps of the FCN-8s model to semantically weight the target. Given the location of the target at (\( e,f \))  in the first (or previous) frame, the proposed method defines a semantic mask \( W_{mask} \) as
\begin{equation}\tag{3}
L_{FCN} \left( i,j \right) = \Bigg \{ \begin{matrix} Z  &  \;\;\; if \;\; max X_{FCN}^{Z} \left( i,j \right) >max_{C \neq Z} X_{FCN}^{C} \left( i,j \right) \\
0  &  otherwise\\
\end{matrix}
\end{equation}
\begin{equation}\tag{4}
W_{mask} \left( i,j \right) = \Bigg\{ \begin{matrix}
1  &  \;\;\; if \;\; L_{FCN} \left( i,j \right) =L_{FCN} \left( e,f \right)  \\
0  &  otherwise\\
\end{matrix}
\end{equation}
in which  \( L_{FCN} \),  \( Z \), and  \(  \left( i,j \right)  \)  are the label map, PASCAL classes  \( \{ C, Z \}   \epsilon   \left( 1, \ldots ,21 \right)  \), and spatial location, respectively. Finally, the fused feature maps are weighted by
\begin{equation}\tag{5}
F=\overline{X}\odot \left( W_{mask}~.  W_{cos} \right)
\end{equation}
where  \( W_{cos} \)  and  \( \odot \) are the conventional cosine (or sine) window and the channel-wise product, respectively. The traditional cosine window weights a deep representation \( X \) with \( d1 \) and \( d2 \) dimensions as
\begin{equation}\tag{6}
X_{ij}^{w}= \left ( X_{ij}-0.5 \right )\sin \left ( \pi i/d1 \right )\sin \left ( \pi j/d2 \right ),  \forall i= 0,...,d1 - 1, j= 0,...,d2 - 1 \;\;\; .
\end{equation}
The matrix form of semantically weighted fused deep representations is defined as
\begin{equation}\tag{7}
B \triangleq FK= \left[ \overline{X}\odot \left( W_{mask}~.  W_{cos} \right)  \right]  .K
\end{equation}
such that the weighted fused feature maps are posed to the continuous domain by an interpolation function. By defining the diagonal matrix of training samples weights as  \(  \Gamma  \), the regularized least square problem (1) is minimized by 
\begin{equation}\tag{8}
\nabla _{H}  \Vert B  \Gamma  H-Y \Vert _{2}^{2}+ \Vert P H \Vert _{2}^{2}=0
\end{equation}
which has the following closed-form solution
\begin{equation}\tag{9}
 \left[ B^{T}  \Gamma  B+P^{T} P \right]  H=B^{T}  \Gamma  Y \;\;\; .
\end{equation}

The proposed method learns multi-channel convolution filters by employing the iterative conjugate gradient (CG) method. Because the iterative CG method does not need to form the multiplications of  \( B^{T} \)  and  \( B \), it can desirably minimize the objective function in limited iterations. To track the target in subsequent frames, the multitasking deep feature maps (\( T \)) are extracted from the search region in the current frame. Next, the confidence map is calculated by
\begin{equation}\tag{10}
CM=F^{-1} \bigg\{  \sum _{j=1}^{n}T^{j} .  \left( H^{j} F^{j} K_{j} \right)  \bigg \} 
\end{equation}
that its maximum determines the target location in the current frame. Then, a new set of continuous convolution filters are trained on the current target sample and the prior learned filters are updated by the current learned filters as
\begin{equation}\tag{11}
H= \left( 1- \gamma  \right)  H_{prior}+ \gamma  H_{current}
\end{equation}
in which  \(  \gamma  \)  is learning rate. Therefore, the proposed method alternatively uses online training and tracking processes of the generic visual target for all video frames. 
%%%%%%%%%%%%%%%%%%%%%%%%%%%%%%%%%%%%%%%%%%%%%%%%%%%%%%%%%%%%%%%%%%%%%%%%%%%%%%%%%%%%%%%%%%%%%%%%%%%%%%%%%%
\vspace{-4mm}
\section {Experimental Results}
\label{sec:5}
\vspace{-2mm}
All implementation details related to the exploited models (including the ResNet, ResNeXt, SE-ResNet, SE-ResNeXt, DenseNet, and FCN-8s models) are the same as in their original papers \cite{ResNet,ResNeXt,SE-Res-CVPR,SE-Res-PAMI,DenseNet,FCN-8s}. The ResNet-based models have been trained on the ImageNet dataset \cite{ImageNet} while the FCN-8s model has been trained on the PASCAL VOC and MSCOCO data-sets \cite{PascalVOC,MSCOCO}. Similar to the exploitation of FENs in state-of-the-art visual tracking methods \cite{DeepSRDCF,DeepMotionFeatures,STRCF,CCOT,ECO,ETDL,FCNT,HCFT,HCFTs,DNT,HDT,LCTdeep,CF-CNN}, these CNNs are not fine-tuned or trained during their performance evaluations on the OTB-2013, OTB-2015, TC-128, and VOT-2018 visual tracking datasets. In fact, the FENs transfers the power of deep features to the visual tracking methods. According to the visual tracking benchmarks, all visual trackers are initialized with only a given bounding box in the first frame of each video sequence. Then, these visual trackers start to learn and track a target, simultaneously. To investigate the effectiveness of the proposed method and fair evaluations, the experimental configuration of baseline ECO tracker is used. The search region, learning rate, number of CG iterations, and number of training samples are set to the \( 5^2 \) times of the target box, 0.009, 5, and 50, respectively. The desirable response map is considered as a 2D Gaussian function which its standard deviation is set to 0.083. Given a target with size \( q1 \times q2 \) and spatial location  \((i,j)\), the penalty function is defined as 
\begin{equation}\tag{12}
p \left( i,j \right) =0.1+3 \left( \frac{i}{q1} \right) ^{2}+3 \left( \frac{j}{q2} \right) ^{2}
\end{equation} 
which is a quadratic function with a minimum at its center. Moreover, the number of deep feature maps extracted from the DenseNet and FCN-8s models are  \( S1=512 \) and  \( S2=21 \). To estimate the target scale, the proposed method utilizes the multi-scale search strategy \cite{DSST} with 17 scales which have relative scale factor 1.02. Note that this multi-scale search strategy exploits the HOG features to accelerate the visual tracking methods (such as in the DSST, C-COT, and ECO). Like the baseline ECO tracker, the proposed method applies the same configuration to all videos in different visual tracking datasets. However, this work exploits only deep features to model target appearance and highlight the effectiveness of the proposed method. This section includes a comparison of the proposed method with state-of-the-art visual tracking methods and the ablation study, which evaluates the effectiveness of proposed components on visual tracking performance.
\vspace{-4mm}
\subsection{Performance Comparison}
\label{sec:5.1}
To evaluate the performance of the proposed method, it is extensively compared quantitatively with deep and handcrafted-based state-of-the-art visual trackers (for which benchmark results on different datasets have been publicly available) in the one-pass evaluation (OPE) on three large visual tracking datasets: OTB-2013 \cite{OTB2013}, OTB-2015 \cite{OTB2015},  TC-128 \cite{TC128}, and VOT-2018 \cite{VOT-2018} datasets. The state-of-the-art deep visual trackers include deep feature-based visual trackers (namely, ECO-CNN \cite{ECO}, DeepSRDCF \cite{DeepSRDCF}, UCT \cite{UCT}, DCFNet \cite{DCFNet}, DCFNet-2.0 \cite{DCFNet}, LCTdeep \cite{LCTdeep}, HCFT \cite{HCFT}, HCFTs \cite{HCFTs}, SiamFC-3s \cite{SiamFC}, CNN-SVM \cite{CNN-SVM}, PTAV \cite{PTAV-ICCV,PTAV}), CCOT \cite{CCOT}, DSiam \cite{DSiam}, CFNet \cite{CFNet}, DeepCSRDCF \cite{DeepCSRDCF}, MCPF \cite{MCPF}, TRACA \cite{TRACA}, DeepSTRCF \cite{STRCF}, SiamRPN \cite{SiamRPN}, SA-Siam \cite{SA-Siam}, LSART \cite{LSART}, DRT \cite{DRT}, DAT \cite{DAT}, CFCF \cite{CFCF}, CRPN \cite{CRPN}, GCT \cite{GCT}, SiamDW-SiamFC \cite{SiamDW}, and SiamDW-SiamRPN \cite{SiamDW}; Also the compared handcrafted feature-based visual trackers are the Struck \cite{StruckTPAMI}, BACF \cite{BACF}, DSST \cite{DSST}, MUSTer \cite{MUSTer}, SRDCF \cite{SRDCF}, SRDCFdecon \cite{SRDCFdecon}, Staple \cite{Staple}, LCT \cite{LCTdeep}, KCF \cite{KCF-ECCV,KCF}, SAMF \cite{SAMF}, MEEM \cite{MEEM}, and TLD \cite{TLD}). These methods are included the visual trackers that exploit handcrafted features, deep features (by FENs or EENs), or both. The proposed method is implemented on an Intel I7-6800K 3.40 GHz CPU with 64 GB RAM with an NVIDIA GeForce GTX 1080 GPU. The overall and attribute-based performance comparisons of the proposed method with different visual trackers on the four visual tracking datasets in terms of various evaluation metrics are shown in Fig. 3 to Fig. 5. 
%%%%%%%%%%%%%%%%%%%%%%%%%%%%%%%%%%%%%%%%%%%%%%%%%%%%%%%%%%%%%%%%%%%%%%%%%%%%%%%%%%%%%%%%%%%%%%%%%%%%%%%
\begin{figure*}
\justify
\subfigure{\includegraphics[width=5.9cm, height=3.8cm]{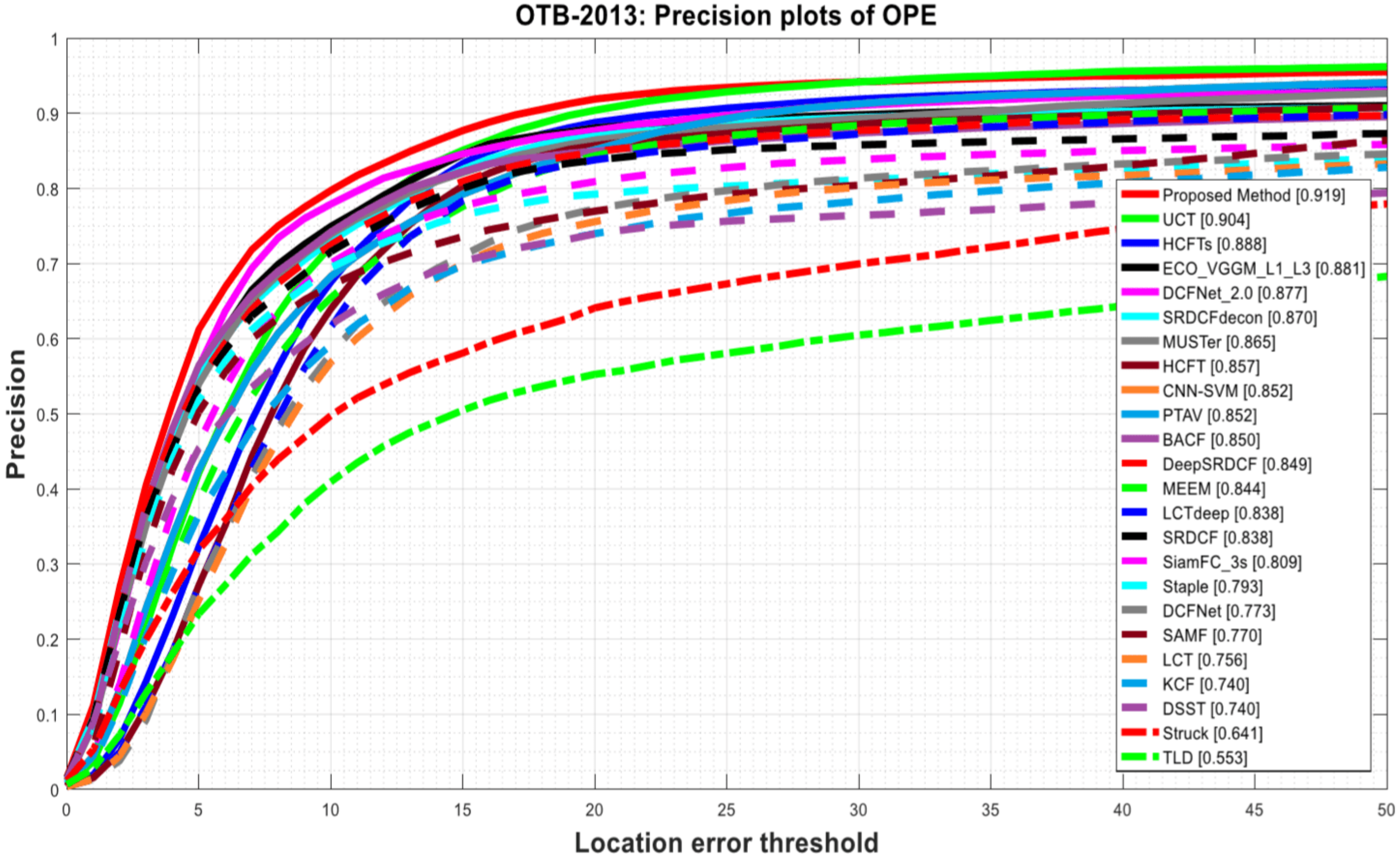}} 
\hspace{0mm}
\subfigure{\includegraphics[width=5.9cm, height=3.8cm]{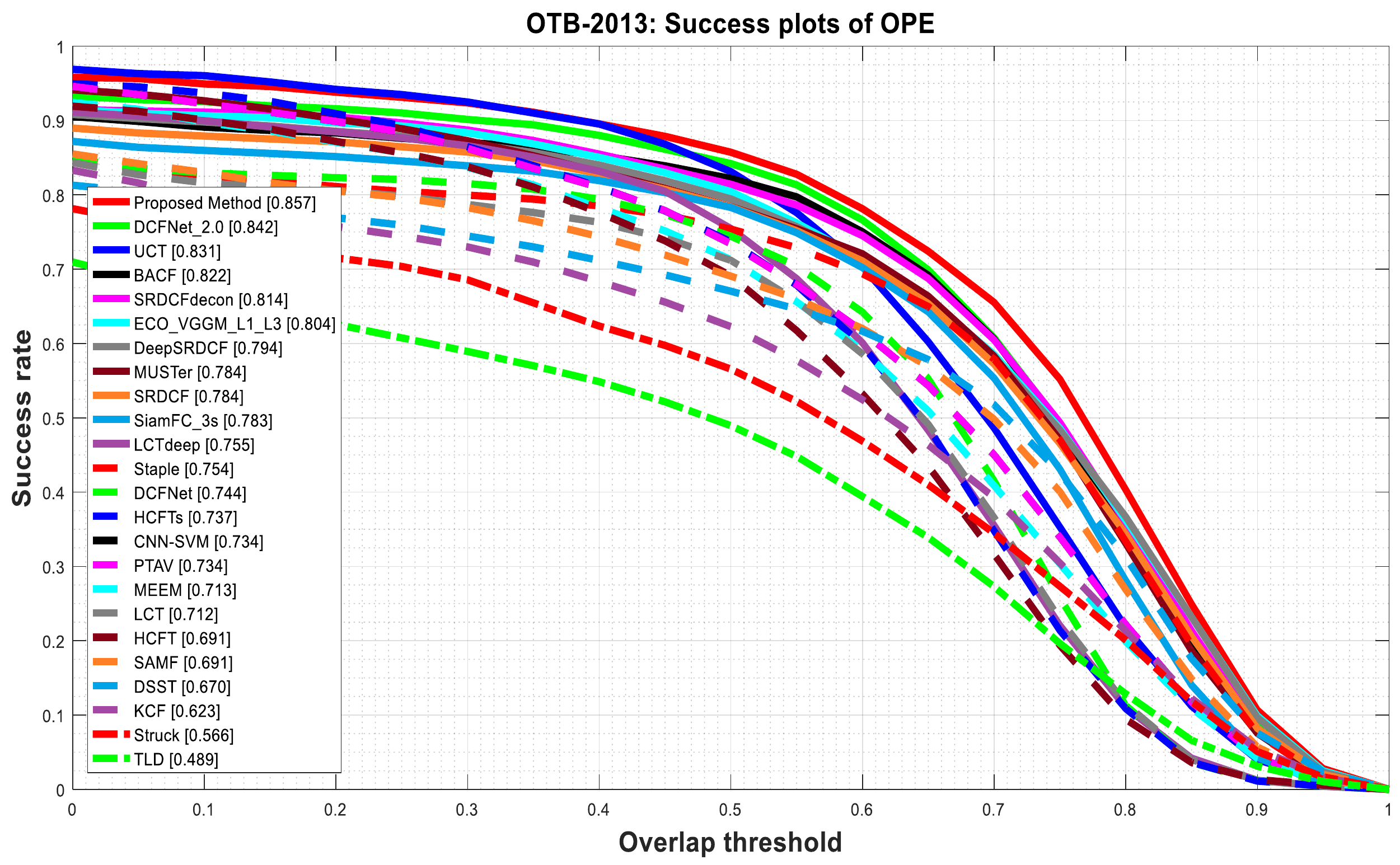}}
\vspace{-8mm}
\justify
\subfigure{\includegraphics[width=5.9cm, height=3.8cm]{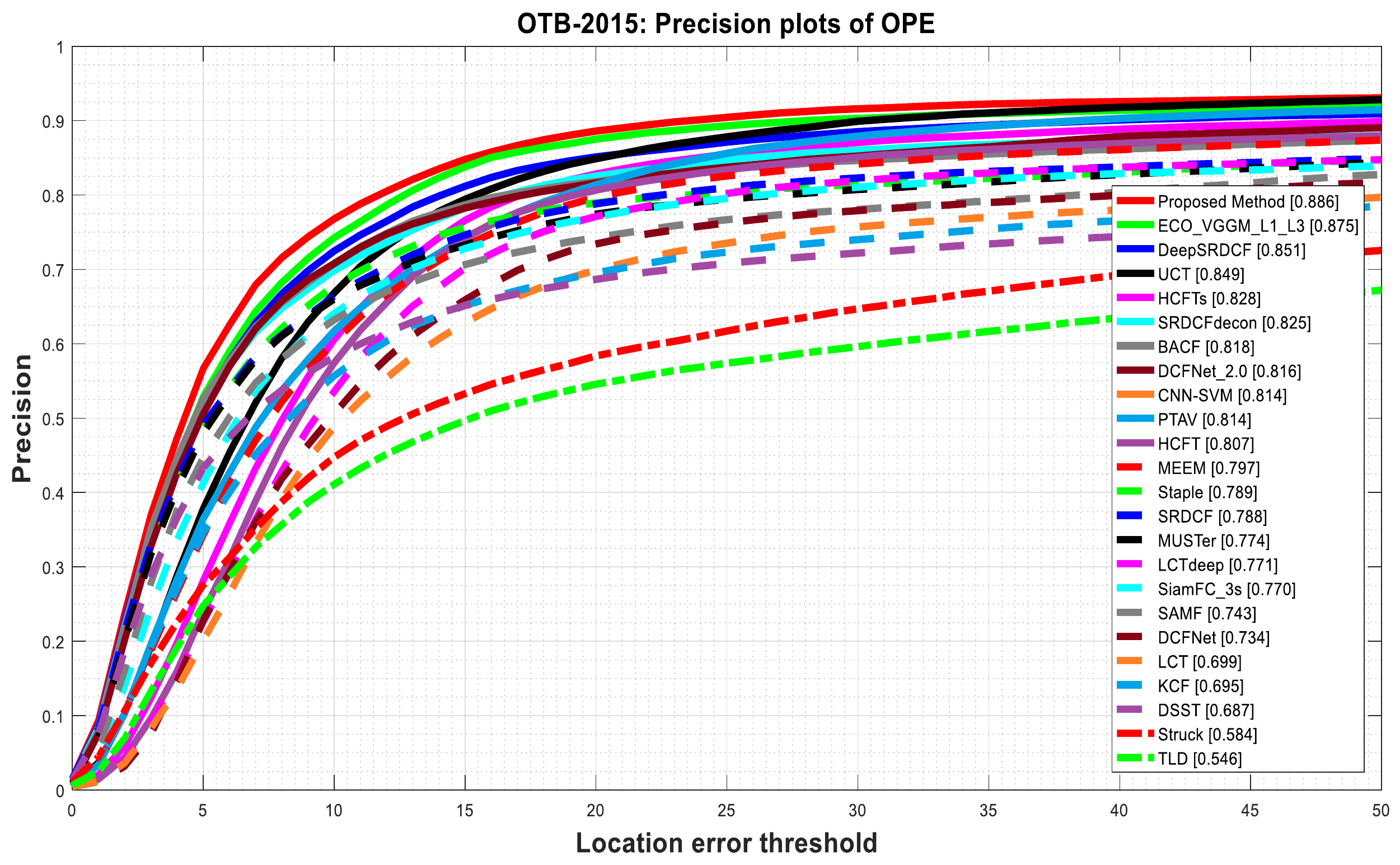}} 
\hspace{0mm}
\subfigure{\includegraphics[width=5.9cm, height=3.8cm]{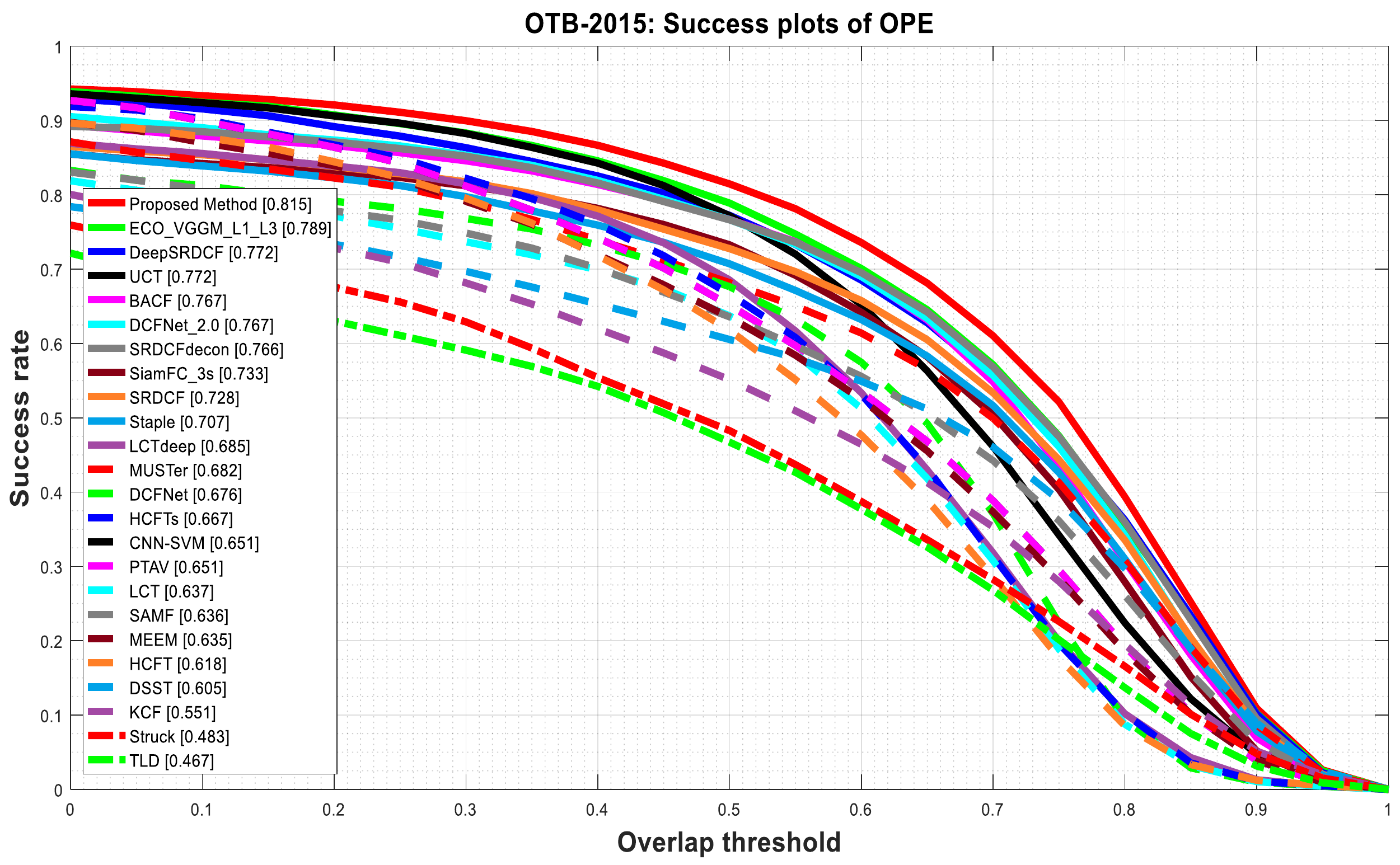}}
\vspace{-8mm}
\justify
\subfigure{\includegraphics[width=5.9cm, height=3.8cm]{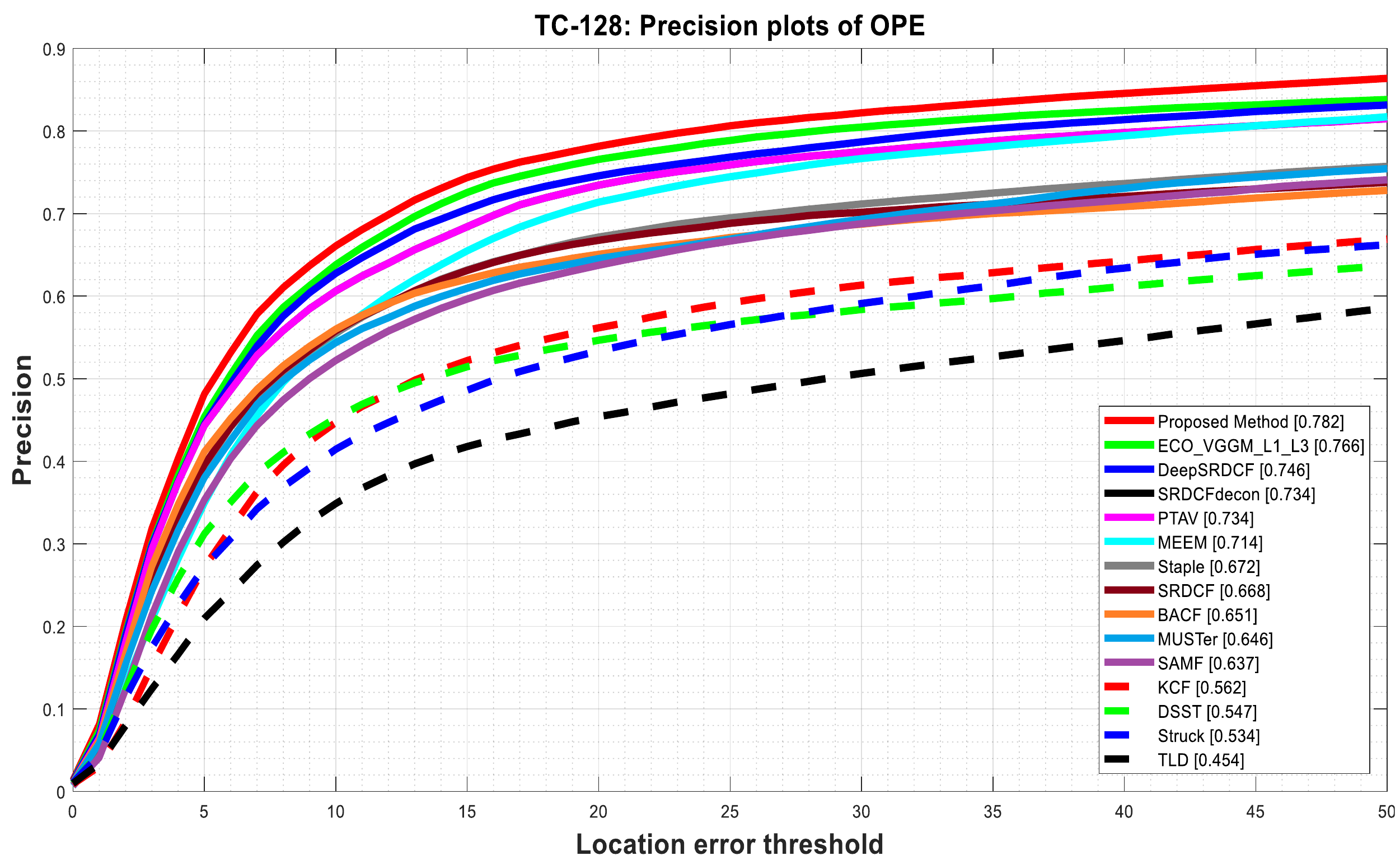}} 
\hspace{0mm}
\subfigure{\includegraphics[width=5.9cm, height=3.8cm]{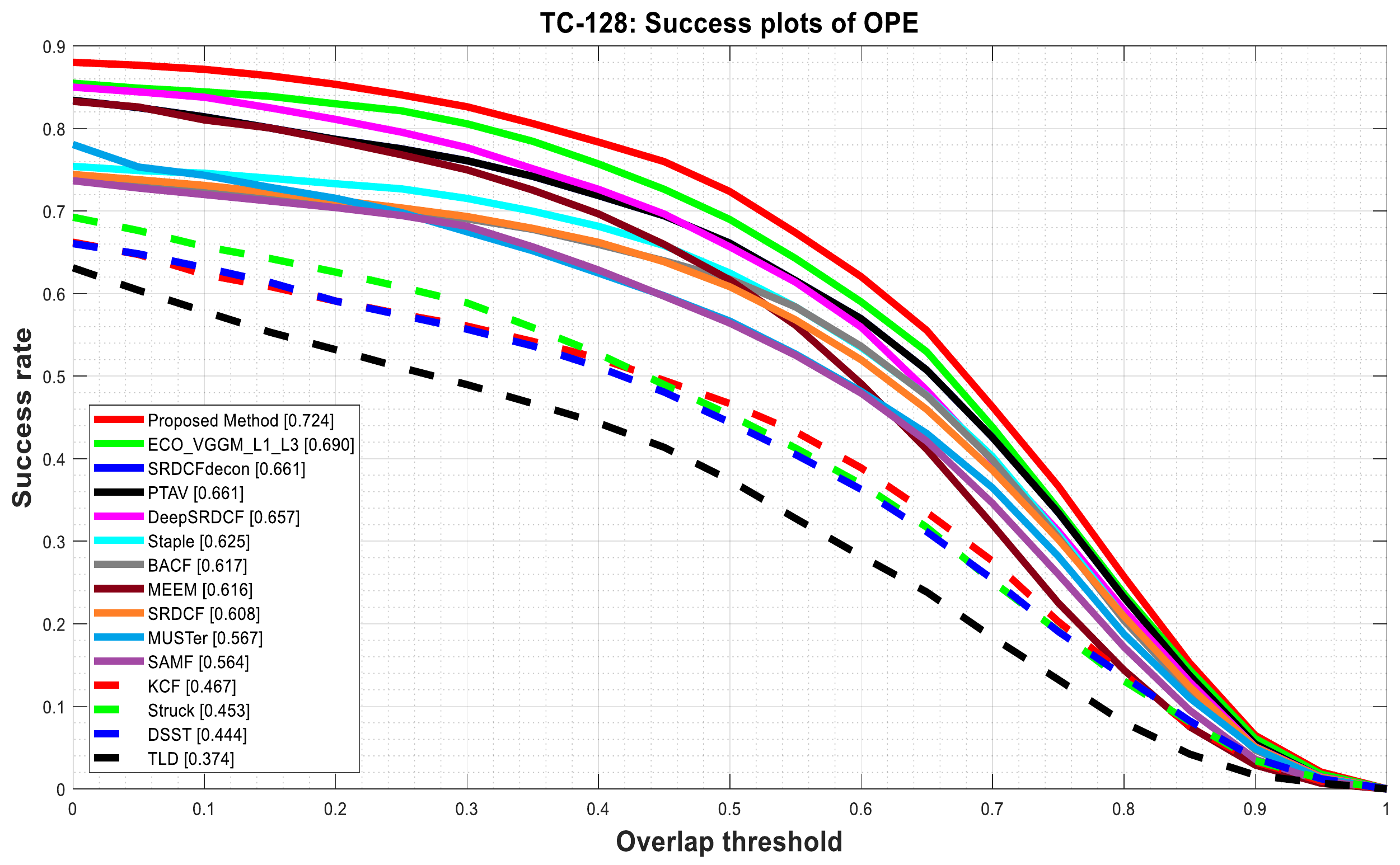}}
\vspace{-4mm}
\caption{Overall precision and success evaluations on the OTB-2013, OTB-2015, and TC-128 datasets.}
\end{figure*}
%%%%%%%%%%%%%%%%%%%%%%%%%%%%%%%%%%%%%%%%%%%%%%%%%%%%%%%%%%%%%%%%%%%%%%%%%%%%%%%%%%%%%%%%%%%%%%%%%%%%%%%
\begin{figure*}
\justify
\subfigure{\includegraphics[width=3.8cm, height=3cm]{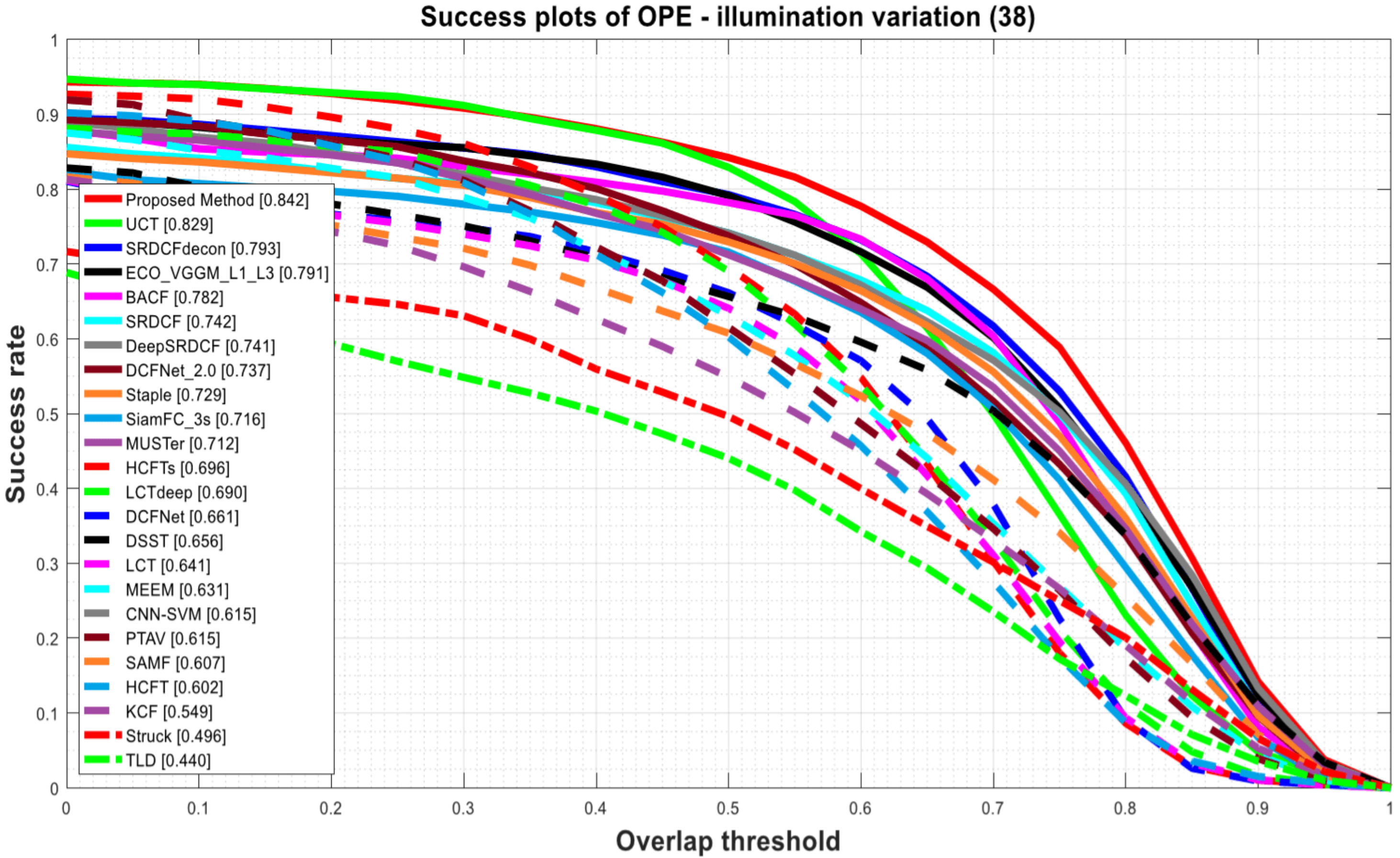}} 
\hspace{0mm}
\subfigure{\includegraphics[width=3.8cm, height=3cm]{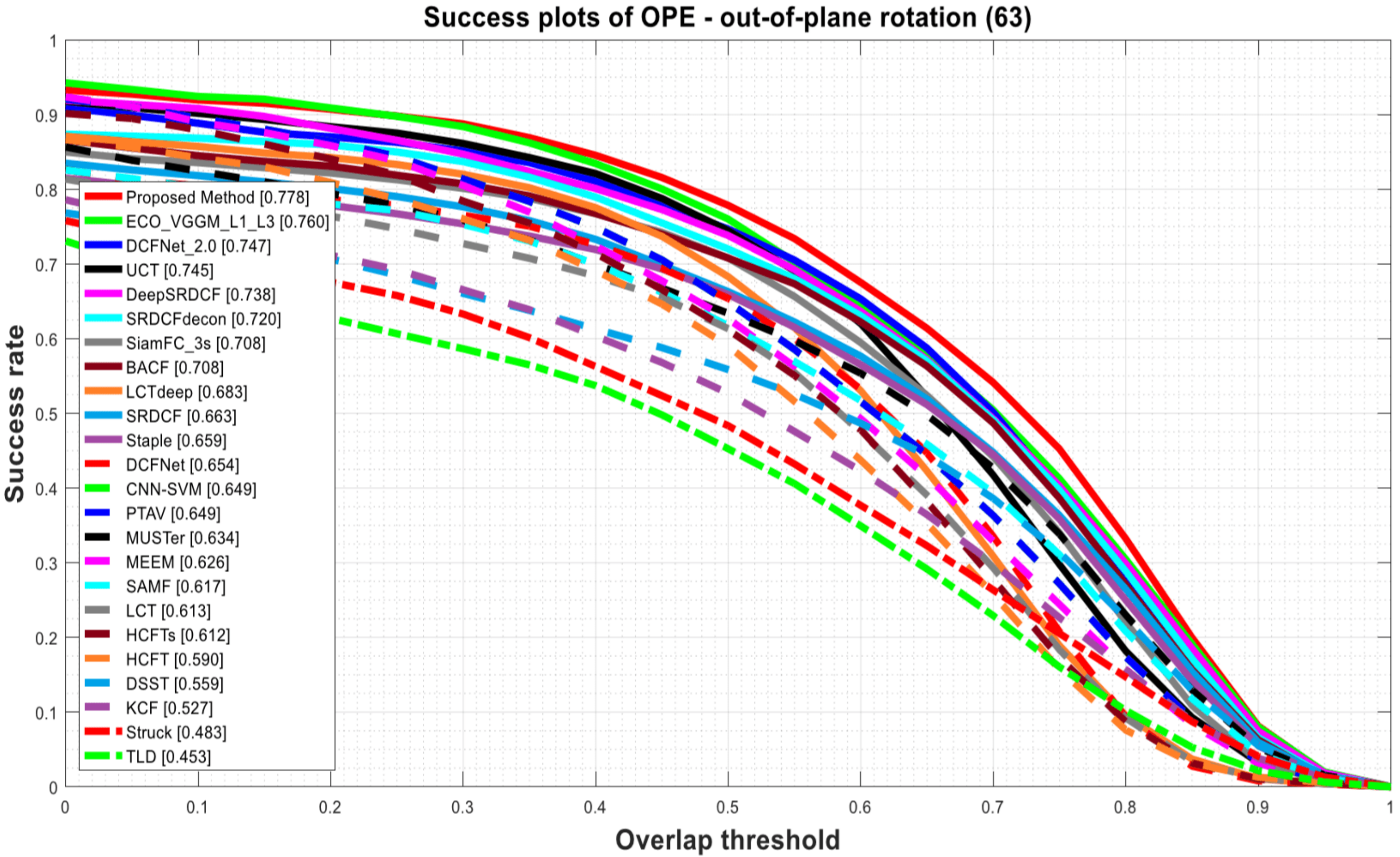}}
\hspace{0mm}
\subfigure{\includegraphics[width=3.8cm, height=3cm]{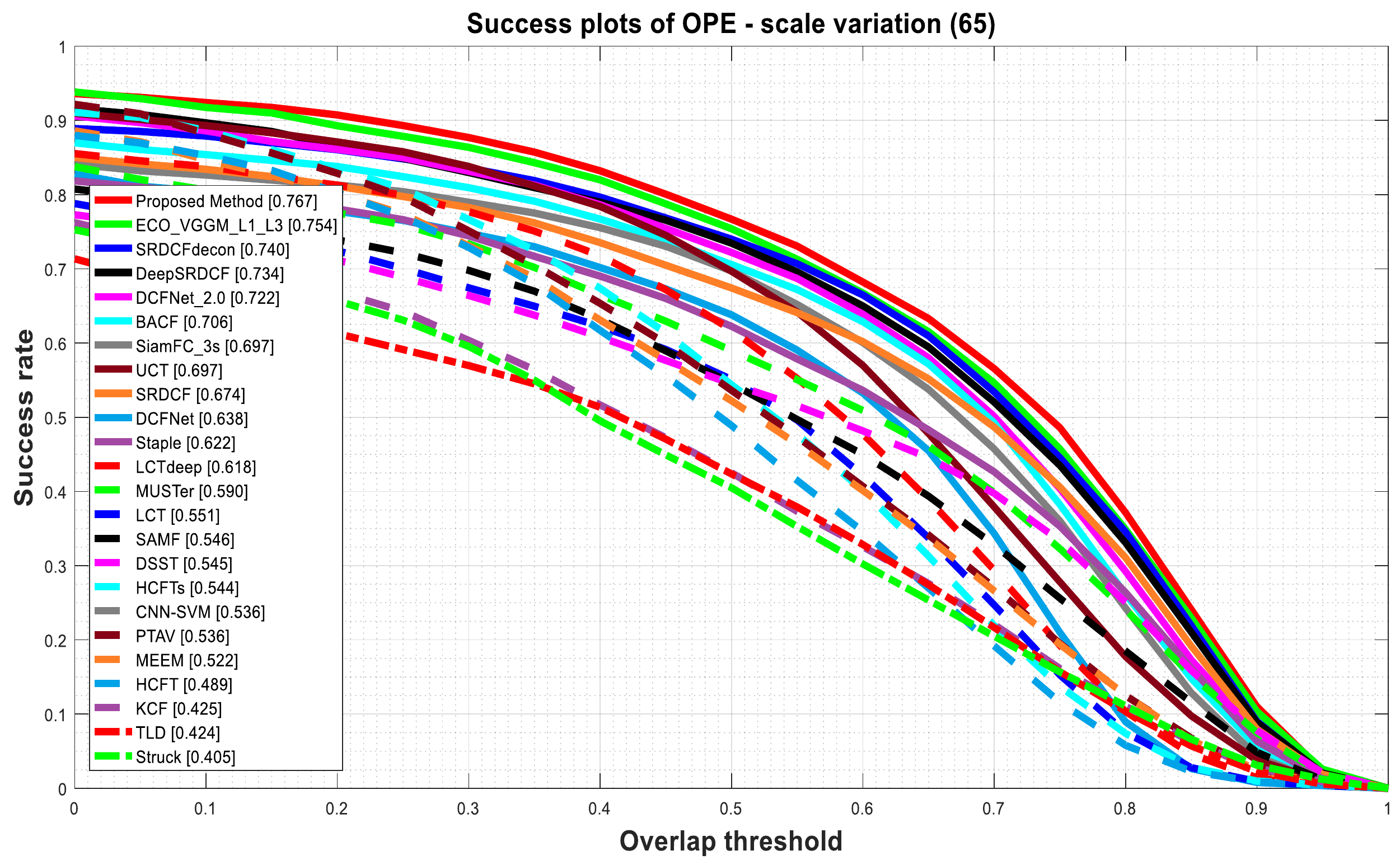}}
\vspace{-4mm}
\justify
\subfigure{\includegraphics[width=3.8cm, height=3cm]{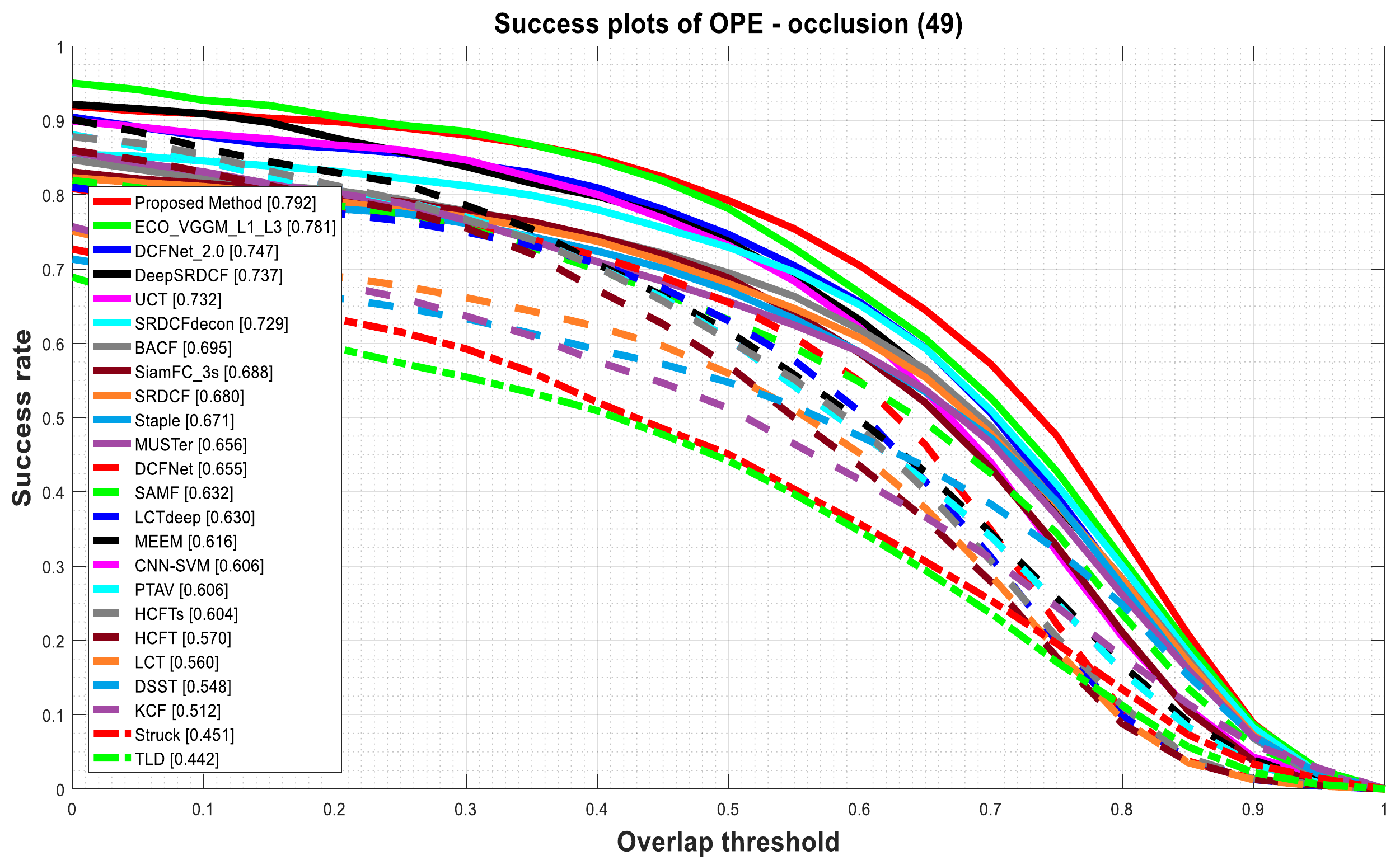}} 
\hspace{0mm}
\subfigure{\includegraphics[width=3.8cm, height=3cm]{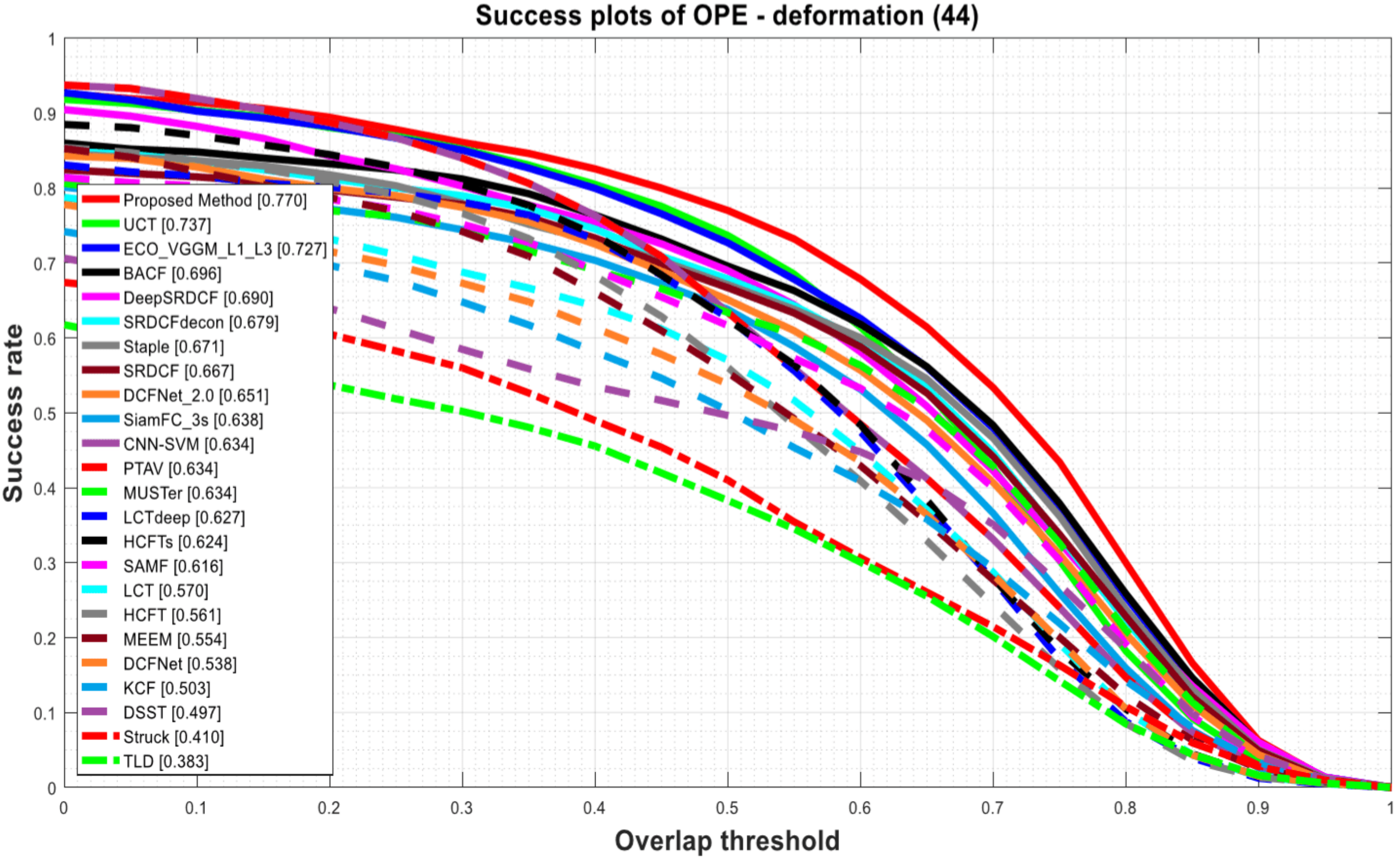}}
\hspace{0mm}
\subfigure{\includegraphics[width=3.8cm, height=3cm]{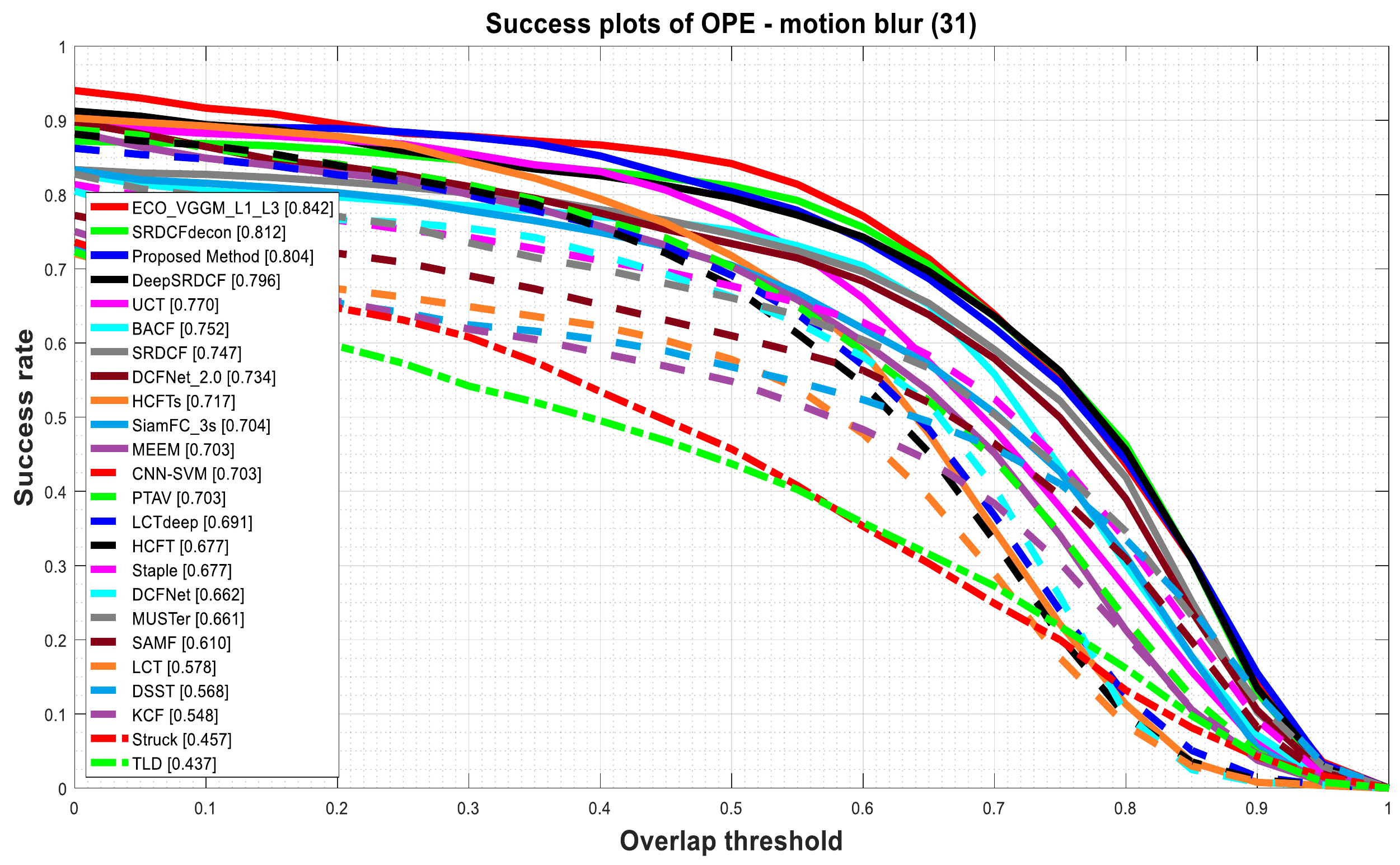}}
\vspace{-4mm}
\justify
\subfigure{\includegraphics[width=3.8cm, height=3cm]{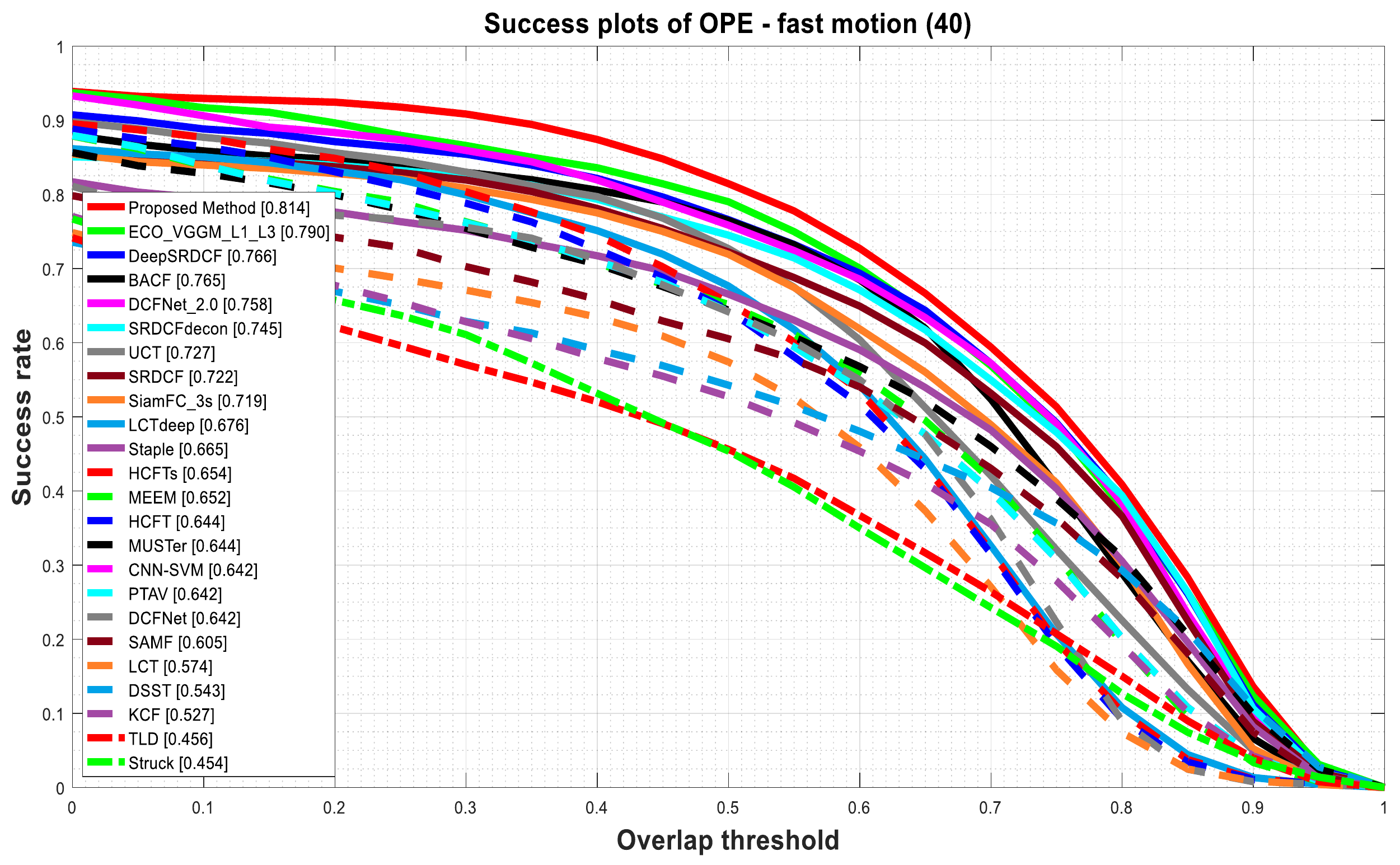}} 
\hspace{0mm}
\subfigure{\includegraphics[width=3.8cm, height=3cm]{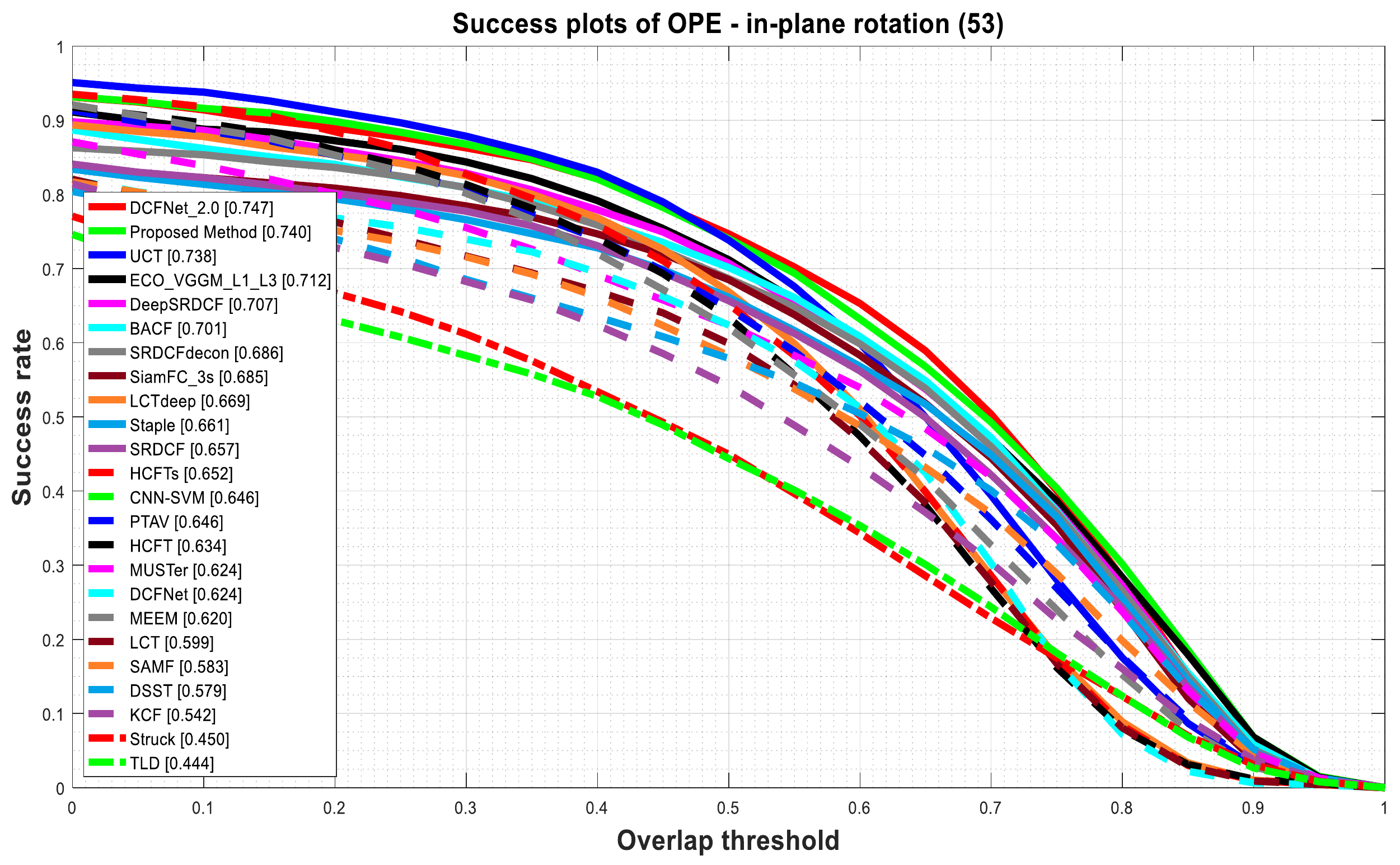}}
\hspace{0mm}
\subfigure{\includegraphics[width=3.8cm, height=3cm]{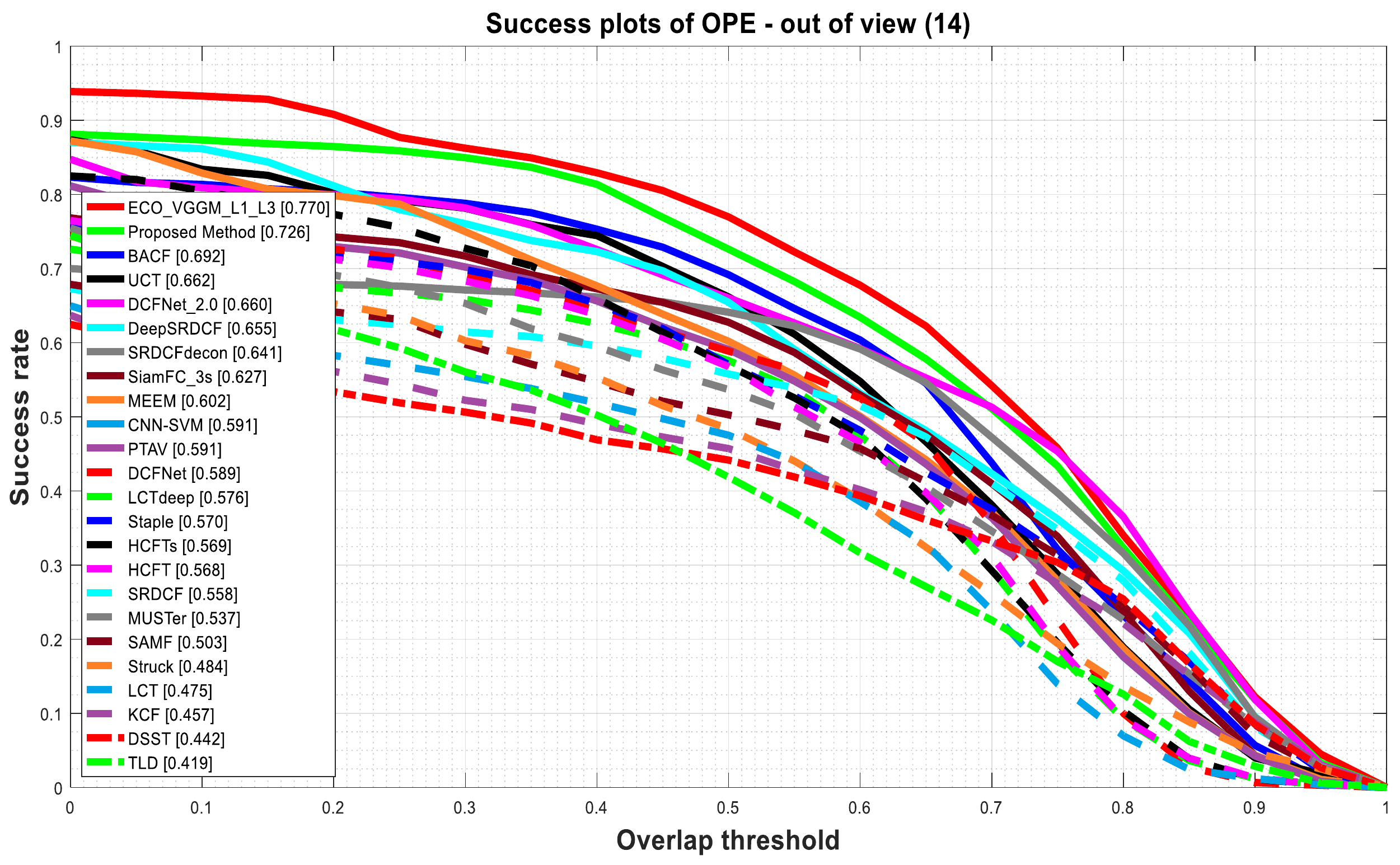}}
\vspace{-4mm}
\centering
\subfigure{\includegraphics[width=3.8cm, height=3cm]{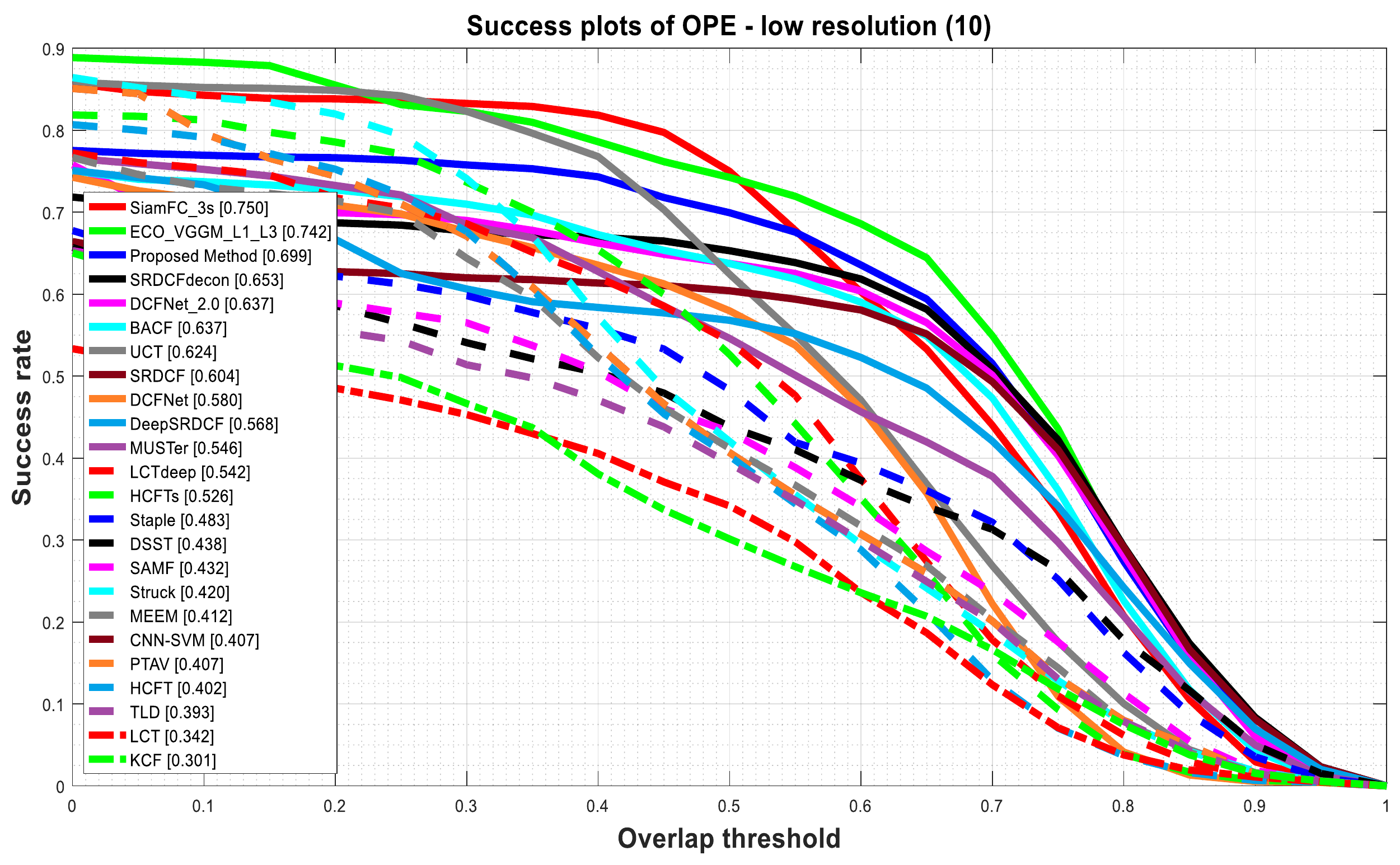}}
\hspace{0mm}
\subfigure{\includegraphics[width=3.8cm, height=3cm]{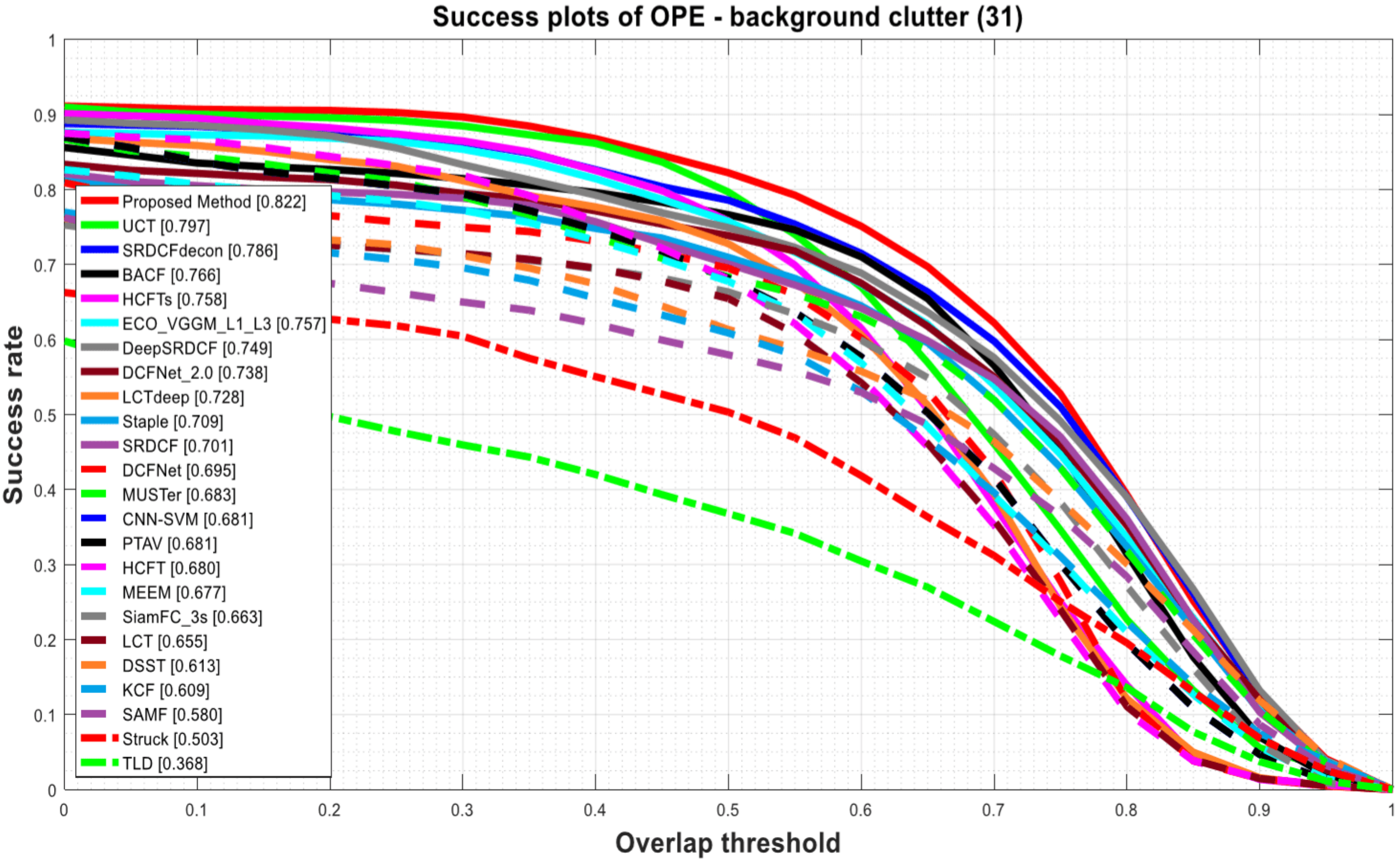}}
\vspace{1mm}
\caption{Attribute-based evaluations of visual trackers in terms of average success rate on the OTB-2015 dataset.}
\end{figure*}
%%%%%%%%%%%%%%%%%%%%%%%%%%%%%%%%%%%%%%%%%%%%%%%%%%%%%%%%%%%%%%%%%%%%%%%%%%%%%%%%%%%%%%%%%%%%%%%%%%%%%%%
\begin{figure*}
\justify
\subfigure{\includegraphics[width=6cm, height=3.6cm]{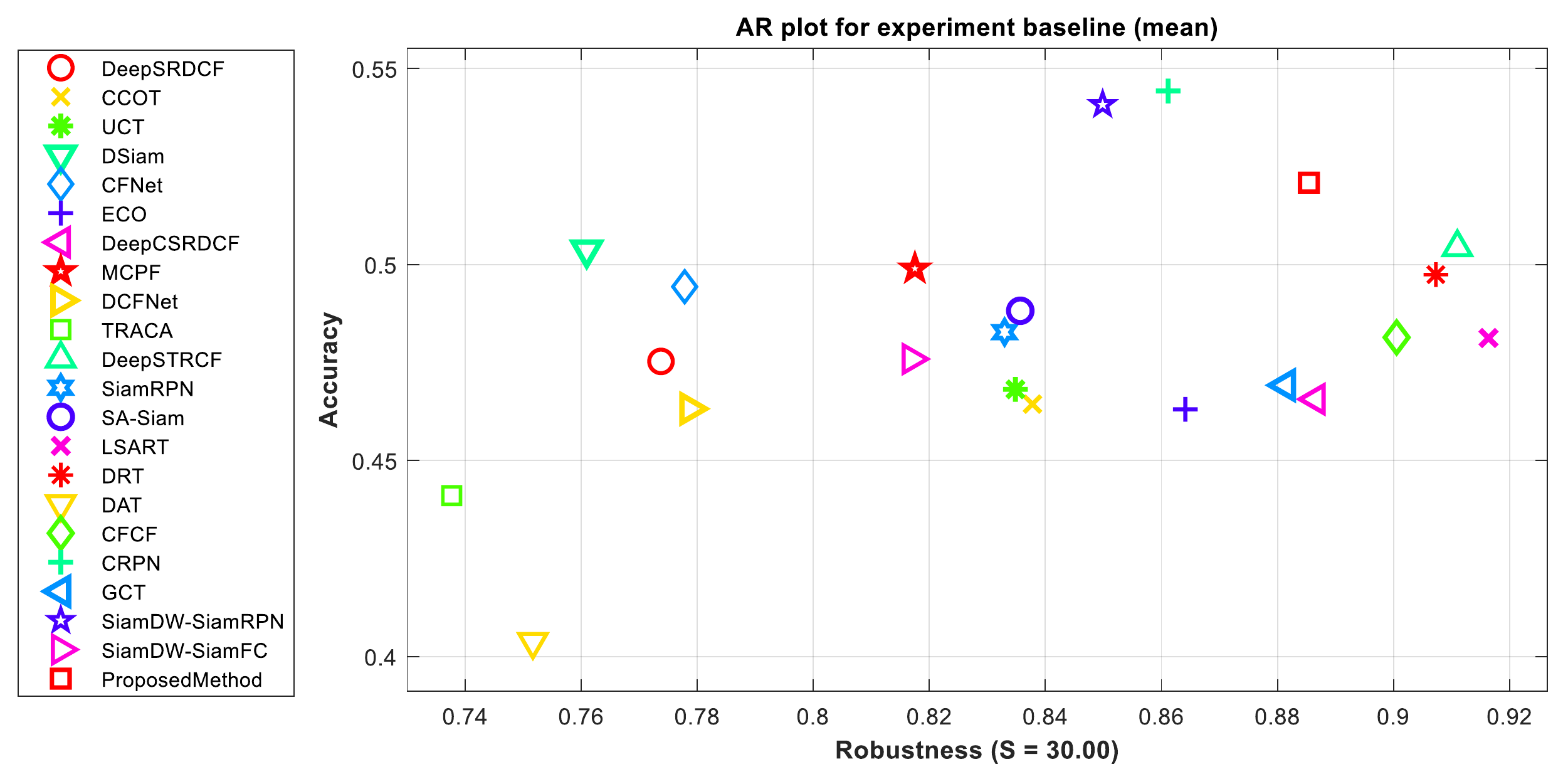}} 
\hspace{0mm}
\subfigure{\includegraphics[width=6cm, height=3.6cm]{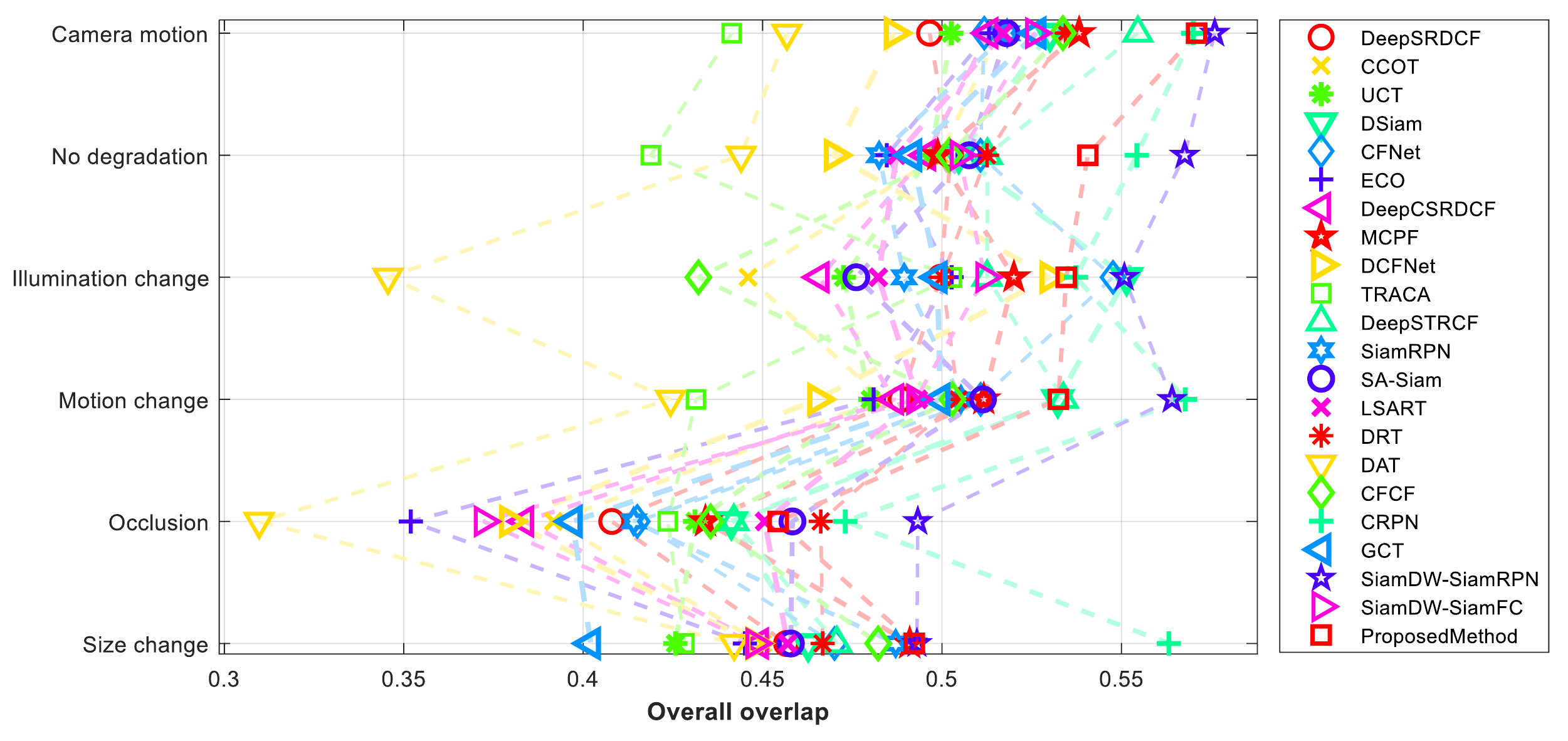}}
\vspace{-6mm}
\caption{Overall and attribute-based performance comparison of visual tracking methods on VOT-2018 dataset.}
\end{figure*}
%%%%%%%%%%%%%%%%%%%%%%%%%%%%%%%%%%%%%%%%%%%%%%%%%%%%%%%%%%%%%%%%%%%%%%%%%%%%%%%%%%%%%%%%%%%%%%%%%%%%%%%

Based on the results of the attribute-based comparison on the OTB-2015 dataset, the proposed method has shown to improve the average success rate of the ECO tracker by up to 5.1\%, 1.8\%, 1.3\%, 1.1\%, 4.3\%, 2.4\%, 2.8\%, and 6.5\% for IV, OPR, SV, OCC, DEF, FM, IPR, and BC attributes, respectively. However, the ECO tracker has achieved better visual tracking performance for MB, OV, and LR attributes. The proposed method outperforms other state-of-the-art visual trackers in most challenging attributes, and the results indicate that the proposed method has improved the average precision rate by up to 3.8\%, 1.1\%, and 1.6\% and the average success rate by up to 5.3\%, 2.6\%, and 3.4\% compared to the ECO-CNN tracker on the OTB-2013, OTB-2015, and TC-128 datasets, respectively. Thus, in addition to providing better visual tracking performance compared to the FEN-based visual trackers, the proposed method outperforms the EEN-based visual tracking methods. For example, the proposed method achieved up to 3.7\%, 7\%, 7.2\%, 11.6\%, and 15.2\% on the OTB-2015 dataset in terms of average precision rate compared to the UCT, DCFNet-2.0, PTAV, SiamFC-3s, and DCFNet trackers, respectively. Furthermore, the success rate of the proposed method gained by up to 4.3\%, 4.8\%, 16.4\%, 8.2\%, and 13.9\%, respectively, compared to the aforementioned EEN-based trackers. Because of the higher performance of deep visual trackers compared to the handcrafted-based ones, the evaluations on the VOT-2018 compare the proposed method just with the state-of-the-art deep visual tracking methods. The achieved results in Fig. 5 demonstrate the proficiency of the proposed method for visual tracking. According to these results, the proposed method not only considerably improve the performance of ECO-CNN framework but also has a competitive results compared to the EEN-based methods which have trained on large-scale datasets. 

The qualitative comparison between the proposed method and the top-5 visual trackers (i.e., UCT, DCFNet-2.0, ECO-CNN, HCFTs, and DeepSRDCF) is shown in Fig. 5, which clearly demonstrates remarkable robustness of the proposed method in real-world scenarios. It is noteworthy that the proposed method provides an acceptable average speed (about five frames per second (FPS)) compared to the most FEN-based visual trackers, which run less than one FPS \cite{MDNet,CCOT}. 
%%%%%%%%%%%%%%%%%%%%%%%%%%%%%%%%%%%%%%%%%%%%%%%%%%%%%%%%%%%%%%%%%%%%%%%%%%%%%%%%%%%%%%%%%%%%%%%%%%%%%%%
\begin{figure*}
\justify
\includegraphics[width=11.8cm, height=14cm]{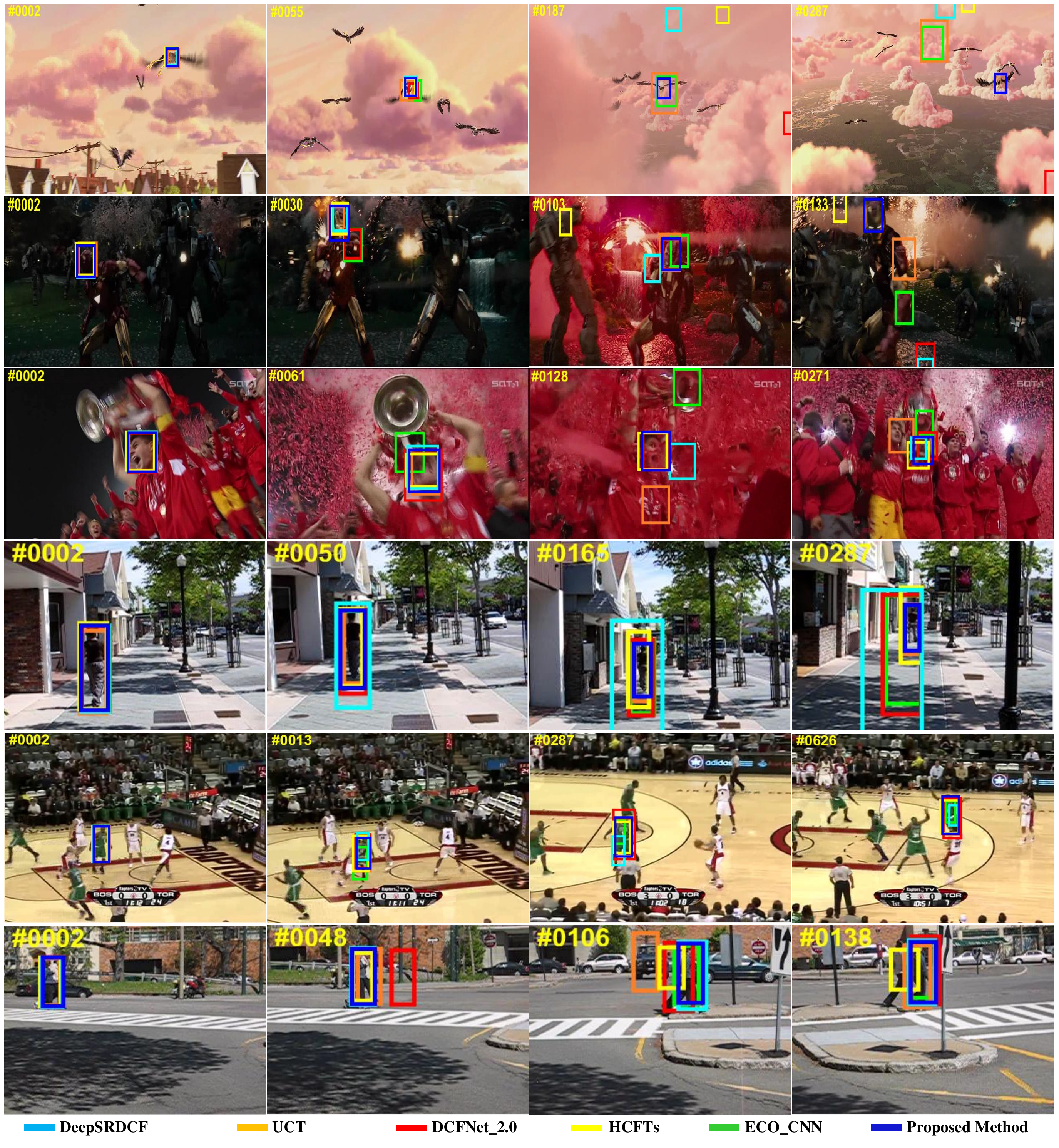}
\vspace{-1mm}
\caption{Qualitative evaluation results of visual tracking methods on Bird1, Ironman, Soccer, Human9, Basketball, and Couple video sequences from top to bottom row, respectively.}\label{QualitativeComparison}
\vspace{-4mm}
\end{figure*}
%%%%%%%%%%%%%%%%%%%%%%%%%%%%%%%%%%%%%%%%%%%%%%%%%%%%%%%%%%%%%%%%%%%%%%%%%%%%%%%%%%%%%%%%%%%%%%%%%%%%%%%
To robustly model the visual targets, the proposed method uses two main components. First, the proposed method exploits the weighted fused deep representations in the learning process of convolution filters in the continuous domain. It extracts deep feature maps from two FENs (i.e., DenseNet-201 and FCN-8s) that are trained on different large-scale datasets (i.e., ImageNet, PASCAL VOC, and MSCOCO) for different tasks of object recognition and semantic segmentation. These rich representations provide complementary information for an unknown visual target. Hence, the fused deep representation leads to a more robust target model, which is more resist against challenging attributes, such as heavy illumination variation, occlusion, deformation, background clutter, and fast motion. Moreover, the employed FENs are efficient and do not have complications with integrating into the proposed DCF-based tracker. Second, the proposed method uses the extracted feature maps from the FCN-8s network to semantically enhance the weighting process of target representations. This component helps the proposed method to model the visual target more accurately and prevent contamination of target appearance with the background or visual distractors. Hence, the proposed method demonstrates more robustness against significant variations of a visual target, including deformation and background clutter.
\vspace{-4mm}
\subsection{Performance Comparison: Ablation Study}
\label{sec:5.2}
\vspace{-4mm}
To understand the advantages and disadvantages of each proposed component, the visual tracking performance of two ablated trackers are investigated and compared to the proposed method. The ablated trackers are the proposed method to disable the fusion process (i.e., this method just exploits the DenseNet-201 representations and semantic windowing process) and the proposed method without semantic windowing process (i.e., this method only fuses deep multitasking representations from the DenseNet-201 and FCN-8s networks to ensure robust target appearance).
%%%%%%%%%%%%%%%%%%%%%%%%%%%%%%%%%%%%%%%%%%%%%%%%%%%%%%%%%%%%%%%%%%%%%%%%%%%%%%%%%%%%%%%%%%%%%%%%%%%%%%%
\begin{figure*}
\justify
\subfigure{\includegraphics[width=5.8cm, height=4.5cm]{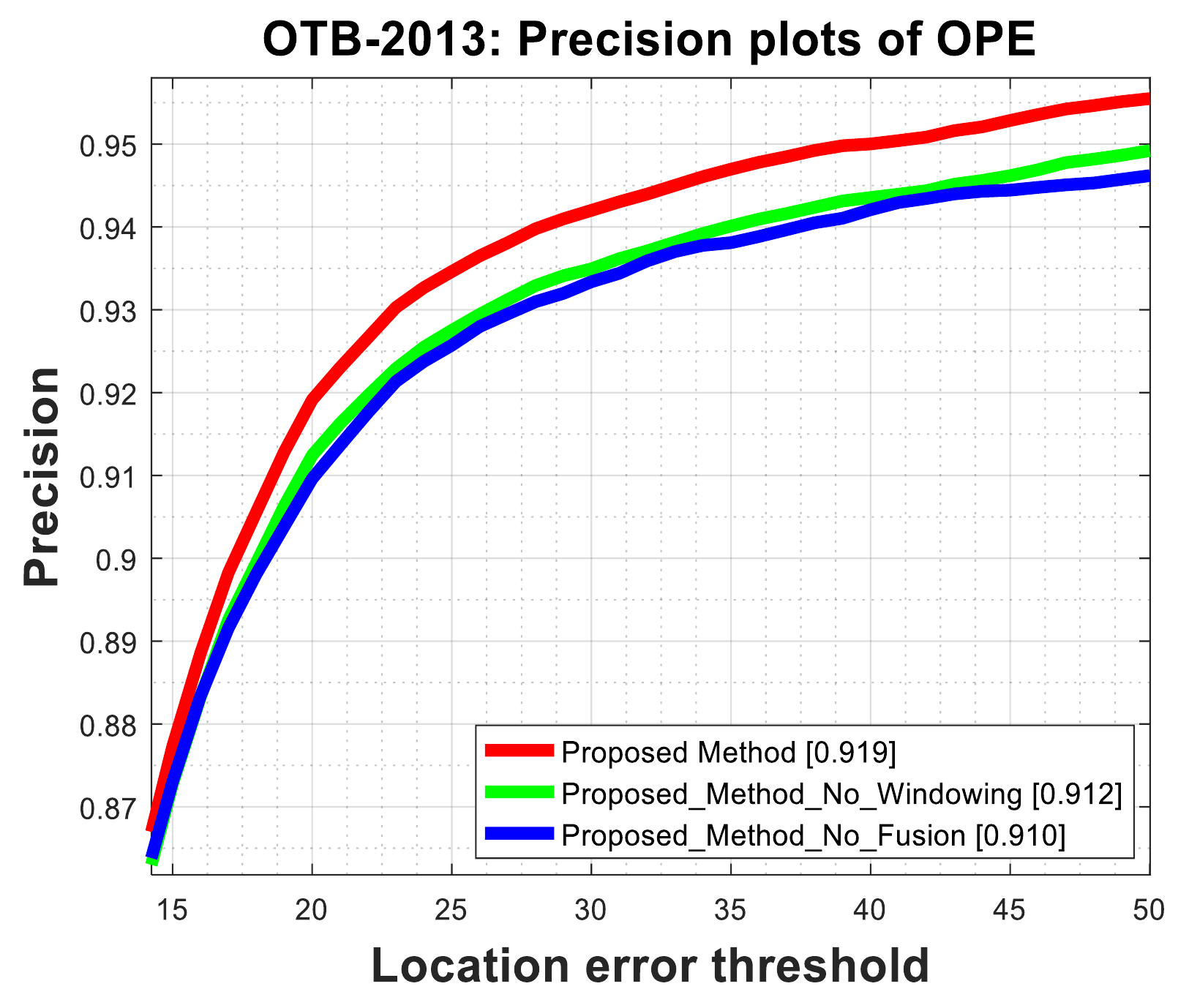}} 
\hspace{0mm}
\subfigure{\includegraphics[width=5.8cm, height=4.5cm]{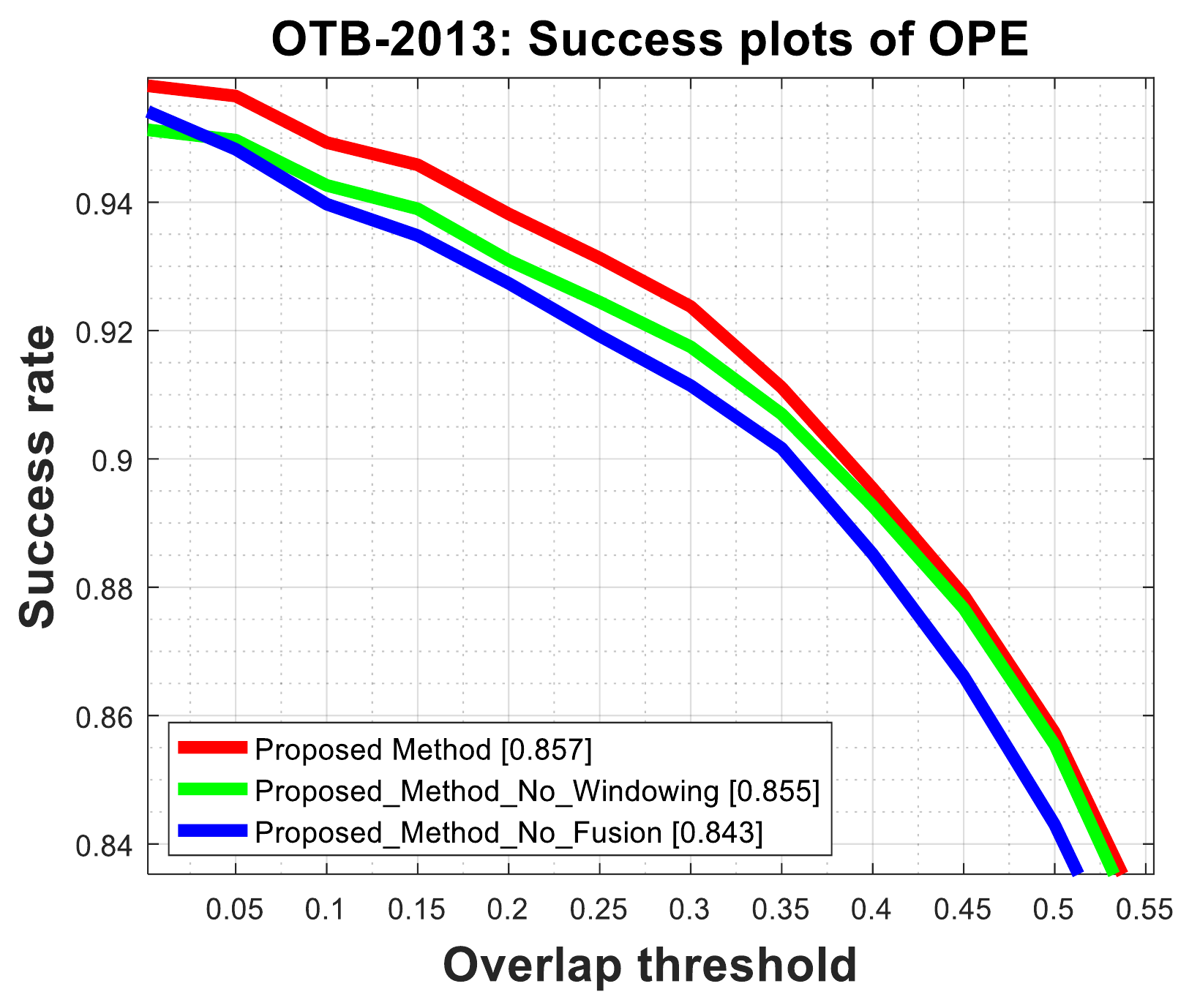}}
\vspace{-4mm}
\justify
\subfigure{\includegraphics[width=5.8cm, height=4.5cm]{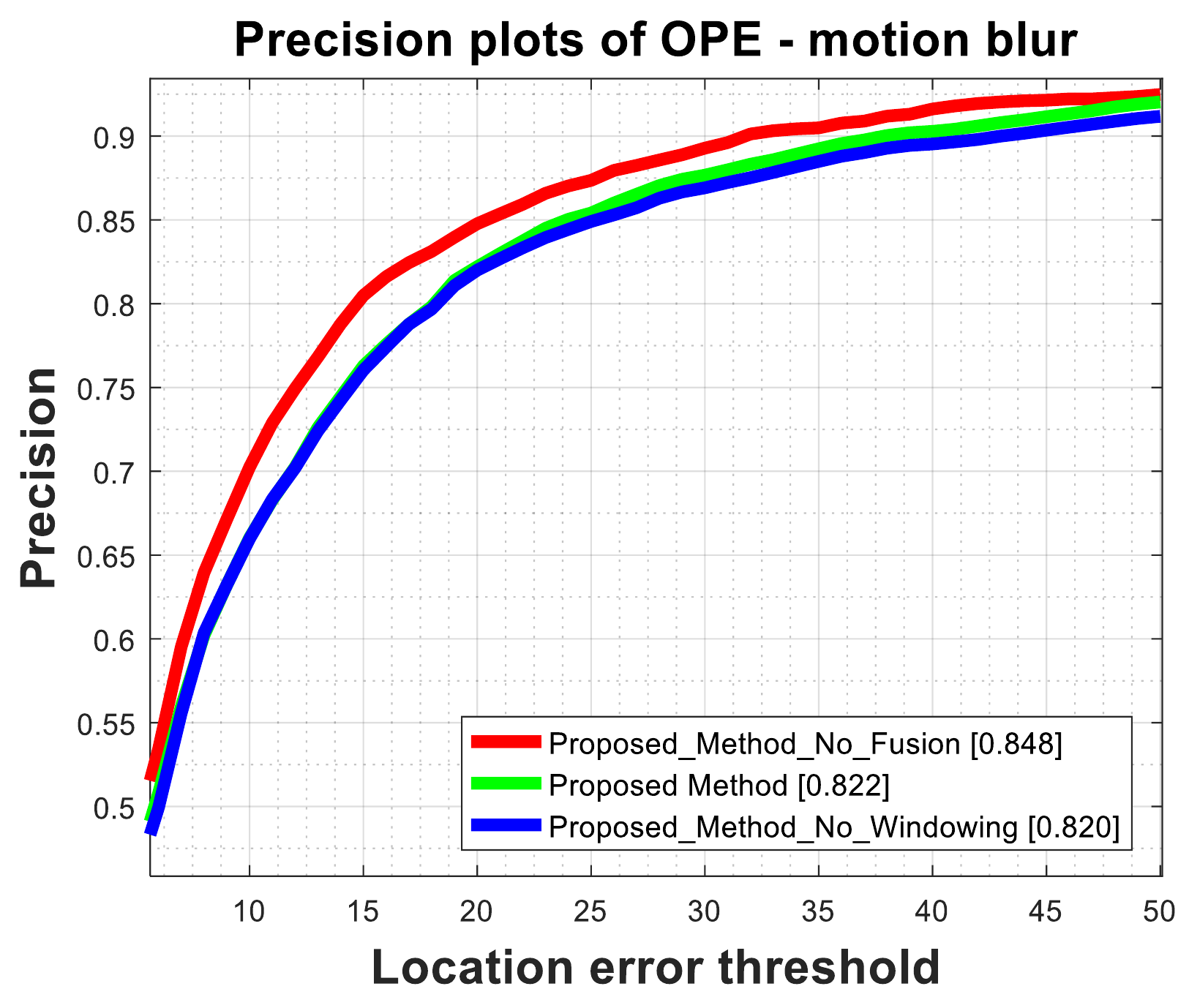}} 
\hspace{0mm}
\subfigure{\includegraphics[width=5.8cm, height=4.5cm]{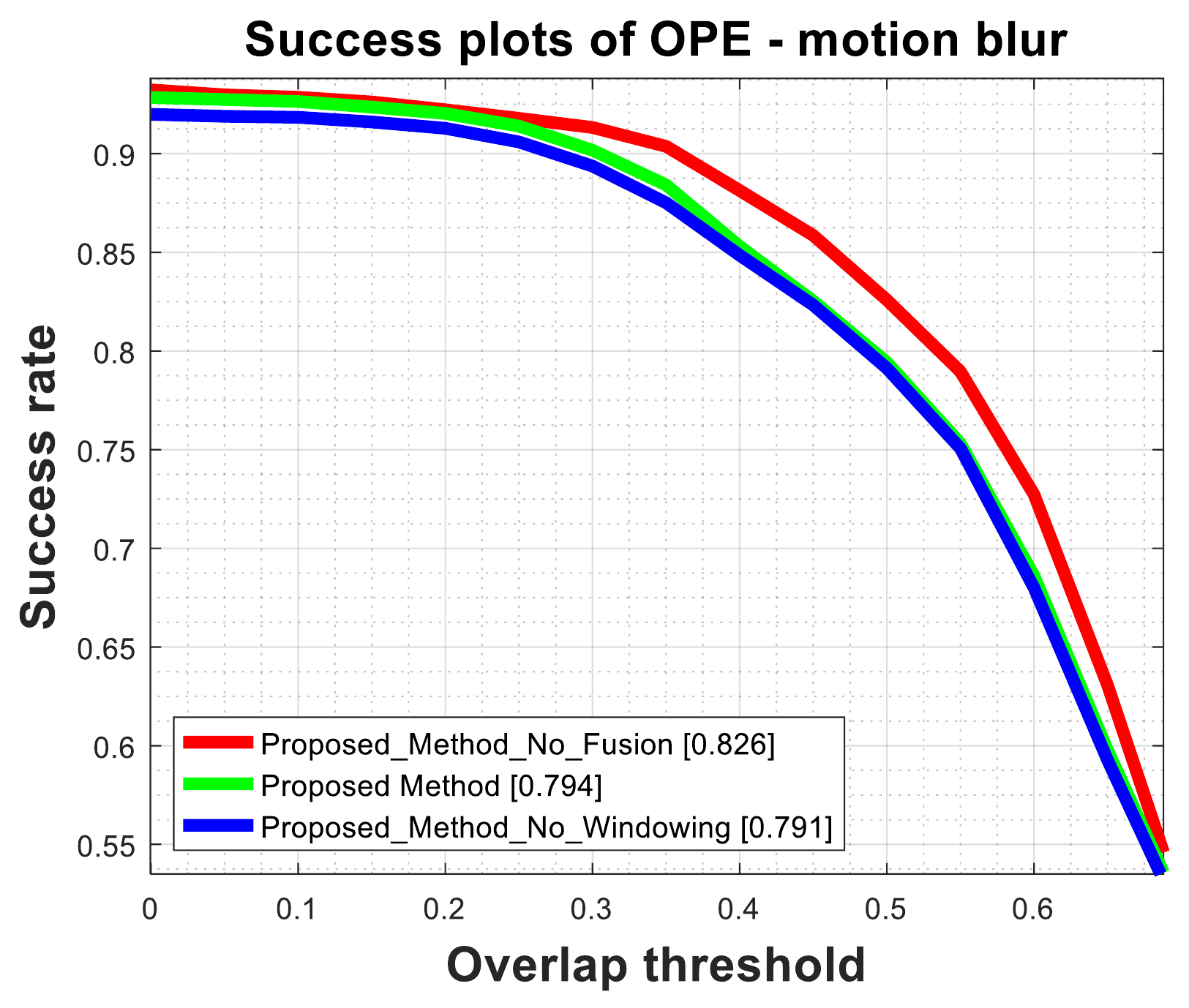}}
\vspace{-4mm}
\justify
\subfigure{\includegraphics[width=5.8cm, height=4.5cm]{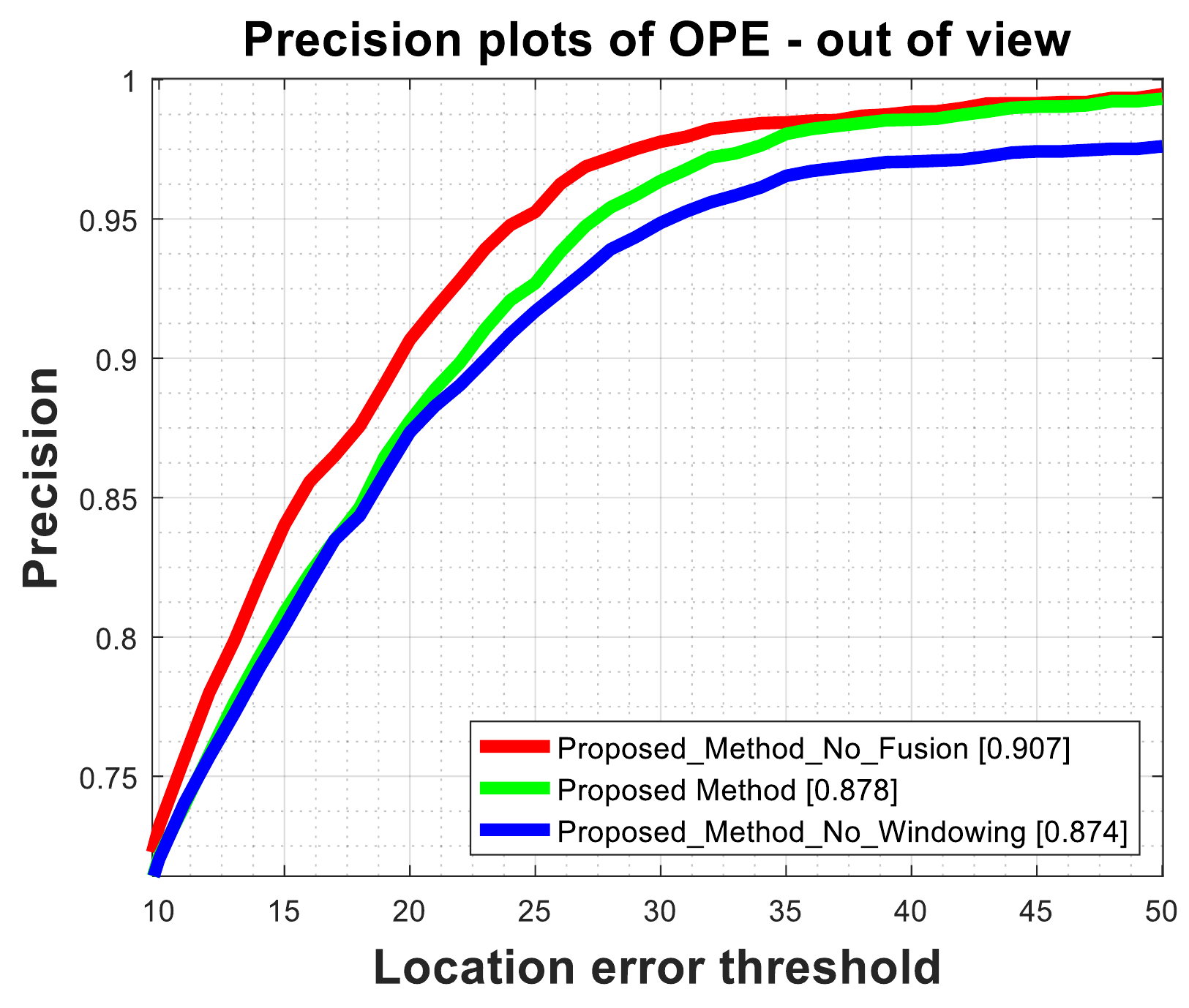}} 
\hspace{0mm}
\subfigure{\includegraphics[width=5.8cm, height=4.5cm]{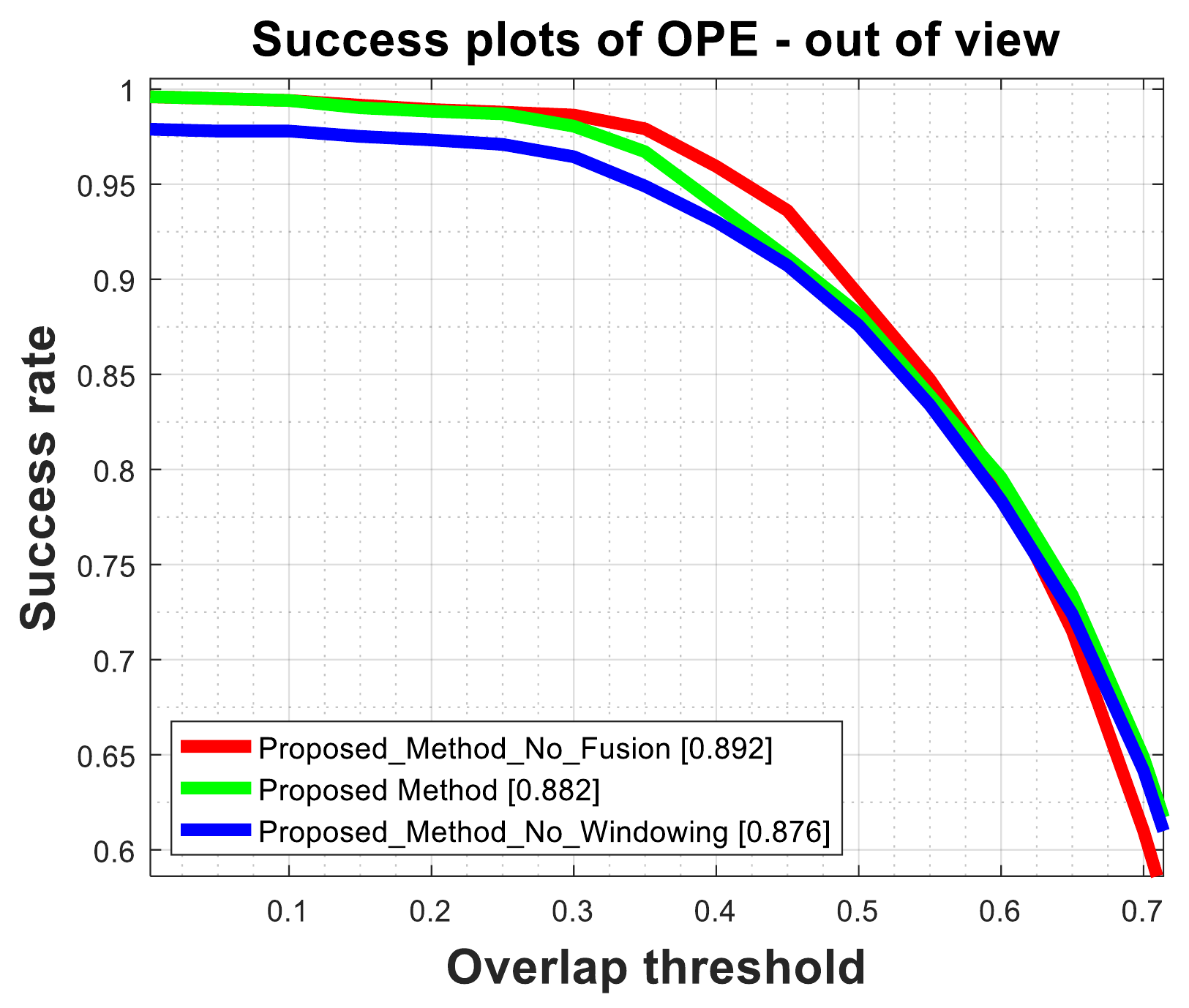}}
\vspace{-4mm}
\justify
\subfigure{\includegraphics[width=5.8cm, height=4.5cm]{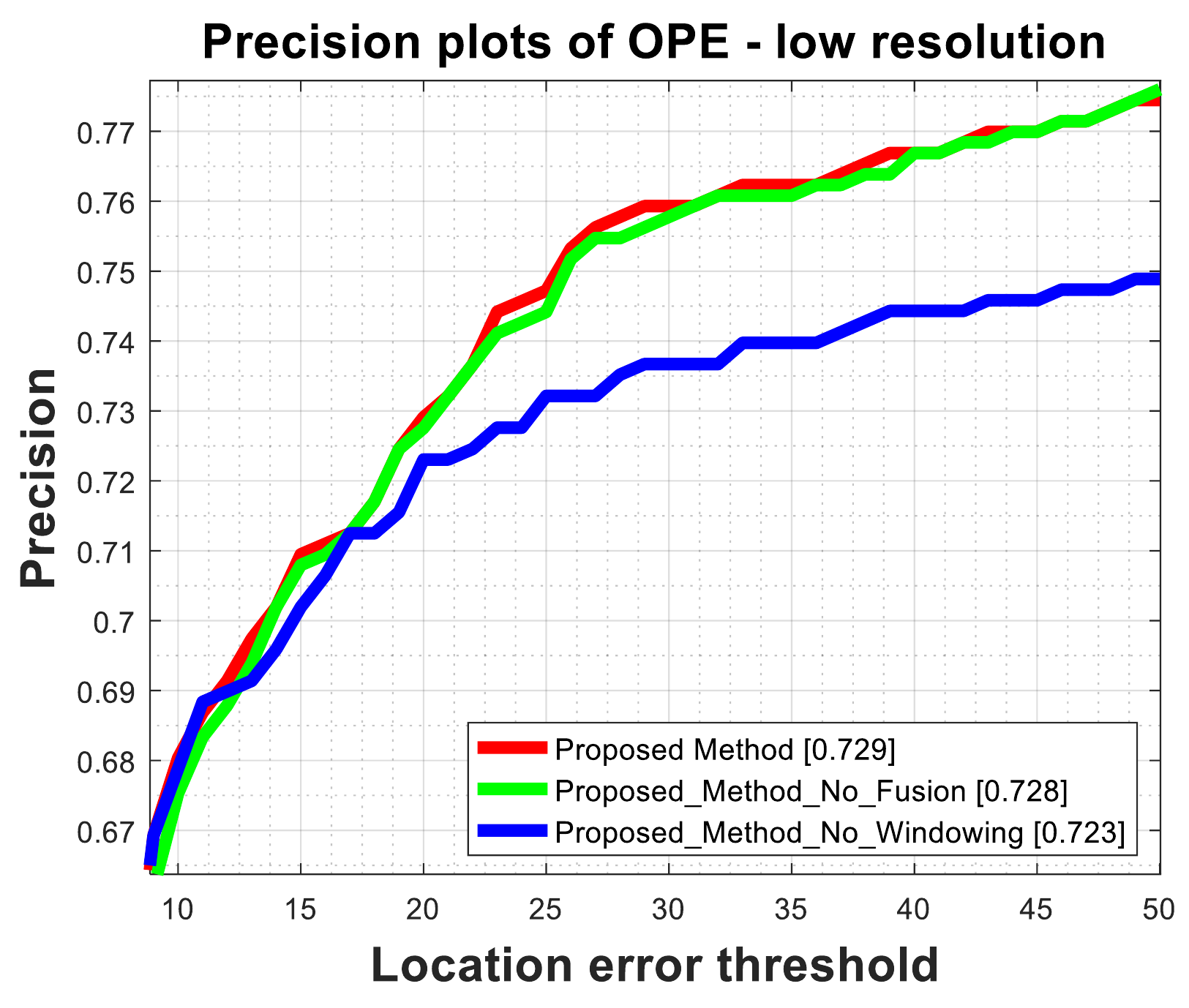}} 
\hspace{0mm}
\subfigure{\includegraphics[width=5.8cm, height=4.5cm]{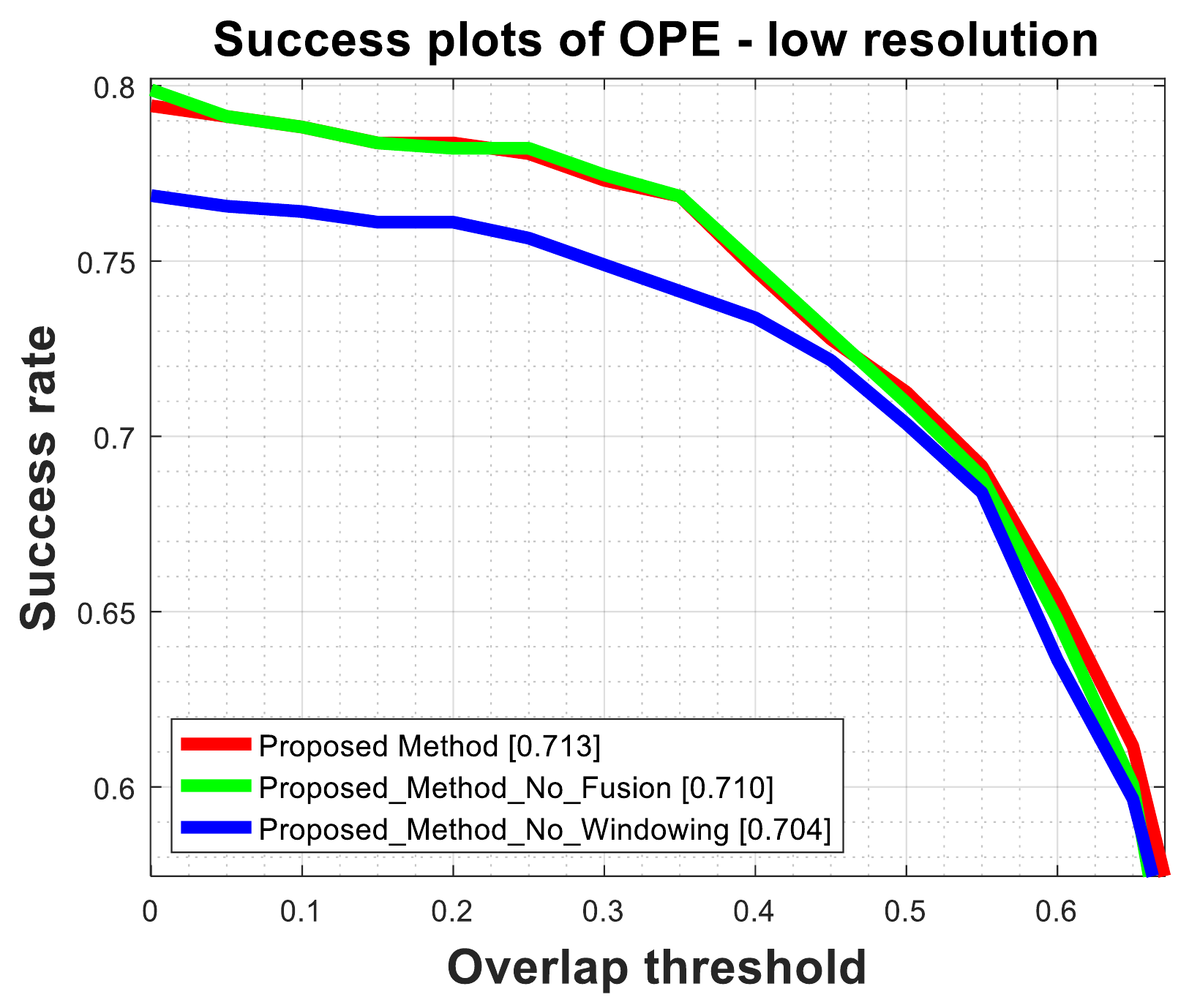}}
\vspace{-2mm}
\caption{Overall and attribute-based ablation study of proposed visual-tracking method on the OTB-2013 dataset.}
\vspace{-4mm}
\end{figure*}
%%%%%%%%%%%%%%%%%%%%%%%%%%%%%%%%%%%%%%%%%%%%%%%%%%%%%%%%%%%%%%%%%%%%%%%%%%%%%%%%%%%%%%%%%%%%%%%%%%%%%%%

The results of the ablation study on the OTB-2013 dataset are shown in Fig. 7. Based on the precision and success plots of the overall comparison, it can be concluded that most of the success of the proposed method is related to the effective fusion process of desirable representations for visual tracking. Moreover, it is clearly demonstrated that the proposed windowing process enhances the proposed method, particularly in terms of precision metrics. To identify the reason for the performance degradation faced for MB, OV, and LR attributes, an attribute-based evaluation was performed. The fusion process provides the main contribution to reducing visual tracking performance when these attributes occur, as seen in the results in Fig. 7. These results confirm that the semantic windowing process can effectively support visual trackers in learning more discriminative target appearance and in preventing drift problems.
%%%%%%%%%%%%%%%%%%%%%%%%%%%%%%%%%%%%%%%%%%%%%%%%%%%%%%%%%%%%%%%%%%%%%%%%%%%%%%%%%%%%%%%%%%%%%%%%%%%%%%%%%%
\vspace{-4mm}
\section{Conclusion}
\label{sec:6}
The use of twelve state-of-the-art ResNet-based FENs for visual tracking purposes was evaluated. A visual tracking method was proposed; this method can fuse deep features extracted from the DenseNet-201 and FCN-8s networks, but it can also semantically weight target representations in the learning process of continuous convolution filters. Moreover, the best ResNet-based FEN in the DCF-based framework was determined (i.e., DenseNet-201), and the effectiveness of using the DenseNet-201 network on the tracking performance of another DCF-based tracker was explored. Finally, the comprehensive experimental results on the OTB-2013, OTB-2015, TC-128 and VOT-2018 datasets demonstrated that the proposed method outperforms state-of-the-art visual tracking methods in terms of various evaluation metrics for visual tracking.\\

\noindent\textbf{Acknowledgement:} This work was partly supported by a grant (No. 96013046) from Iran National Science Foundation (INSF).\\

\noindent\textbf{Compliance with Ethical Standards (Conflict of Interest):} All authors declare that they have no conflict of interest.

\vspace{-4mm}
\bibliographystyle{plainnat}
\bibliography{ref.bib}
\end{document}